\documentclass[review]{elsarticle}

\usepackage{lineno}
\modulolinenumbers[5]

\usepackage[colorlinks = true,
            linkcolor = blue,
            urlcolor  = blue,
            citecolor = blue,
            anchorcolor = blue]{hyperref}

\usepackage{enumitem}
\usepackage{algorithm}
\usepackage{algorithmic}
\usepackage{graphicx}
\usepackage{caption}
\usepackage{textcomp}
\usepackage{subfig}
\usepackage{multirow}
\usepackage{multicol}
\usepackage{makecell}
\usepackage{longtable}
\usepackage{lipsum}
\usepackage{tabularx}
\usepackage{float}
\usepackage{url}
\usepackage{booktabs}
\usepackage{dblfloatfix}
\usepackage{mathtools}
\usepackage{cuted}
\usepackage{xcolor}
\usepackage[left=2.5cm, right=2.5cm, top=2.5cm]{geometry}
\usepackage{changepage}
\usepackage{gensymb}

\newcolumntype{L}[1]{>{\raggedright\let\newline\\\arraybackslash\hspace{0pt}}m{#1}}
\newcolumntype{C}[1]{>{\centering\let\newline\\\arraybackslash\hspace{0pt}}m{#1}}
\newcolumntype{R}[1]{>{\raggedleft\let\newline\\\arraybackslash\hspace{0pt}}m{#1}}

\journal{Journal of ...}

\biboptions{authoryear}

\begin{document}

\begin{frontmatter}

\title{Applications of Machine Learning in Biopharmaceutical Process Development and Manufacturing: Current Trends, Challenges, and Opportunities}

\author[utsaddress]{\corref{mycorrespondingauthor}Thanh Tung Khuat}\ead{Thanhtung.Khuat@uts.edu.au}

\author[csladdress]{Robert Bassett}
\ead{Robert.Bassett@csl.com.au}

\author[csladdress]{Ellen Otte}
\ead{Ellen.Otte@csl.com.au}

\author[csladdress]{Alistair Grevis-James}
\ead{Alistair.Grevis-James@csl.com.au}

\author[utsaddress]{Bogdan Gabrys}
\ead{Bogdan.Gabrys@uts.edu.au}


\cortext[mycorrespondingauthor]{Corresponding author}

\address[utsaddress]{Complex Adaptive Systems Laboratory, The Data Science Institute, University of Technology Sydney, NSW 2007, Australia}
\address[csladdress]{CSL Innovation, Melbourne, VIC 3000, Australia}

\begin{abstract}
While machine learning (ML) has made significant contributions to the biopharmaceutical field, its applications are still in the early stages in terms of providing direct support for quality-by-design based development and manufacturing of biopharmaceuticals, hindering the enormous potential for bioprocesses automation from their development to manufacturing. However, the adoption of ML-based models instead of conventional multivariate data analysis methods is significantly increasing due to the accumulation of large-scale production data. This trend is primarily driven by the real-time monitoring of process variables and quality attributes of biopharmaceutical products through the implementation of advanced process analytical technologies. Given the complexity and multidimensionality of a bioproduct design, bioprocess development, and product manufacturing data, ML-based approaches are increasingly being employed to achieve accurate, flexible, and high-performing predictive models to address the problems of analytics, monitoring, and control within the biopharma field. This paper aims to provide a comprehensive review of the current applications of ML solutions in a bioproduct design, monitoring, control, and optimisation of upstream, downstream, and product formulation processes. Finally, this paper thoroughly discusses the main challenges related to the bioprocesses themselves, process data, and the use of machine learning models in biopharmaceutical process development and manufacturing. Moreover, it offers further insights into the adoption of innovative machine learning methods and novel trends in the development of new digital biopharma solutions.
\end{abstract}

\begin{keyword}
biopharmaceuticals \sep machine learning \sep upstream \sep downstream \sep bioprocesses \sep digital twin \sep soft sensors
\end{keyword}

\end{frontmatter}


\section{Introduction}
\begin{figure}[!ht]
    \centering
    \includegraphics[width=0.8\textwidth]{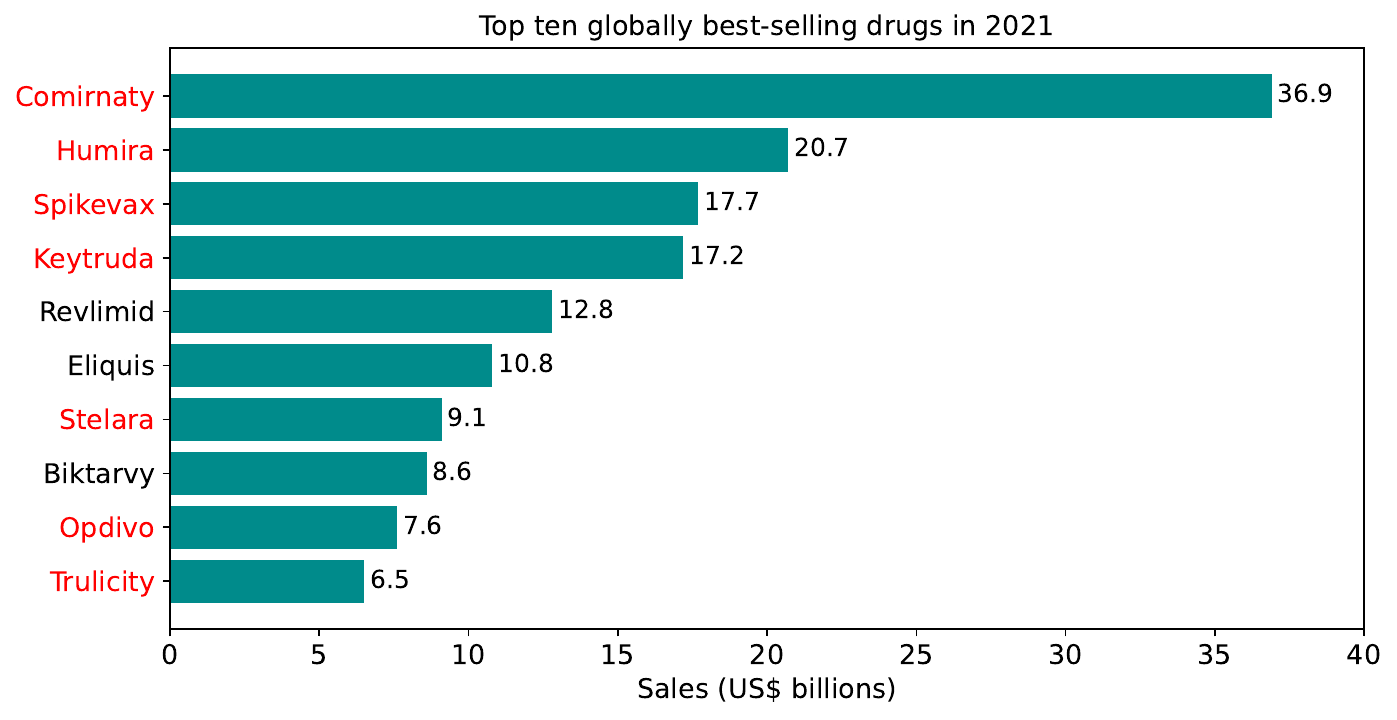}
    \caption{Top ten drugs by sales globally in 2021 (produced from the data in \cite{ur22}). The biological products are highlighted by red color.}
    \label{fig_best_selling_21}
\end{figure}

\begin{figure}[!ht]
    \centering
    \includegraphics[width=0.95\textwidth]{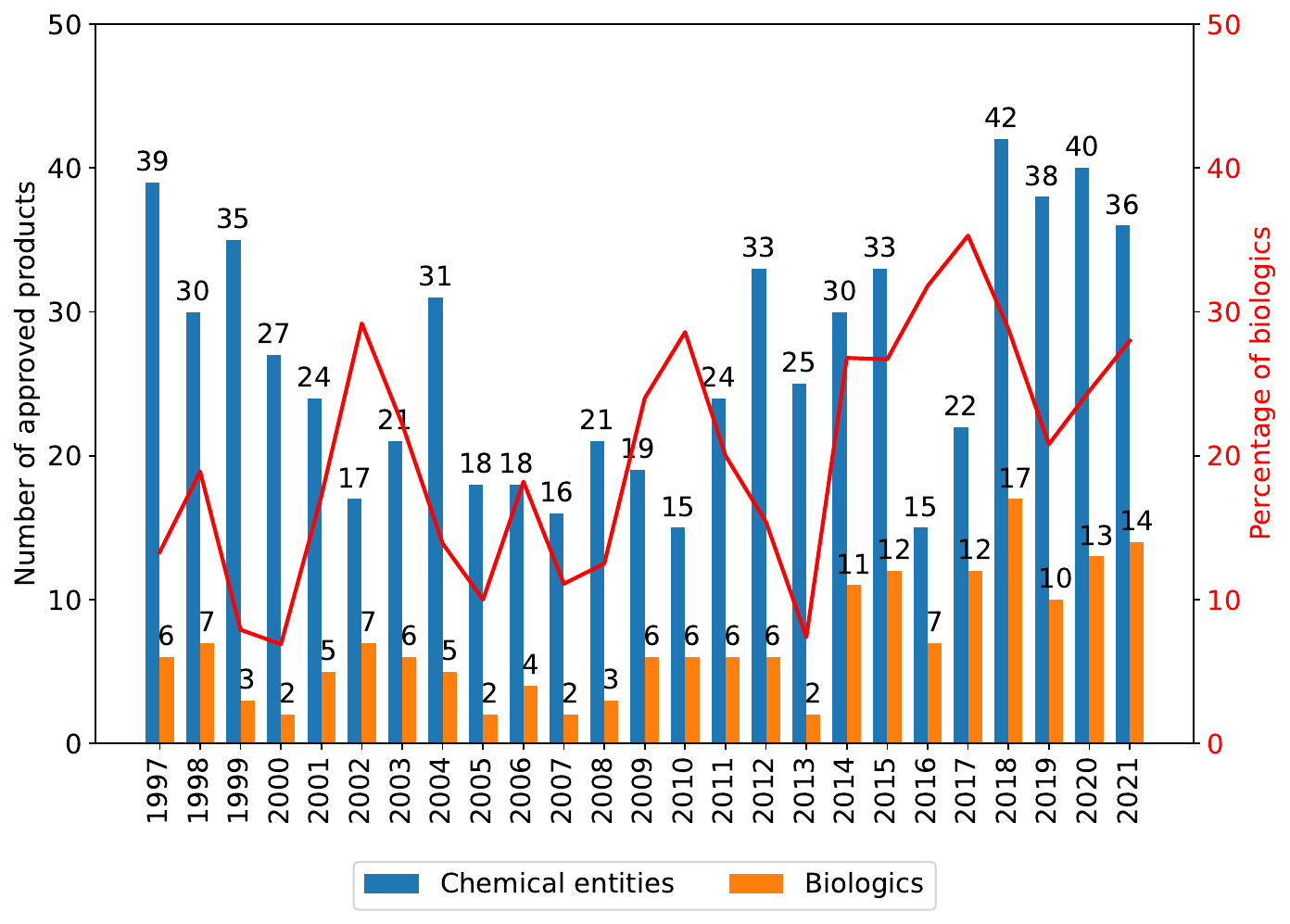}
    \caption{New drugs have been approved by FDA over the last 25 years and the percentage of biologics over total drug approved each year (adapted from the data in \cite{deal22}).}
    \label{fig_products_approved}
\end{figure}

Over the past few years, biopharmaceutical products, also known as biologics, such as monoclonal antibodies (mAbs) and therapeutic proteins have become the best-selling drugs in the pharmaceutical market \citep{luhw20}, and in 2021, seven of the top ten best-selling drugs worldwide were biologics \citep{ur22} (see Fig. \ref{fig_best_selling_21}). According to the definition given by \cite{fda18}, biologics include vaccines, monoclonal antibodies, blood and blood components, allergenics, somatic cells, tissues, gene therapy, and therapeutic proteins. Unlike small molecule and chemically synthesized drugs, biologics are complicated, large mixtures of sugars, proteins or nucleic acids, and other substances which are not easy to identify and exactly characterise \citep{pehe20}. Most of the biologics are produced by biotechnology in a living system such as microorganisms, plant, animal, or human cells, in which mammalian cells like the Chinese hamster ovary (CHO) cells, mouse myeloma (NS0), baby hamster kidney (BHK), human embryo kidney (HEK-293) and human retinal cells are typically used \citep{wu04}. In recent years, the market for biologics has explosively grown with a percentage of new biological products approved by FDA every year since 2014 for treating various human diseases including cancers, autoimmune, metabolic and infectious diseases, always exceeding 20\% of the total number of new approved drugs \citep{deal22} (see Fig. \ref{fig_products_approved} for more details). 

Among biological products, mAbs emerge as the leading product in the rapidly growing market of high-valued biologics \citep{pabu19}. As reported by \cite{gr22}, the global therapeutic monoclonal antibody market was valued at about US\$185.5 billion in 2021 and is expected to achieve a compound annual growth rate of 11.30\% from 2022 to 2030 to reach a revenue of around \$494 billion by the end of 2030. To achieve this target, it is necessary to employ more cost effective solutions and smart manufacturing while simultaneously not sacrificing the process robustness and product quality when aiming for higher process productivity, more affordable end-products and shorter production times \citep{xe15}.

The manufacturing process of mAbs and therapeutic proteins is complex with many interconnected elements. It begins with the culture of genetically modified cells in small- and medium-sized bioreactors and expanding them to the large production bioreactors within upstream processes. After that, various downstream purification steps are performed to separate the product of interest from cell-derived and process-related impurities \citep{pako20}. Before the Quality by Design (QbD) has been introduced by FDA as a standard technique \citep{rawi09}, the assurance of final product quality in the conventional manufacturing of mAbs and therapeutic proteins has usually relied on costly and time-consuming quality by test methods \citep{zhma17}. In the quality by test approaches, the root causes of failure and/or significant levels of uncertainty within bioprocesses are often not well understood. Hence, it is frequently time-consuming, costly, and potentially adversely affecting product safety when the manufacturing procedures have to be repeated until the root causes of failure are detected and resolved. In contrast to the quality by test, QbD methods are based on the holistic understanding of products and how materials and critical process parameters (CPPs) influence the critical quality attributes (CQAs) and yield of the final products via different unit operations \citep{zhma17}.

In the QbD framework, the product specifications, also known as the quality target product profiles (QTPPs), are placed at the heart of the design strategies of experiments to guarantee the quality and consistency of the final products \citep{pako20}. This leads to the crucial roles of in-process analytical measurements and process analytical technology (PAT). Moreover, it is desired to understand the relationship between process conditions and product quality. These characteristics provide opportunities for the use of computational models as promising alternatives to experimental approaches to optimise and control the process behaviors more systematically \citep{smju20}, to extract useful knowledge from acquired experimental data and to build predictive models for in-silico experiments. As presented in \cite{pako20}, there are three main challenging areas in the implementation of QbD principles in biomanufacturing: (i) product heterogeneity, (ii) various limitations of measurements for process monitoring, and (iii) the lack of online process control strategies. These challenges can be mitigated by taking advantage of knowledge-driven and data-driven approaches for (real-time) bioprocess monitoring and control based on the collected bioprocess data. 

By monitoring the relationships among variables and extracting valuable knowledge from bioprocess data, machine learning (ML) models can acquire novel insights into the interdependence between CPPs and CQAs in biopharmaceutical process development and manufacturing. These insights have the potential to assist in the development of more effective process control strategies. Apart from real-time monitoring and control of upstream and downstream bioprocesses, supervised ML algorithms can be applied to discovery and design of biopharmaceuticals such as mAbs selection, media screening, design of feeding strategies, and selection of substrate for purification. ML algorithms can also be used in product formulation and stability steps. As a result, the deployment of ML methods for bioprocess development and manufacturing can contribute to reducing the time and production cost of bioproducts and, in turn, lowering the price of cell and protein therapy \citep{puda22, rasi15a}. Over the past few years, there has been observed a significant increase in the use of multivariate data analysis (MVDA), which is often considered to be a sub-field of ML focusing on a group of statistical approaches, as a PAT tool for cell cultivation \citep{baal21, rasi15a} to explore the available data from pilot or manufacturing plants aiming to find the correlation between process variables and product quality attributes, enhance the bioprocess comprehension, and online monitoring and control of unit operations. As summarised in \cite{medi14}, popular applications of MVDA consist of the analysis of spectroscopic measurements and data profiles of unit operations like cell culture and chromatography, quantitative evaluation of process comparability, root cause analysis of failures, and raw material selection. MVDA methods have been recommended by regulators as a critical tool for QbD and PAT frameworks to improve understanding of bioprocesses and to increase the levels of process monitoring and control and, in turn,increasing the probability of achieving the desired final product quality \citep{medi14}.

As shown in \cite{baal21}, it should not be surprising if the use of ML algorithms within biopharmaceuticals would grow explosively in coming years to take benefits from explored insights and the power of predictive models to optimise, monitor, and control bioprocess development and manufacturing operations. The main trends can focus on the use of advanced data analytics, the integration of soft sensors and PAT approaches to formulate advanced control strategies \citep{magu14}, and the development of hybrid data-driven and knowledge-driven models associated with integrated designs of production elements such as process, product, and cells \citep{basu20, wamy22}. Moreover, the biopharma field could witness a rapid shift to digital transformation with the increased adoption of queryable and structured centralised repositories such as a data lake or data warehouse for recorded data \citep{stbo19}. This digitisation would allow the application of ML models to all data gathered from multiple sites across various scales and unit operations of biomanufacturing plants thanks to data consolidation and accessibility \citep{baal21}. This creates an important premise for the fully automated data-driven biopharmaceutical manufacturing facilities equipped with knowledge of physicochemical properties of substances, bioprocesses, and products in near future.

This paper aims to provide a comprehensive review of the existing applications of ML to various stages of process development and manufacturing for mAbs and therapeutic proteins, identify challenges regarding bioprocesses themselves and process data, then proposing potential research directions in this area. We do not aim to provide statistics of the numbers of algorithms used for specific problem domains year over year. The interested readers can found more information for this theme in recent reviews \citep{baal21, gugl18, mosa21, phma22}. Instead, existing trends regarding how ML algorithms are used in the process development and manufacturing of biological products across different unit operations are reviewed in detail. This paper is an expansion of the recent review paper \citep{puda22}, which covers only typical applications. On the contrary, this paper presents a much larger number of applications of ML to biopharmaceuticals and identifies numerous potential directions for the applications of advanced ML methods to address existing challenges within bioprocess data aiming to the development of digital twin systems for biopharma 4.0.

The rest of this paper is organized as follows: Section \ref{biopharma_process} summarises some background of a typical biopharmaceutical process for manufacturing mAbs and therapeutic proteins. Next, Section \ref{challenges_biopharma} presents challenges regarding bioprocesses and process data characteristics, which can cause difficulties in building ML models. After that, Section \ref{section_early_stage} reviews popular applications of ML in early stages of bioproduct development, while Section \ref{upstream} covers different applications of ML models in upstream processing processes. Section \ref{downstream} is dedicated to presenting typical applications of ML in the downstream purification process. The use of ML algorithms to deal with problems in product formulation and stability is presented in Section \ref{production_stab}. Section \ref{open_datasource} collects valuable open data sources to support ML research in biopharmaceuticals, while Section \ref{potential_direction} discusses potential research directions of ML in the construction and development of Biopharma 4.0. Concluding remarks of this paper are described in Section \ref{conclu}.

\section{A typical biopharmaceutical process for production of monoclonal antibodies and therapeutic proteins}\label{biopharma_process}

\begin{figure}[!ht]
    \centering
    \includegraphics[width=1\textwidth]{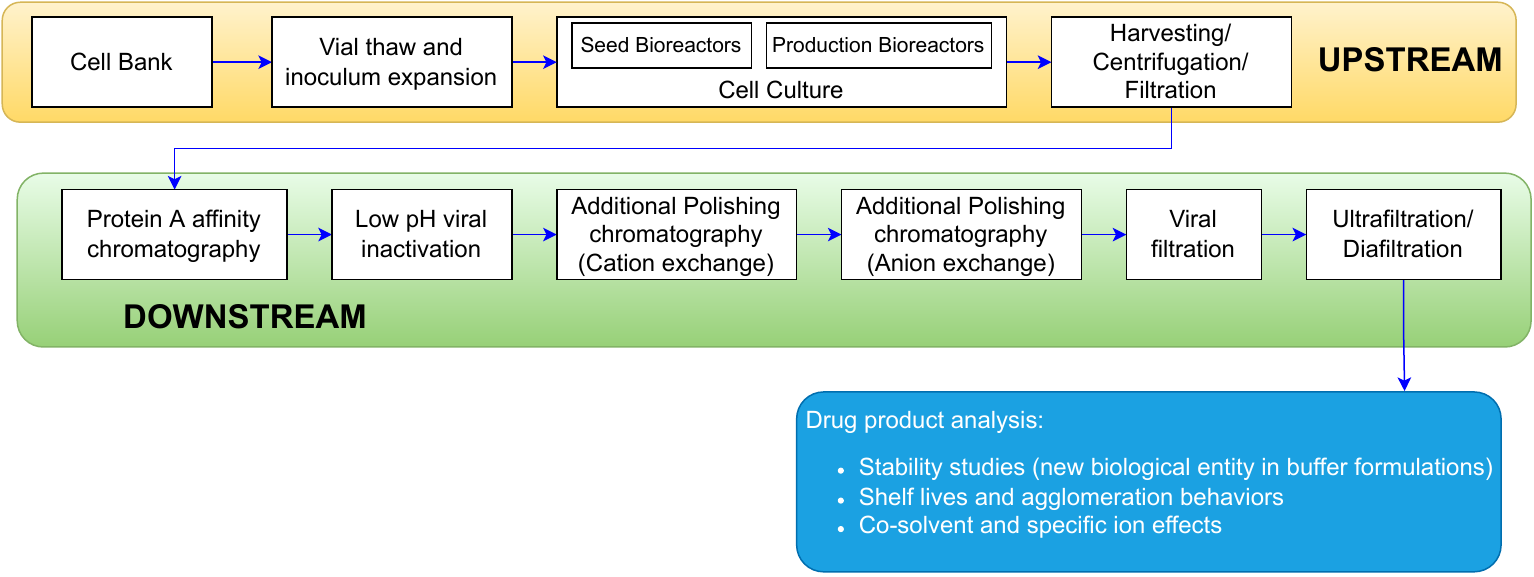}
    \caption{Production process of mAbs and therapeutic proteins (adapted from \citep{pabu19, smju20, shth10, hose18, basu20})}
    \label{bioprocess_mabs}
\end{figure}

The standard manufacturing and purification process of mAbs and therapeutic proteins includes two main stages: upstream (USP) and downstream processing (DSP) in combination with the final product formulation and stability phase. The upstream process comprises the cultivation of cells in a series of bioreactors and the production of the interested product, while the downstream process includes a sequence of separation/purification steps to capture the product of interest and eliminate various process, host cell, and product related impurities with minimal yield loss \citep{shth10, hose18, pabu19, basu20, pako20}. Currently, these operations are performed in fed-batch cultivation systems and batch separation processes \citep{xe15}, and sometimes continuous cultivation systems also use these procedures. Fig. \ref{bioprocess_mabs} represents a flow diagram for a typical manufacturing process platform of mAbs and recombinant proteins.

The upstream process begins with a vial thaw followed by the inoculum growth of cells from a cell bank and suitable media, which is performed in shake flasks or spinner flasks. After that, the upstream process formally starts with the cell culture in seed bioreactors with the progressive increase in size and/or volume. The cells are then transferred to the production bioreactor for protein expression in the medium \citep{shth10, pabu19}. Finally, a typical harvest procedure is conducted to eliminate cells and cell debris by centrifugation followed by depth and membrane filtration prior to be transferred to purification steps in the downstream \citep{shka08}. Fed-batch approach is usually used for the upstream process, where small volumes of feed nutrients are added to the bioreactors during the cell growth stages. Apart from the monitoring of nutrient and metabolite concentration, other factors of bioreactors in the upstream processing are also frequently controlled during the cell cultivation including dissolved oxygen (pO2), pH, temperature and mass transfer of oxygen and CO2 \citep{shth10}.

The downstream process for the manufacturing of mAbs and therapeutic proteins starts with Protein A chromatographic capture. This type of affinity chromatography relies on the specific binding interactions between the Fc area of mAbs and the Protein A ligand. As indicated in \cite{shth10, pabu19}, Protein A affinity chromatography step can lead to over 98\% purity because of high binding affinity and specificity of the Protein A to Fc-fusion proteins. This specificity enables impurities such as host cell proteins, DNA and other contaminants to pass through while the products of interest binds to the stationary phase. The products are then eluted from Protein A affinity adsorbents under low pH conditions for virus inactivation and reduction of the affinity of Protein A to the Fc regions. Next, one or multiple polishing chromatographic steps are conducted to further remove process and product related impurities, which cannot be removed by the Protein A affinity chromatography such as aggregates and product variants because of their chemical similarity with the derived protein as well as a byproduct related to the process like leached protein A. Cation exchange (CEX) chromatography and anion exchange (AEX) chromatography are the most commonly used polishing approaches \citep{shth10}, which may be used individually or in combination. After polishing steps, a viral filtration step is deployed for virus removal. The final step in the downstream process is typically an ultrafiltration/diafiltration (UF/DF) to decrease storage volumes and pass the product through the buffer for the formulation of the drug substance.

The last stage of the manufacturing process of mAbs and related proteins focuses on optimising and analysing the therapeutic products. Main steps of this stage include the studies regarding the design of product formulations assuring long shelf-lives and high stability of the therapeutic protein in the buffer solution in line with the desired delivery method \citep{smju20}.

Within the QbD scheme, each biopharmaceutical manufacturing process has to comply with predefined standards to ensure the QTPPs of the final product. In particular, the QbD method introduces a set of CQAs which present the suitable limit, range, or distribution of  physical, chemical, biological, or microbiological properties or characteristics of in-process and finished products \citep{mi13}. As claimed in \cite{eobr12}, these CQAs are frequently chosen based on risk-based analyses, where potential or known product quality attributes are assessed for their potential effects on patient safety and product efficacy. Quality attributes of recombinant therapeutic proteins can be divided into three groups, i.e., product-related impurities and substances, process-related impurities, and contaminants \citep{eobr12, pabu19}. These commonly used attributes are presented in Table \ref{cqa_table}.

\begin{table}[!ht]
\footnotesize{
    \centering
    \caption{Critical quality attributes are commonly used to assess the final recombinant therapeutic proteins (adapted from \cite{eobr12, pabu19, ch19})}
    \label{cqa_table}
    \begin{tabular}{L{5.5cm}L{3.5cm}L{6.5cm}}
    \toprule
      Product-related impurities and substances & Process-related impurities & Contaminants \\
    \midrule
    Aggregation & Residual host cell proteins (HCPs) & Adventitious agents (e.g. potential virues, bacteria, mycoplasma, bioburden, and fungi) \\
    Fragmentation & Residual DNA & Endotoxins \\
    N/C-terminal modifications & Raw material-created impurities (e.g. leached protein A, cell culture media, and buffers) & \\
    Oxidation & Leachables (from product contact materials) & \\
    Deamidation/Isomerization & & \\
    N-linked Glycosylation & & \\
    O-linked Glycosylation & & \\
    Glycation & & \\
    Conformation & & \\
    Disulfide bond modifications/free thiols & & \\
    GlcNAc, N-acetylglucosamine & & \\
    \bottomrule
    \end{tabular}
    }
\end{table}

This section only summarises the main steps and their functionality within a typical manufacturing platform of mAbs and recombinant therapeutic proteins. The review of impacting factors and the way of enhancing productivity and modifying product quality along with development strategies can be found in \cite{rawe18} for the upstream and in \cite{bahu17, raka14} for the downstream processes. According to \cite{basu20}, the optimisation activities of the upstream cultivation processes are usually conducted aiming to raise product quantity represented by titer and productivity, while the optimisation of the downstream purification processes is performed to obtain the desired product profile.

The manufacturing process of the mAbs and theurapeutic proteins is associated with tight regulations defining the purity and composition of the final products and the limitation of product and process-related impurities \citep{pabu19}. Therefore, rigorous process monitoring becomes critical for the manufacturing process. In that regard, control strategies for tracking the real-time values of impurities and secondary products may be of great potential and high demand. These are the opportunities for the deployment of data-driven modelling and optimisation for the whole biopharmaceutical manufacturing process from the discovery and design stages to upstream, downstream, and product formulation and stability. This paper will review key areas where ML algorithms can be used for these purposes. However, first we will discuss challenges associated with processes and bioprocess data.

\section{Challenges of biopharmaceutical processes with respect to ML solutions}\label{challenges_biopharma}
\subsection{Challenges associated with characteristics of biopharmaceutical processes}
The unique aspects and specificities of the biopharmaceutical industry give rise to various challenges related to data management and analysis, as well as the comprehension of complete bioprocesses. According to \cite{stbo19}, these obstacles can be tackled by the deployment of digital transformation solutions and the use of smart algorithms for monitoring, controlling, and optimisation of manufacturing processes. These current obstacles and potential solutions are presented in Table \ref{obstacle_sol}. In this section, we explore the distinctive features of the bioprocess life cycle that may present difficulties and challenges when it comes to creating and integrating digital solutions for the development and production of mAbs and therapeutic proteins.

\begin{table}[!ht]
\footnotesize{
    \centering
    \caption{Current challenges of biopharmaceutical manufacturing processes and potential digital solutions for addressing them (taken from \cite{stbo19})}
    \label{obstacle_sol}
    \begin{tabular}{L{8cm}L{8cm}}
    \toprule
      \textbf{Current challenge} & \textbf{Potential digital solution} \\
    \midrule
    Development and approval of new biologics is costly and the time from production to clinic is too long & Increased use of digital tools for experimental design, data-driven decision-making, and bioprocess optimisation\\ \midrule
    Bioprocesses are not transparent & Using predictive soft sensors to indirectly estimate difficult or impossible-to-measure variables\\ \midrule
    Routine manufacturing processes are not cost-effective, operate sub-optimally, or lead to failed batches & Applying optimisation approaches to optimise performance of already established processes; implement enhancements for product life-cycle management\\ \midrule
    Impact of multidimensional changes in process parameters on CQAs is usually unknown & Use process modelling approaches to achieve data-driven estimates for the changes of process parameters in consideration of whole process value chain\\ \midrule
    Shortage of highly skilled workers in statistical analysis and machine learning & Deploy automated algorithms to replace the manual data analyses\\ \midrule
    It is difficult for humans to analyse and understand high-dimensional bioprocess datasets & Deploy data-driven methods for data interpretation\\
    \bottomrule
    \end{tabular}
    }
\end{table}

\subsubsection{Challenges in achieving an effective manufacturing process}
In the manufacturing process of therapeutic proteins, cell culture plays a vital role and a high-performing media formulation is one of the key factors to obtain high cell densities, high productivity levels and quality protein products \citep{bige22}. Therefore, enhancement and optimisation of cell culture media formulations becomes one of the key challenges to achieve the high performance and quality of the final products across multiple cell lines \citep{riwu18}. In addition to cell culture media optimisation, the effective nutrient usage is also an important factor to achieve high performing cell cultures. As a result, monitoring and optimisation of process inputs represent a major theme to reach more efficient bioprocess development and manufacturing \citep{va21}. Therefore, PAT tools are frequently deployed to measure and quantify in-process parameters. As an example for the use of PAT tools, new sensors have started to be frequently used for real-time process monitoring and control in addition to routine off-line analytics \citep{du21}. PAT data can also be employed by ML algorithms for the analysis and optimisation of production and improving the product quality. Together with the development of automation and intelligent devices, smart learning algorithms have enormous potential to be integrated into bioprocess manufacturing systems. This integration enables enhanced monitoring and adaptation of unit operations to accommodate dynamic changes in cell cultures and the purification of biological products. The ultimate goal is to increase yields and improve the quality of the final products. Although automated and adaptive ML solutions have been widely used for many industry processes \citep{sabu16, baga16, baga15, kagr11, kaga11}, there is still no study in the literature deploying automated and adaptive ML methods for real-time monitoring and control of biopharmaceutical processes. This is due to the high complexity and rigorous regulations regarding assessment of process and product related measurements besides the challenges in bioprocess data and a process comprehension. With the recent innovation in automated and autonomous machine learning (Auto/AutonoML) \citep{khke22, doke21, bafa21, kemu20}, however, the integration of Auto/AutonoML into the whole manufacturing processes from upstream, to downstream and formulation stages is more likely to become a feasible solution to address challenges in real-time monitoring, control, and improvement of product quality of bioprocessing.

\subsubsection{Domination of batch based bioprocesses}
Although continuous processes are very common in many process industries \citep{kaga09}, biopharmaceutical manufacturing relies mainly on the batch processes, in which each process can be split into serially performed unit operations \citep{shth10, pabu19, stbo19, smju20}. Given a process, the numbers of unit operations, splitting and pooling steps depend on the product. The variation in one of the CPPs of one unit operation can result in the change in process conditions and CQAs of the products in subsequent steps \citep{stbo19}. This poses challenges and opportunities for the use of ML solutions for real-time monitoring of CPPs based on the feedback data of offline analytics and online PAT data. 

In contrast to continuous processes, batch processing has to be fully completed prior to running another batch. This characteristic offers several benefits, including the ability to control mutations in a cell line, which can impact process efficiency and product quality, as well as providing flexibility in the production schedule. Notwithstanding, batch bioprocessing shows downsides in the production time and scaling up abilities \citep{sc19}. The quality and uniformity of each batch needs to be verified to improve the performance of future batch runs which leads to an increase in the production time. In addition, the products are harvested only once from each batch, so batch manufacturing is a labor-intensive process, ideally, to be performed at a large scale. To meet sufficient number of cells for bioprocessing, the batch will be extended in successively larger bioreactors. The cost of the equipment for the growth of successively larger cell cultures is significant. Hence, fitting a new batch process into existing equipment is considered as one of the biggest challenges that bioprocess manufacturers have to face \citep{ma20b}. As a result, the shift to continuous operation modes can lead to the significant reduction in operating time and machinery cost for scaling up, which in turn also reduces the problems related to mathematical modelling concerned with the effects that scaling up can produce \citep{ma20b}. The transition to continuous manufacturing is a potential opportunity for the deployment of automated and adaptive ML solutions for the real-time monitoring and dynamic optimisation of control strategies for the whole manufacturing process. It is because the process control is crucial for achieving the steady-state of continuous processing and managing the potential dynamic effects and drift associated with biological processes and the product quality concerns over time. All of the above issues can be, in principle, addressed by the use of automated and adaptive ML algorithms.

\subsubsection{Diversity in interoperability of unit operations}
As mentioned above, the manufacturing process of biologics includes several serial process unit operations (PUOs), also known as the bioprocess trains, each having a specific impact on the final CQAs and overall productivity. The interaction among the parameters of various unit operations can lead to the generation of many impacts on the CQAs of the final products similarly to a contribution of parameters of an individual unit operation \citep{jise17}. Hence, the CQA profiles are desired to be known prior to the execution of the next PUO for the effective operation of interconnected bioprocess trains. To achieve this goal, online CQAs measurements should be specified at every PUO if possible \citep{jise17}. As presented in \cite{eobr12}, the product CQAs are defined based mainly on the risk assessment of the process development within a QbD framework, so the implementation of sensory systems across multiple PUOs can play a crucial role for this purpose. As a result, the integration of the monitoring, data analysis and finally modeling of CQAs across the various PUOs of the bioproduct manufacturing process shows a crucial importance, especially in a continuous operation mode \citep{gust19}.

One of the characteristics of an industrial production of bioproducts is the large-scale operation. However, in practice, large-scale heterogeneity usually causes higher variability of CQAs \citep{gust19, brfr17}. Though several studies in the literature have presented the effects of CPPs on CQAs, few studies describe the correlation between process heterogeneity and the variability of CQAs and the level of metabolic reactions \citep{gust19}. The main reason for this fact is the scarcity of real-time monitoring tools for product CQAs during larger scale operations, which limits the comprehensive understanding of the dynamic operational environment. Therefore, it would be very desirable to take full advantage of automation and machine learning solutions for monitoring and analysis of all related PUOs and generation of detailed understanding of bioprocesses as well as dynamic changes during the operational time to improve the yield and quality of final products, especially in industrial large scale operations. Nevertheless, the large amount of collected data could pose new challenges for information technology infrastructure, data storage and governance as well as effective raw data mining algorithms.

\subsubsection{Challenges in time to market}
As stated in \cite{de19}, one of the main challenges in the development and manufacturing of biologics is the constant pressure in delivering novel and high quality therapeutics to the market faster and more cost effectively, while still meeting high standards of regulatory compliance. The sooner the process runs and regulatory approvals for the products are completed, the sooner those essential therapeutics become available for patients. However, due to the complexities of biologics and long production cycles, it is tough for biopharma companies to balance the specificity, quality, cost, and speed. Any small accident or production failure can result in destructive ripple effects.

To accelerate the manufacturing life-cycles, biopharma organisations have invested heavily in innovative solutions such as the high throughput process development, the expanding use of single-use technology \citep{du21}, process analytics, cross-scale predictive modelling, and continuous processing \citep{de19}. These changes pose new challenges but also open opportunities for the implementation of automation and ML solutions for real-time data analytics, continuous monitoring and control of the whole system from the upstream to the downstream processes.

\subsubsection{Constant pressure to lower costs}
No business can avoid the cost pressures including the pharmaceutical industries. Process intensification may be considered as an efficient approach to lower the cost of goods through enhanced process efficiency without sacrificing productivity \citep{bofl22}. Bioprocess intensification is defined as a rise in a bioproduct output related to cell concentration, time, bioreactor volume or cost \citep{wh20}. The implementation of process intensification may reduce up-front capital investment, optimise outputs, streamline labor expenses, and reduce other overhead costs like water, energy, waste disposal, and cleanroom maintenance \citep{bofl22}. 

There are some similarities between the intensification and optimisation, in which both methods aim to enhance the performance to increase productivity \citep{ke18}. However, the intensification is associated with more step changes in equipment and/or process design leading to not only a significant increase in productivity but also a considerable enhancements in other environmental and economic metrics such as energy consumption and carbon footprint \citep{bofl22, boha13}. One of the fundamental strategies in terms of the philosophy behind the process intensification is the shift from batch to continuous processes \citep{lode21}. Continuous bioprocessing is attracting more attention because of its advantages in terms of constant nutrient conditions, elimination of by-products, a capability of formulating high density cultures and no lag phase once the system is operating \citep{bofl22}. 

As shown in \cite{lode21}, ML is an important tool for the innovation of the process intensification with the ability to extract insights, pattern recognition and predictions in the process industry. These abilities can assist in the sustainability design, system integration, advanced process and quality control, intensification, and data-driven process modeling. However, the applications of ML for the process intensification face several following challenges:

\begin{itemize}
    \item \textit{Real-time operation}. This challenge requires the ML models to capture, interpret, and analyse the large amount of data instantly. This feature becomes prominently important in the context of continuous processes to handle constantly streaming data. The real-time operation ability could intensify closed-loop systems to result in automated and self-driven assemblies \citep{sale21}.
    \item \textit{Interoperability}. Bioprocess manufacturing operations require the exchange and communication of information among multiple unit operations.  This poses challenges in data synchronisation between various interconnected equipment and devices through technology infrastructure to allow the coordinated actions among devices, sensors, software, process models, and humans \citep{sabh21}.
    \item \textit{Information transparency}. This requirement poses the challenges in readiness of information from the entire process and external knowledge. It requires the continuous process and visualisation of the large amount of data in an interpretable approach which could be utilised immediately for the adaptation of the production processes to meet the market demands and supporting the process enhancements and intensification strategies \citep{side21}.
    \item \textit{Distributed decision making}. This feature requires the subsystems or unit operations to be capable of making localised data-driven decisions yet also having the ability to coordinate and cooperate among constituent subsystems \citep{daal19}. This results in the challenges in computational efficiency, flexibility, coordinated and parallelised communication of data-driven decision making algorithms within each unit operation.
\end{itemize}

\subsection{Challenges stemming from various characteristics of bioprocess data}
As stated in \cite{ma20b}, some of the greatest challenges in bioprocessing today are the democratisation and the implementation of potential digital solutions, especially focusing on ML models, for knowledge extraction and supporting informed decisions based on the data sources derived from cell-line development to manufacturing and product formulation. However, due to the complexities of biologics, process manufacturing and mechanisms underlying the biological systems, bioprocess data present many challenges for the ML modelling and analysis. This section presents several typical characteristics of bioprocess data including the diversity of data sources, heterogeneity, high dimensionality, data integrity, big data, small training data with infrequent feedback, and the lack of open datasets. These characteristics can result in many challenges in the deployment of ML solutions for biopharmaceutical process development and manufacturing.

\subsubsection{Diversity of data sources}
According to \cite{dubo22}, the toughest challenge for automated model building which needs to be overcome is the diversity, distribution, and multi-structures of data sources. \cite{stbo19} summarised the most critical data sources during the bioprocess life cycle (see Table \ref{data_source}). As we can be see there are often many different data sources used in the biopharmaceutical development and manufacturing life cycle. Several data sources such as file-based exports are easy to set up but they represent several disadvantages for analysis and model building due to the inconsistency in formats and much effort needed to restructure and clean such data. Some of the data sources from data warehouses and historians contain structured and standardised data which can be useful and easy to access for the development of ML models. However, these structured databases require significant effort to set up. Although the diversity of data sources, as presented in Table \ref{data_source}, can present many challenges when building effective machine learning models, optimal solutions for the practical problem typically require the usage of multiple data sources (e.g., an Open Platform Communications (OPC) connector exporting to a data warehouse).

\begin{center}
\scriptsize{
\begin{longtable}[t]{L{2.4cm}L{3cm}L{3.5cm}L{3cm}L{2cm}L{1.3cm}}
    \caption{The diversity of data sources within a bioprocess life cycle (data are taken from \cite{stbo19})} \label{data_source} \\
    \toprule
       \textbf{Source} & \textbf{Examples} & \textbf{Pros} & \textbf{Cons} & \textbf{Rapid setup and low IT/OT costs} & \textbf{Usability for ML}  \\
      \midrule
      Data warehouses & Microsoft SQL Server, Oracle Database, SAS Data Management, Amazon Redshift & Easy and fast access for analytical purposes; store data from various departments and unit operations & Considerable effort to set up the systems & * & ***** \\ \midrule
      Data historians & Wonderware, OSI Pi & Able to store large amounts of mostly time-series process data; often offer various APIs to acess to data & Considerable effort to set up and maintain the systems & * & ***** \\ \midrule
      Tabular files & .csv, .txt files, Excel, Ambr exports automatically or manually generated & Easy to set up and use; portable file format without requiring complex IT environment & Too many various variants and inconsistencies due to manual file changes result in vast effort for data collection and cleaning when analysing a large number of files & ***** & *** \\ \midrule
      LIMS systems & Starlims, McKesson Lab, LabWare, LabVantage & Standardized data source for analysis results from laboratory & Need additional meta data to interconnect these data with process data from other systems & * & *** \\ \midrule
      SCADA/MES systems & Sartorius MFCS, Infors Eve, Applikon BioXpert & Structured data formats and easy to access using standardised interfaces; many available process data & Require set up and availability of interface; metadata to the process itself are usually missing & *** & *** \\ \midrule
      Electronic batch record & Emerson Syncade, ampleLogic eMBR & Extremely valuable data but requiring less effort for data collection and cleaning than paper-based batch record & Huge amount of data; still need much manual work and comprehension & * & *** \\ \midrule
      Direct interface to analytical devices or sensors & OPC, serial ports, field bus & Simple set-up; real-time direct access to data; most direct manner to implement smart sensors & Usually missing metadata; only simple algorithms can be deployed without any further additional data sources & *** & ** \\ \midrule
      Distributed control systems & Emerson Delta-V, Siemens Simatic PCS7 & Real-time connection to connected systems & Often difficult to access (historical) data due to its main purpose is automation, not data delivery & ** & ** \\ \midrule
      Data lakes & Distributed file systems (e.g., Apache Hadoop, Azure Data Lake, Amazon S3) & Able to store structured and unstructured data easily; no data cleansing and preparation necessary to store data & Needs data preparation for unstructured data prior to analysis & *** & ** \\ \midrule
      Hand-written laboratory results & Paper-based laboratory notebook & No IT require & Not able to perform automated data analysis; need to transfer data into another system first & ***** & * \\ \midrule
      Paper-based batch records & - & Extremely valuable data that cannot be found anywhere else (e.g., holding times, exceptions, or out-off-specification events) & Extremely significant effort for data collection and cleaning & *** & * \\ \midrule
      Others & .pdf,.docx files, images & Large amounts of available data in these formats & Data are usually unstructured and need much manual work to make them usable for ML tools & **** & ** \\
      \bottomrule
    \multicolumn{6}{l}{\scriptsize *IT: Informational Technology,  OT: Operational Technology} \\
\end{longtable}
}
\end{center}

\subsubsection{Big raw data issues}
Pharmaceutical organisations usually derive and store a large amounts of data from many different data sources for the business and research purposes. As indicated in \cite{tofe18}, the estimation of the data under storage of one large pharmaceutical company for all purposes are around 6 petabyte of which 2 petabyte are generated from the R\&D departments. The biopharmaceutical production sectors also create a lot of data. Hundreds of process parameters including material inputs, process outputs, control actions and physiochemical parameters from both production scales and the cell expansion in the inoculum trains are regularly collected and stored electronically \citep{chhu08, chle10}. With advances in real-time PAT tools and the increased use of high throughput and automated bioreactor systems, the development and manufacturing processes of mAbs and therapeutic proteins are generating more and more bioprocess data \citep{phma22}. Consequently, the biopharmaceutical industry could face the challenges in data synchronisation and moving from a data-heavy environment to the one where data are converted to information resulting in useful knowledge. Huge amounts of raw data also poses many obstacles for ML algorithms in data preparation and engineering, model building, and knowledge extraction because analytics tools need large amounts of good quality data rather than huge amounts of raw data with noise and redundancies. To overcome this challenge, a significant amount of resources and effort should be allocated to implementing solutions for data preprocessing and synchronisation prior to the development of ML models.

\subsubsection{Data heterogeneity}
Data heterogeneity is identified as the most prominent problem in the process development and manufacturing of mAbs and therapeutic proteins \citep{phma22, chhu08}. Two sources resulting in bioprocess data heterogeneity are time scale and type  \citep{chhu08, chle10}.

Bioprocess parameters can be classified into online and offline types depending on the frequency of measurements. While a vast majority of process parameters are measured online such as pH, temperature, dissolved oxygen and CO2 concentration, PAT data, and gas flow, a few critical parameters, such as viable cell density and product concentrations and several metabolite and nutrients concentration measurements, are acquired off-line \citep{chhu08}. Many online measurements are collected in real-time, while measuring the off-line parameters is performed periodically within a production cycle. In addition, several key parameters such as product concentration and quality indices are only available at specific time points during the operational process due to the time and expense associated with their acquisition. These unique characteristics of bioprocess data result in high level of heterogeneity with regard to the measurement frequency. Many ML algorithms are not designed to work well on heterogeneous data collected over different sampling time intervals \citep{hose18}. It is required to modify the inner mechanisms of learning algorithms \citep{chle10, leka12, seva15} or use techniques to preprocess time heterogeneity such as dynamic time warping \citep{uewi02, vaga16}.

In addition to different measurement frequencies, the data type is also a source contributing to the heterogeneity of the bioprocess data. While many process parameters such as nutrient and metabolite concentration, cell and product concentration, pH, and temperature are continuous, other parameters are discrete or even binary like valve settings for nutrient feeding and gas sparging as ON/OFF state \citep{chhu08}. The data type heterogeneity can be tackled in data preprocessing stages prior to model building by encoding techniques, or using learning models which have the ability to learn from mixed-attribute data such as CART decision trees \citep{brfr17b} or fuzzy-neural networks \citep{khga23}.

\subsubsection{Small training data issues}
While for established or legacy biologics, many pharmaceutical companies have collected large amounts of historical data of experiments spanning years or decades, process data for novel bioproducts are limited, sometimes amounting to as little as only one or two runs for a new product at the manufacturing sites \citep{tuga18, baal21}. This characteristic results in a Low-N problem, in which the number of training samples is much smaller than the number of data dimensions. Furthermore, in practice, products are regularly moved to various production sites to accommodate other products and their life cycles. This operation type also contributes to the Low-N problem as there is a limited number of historical experiments available at new manufacturing facilities. With the increased use of PAT tools within bioprocess steps, large amounts of diverse measurements and information can be collected in real-time while there are only a small number of experiments that can be conducted \citep{baal21}. This typically makes it quite challenging for ML algorithms.

The limited number of observations together with high input dimensionality results in sparse sample spaces. Learning models which are built using such training sets have a much greater risk of being overfitted \citep{vefr05} leading to their potential poor predictive performance (i.e. something that in the ML literature is referred to as an ML model's poor generalisation ability). A possible solution to overcome the challenges within small training sets is the generation of larger training data sets necessary for ML models using probabilistic algorithms such as Gaussian Processes to derive artificial data from small data sets with several assumptions \citep{tuga19, tuga18a}. Another solution includes the deployment of Digital Twins to yield unlimited simulated data sets, which can be used to build and assess the ML models while waiting for the availability of experimental data \citep{godu19}. Despite several existing solutions, the Low-N problem is still a longstanding challenge which has not been completely tackled when applying ML solutions to bioprocess data.

\subsubsection{High dimensionality}
Bioprocess data are high dimensional with routinely tens or even hundreds of different parameters related to the process and products being monitored over time to assure a high-quality process \citep{puda22}. Moreover, with the integration of PAT tools such as spectrometers into the bioprocess steps for online monitoring of process parameters, the number of dimensions of the obtained data can be thousands of variables as in the Raman and infrared spectroscopy data \citep{gava15}. This, in turn, results in the issue known in the ML literature as the curse of dimensionality. Learning from high dimensional feature spaces with a limited number of training samples is a huge challenge for ML algorithms.

To reduce the impact of high dimensionality on the performance of learning models, feature selection and dimensionality reduction methods need and are commonly employed to select a small subset of features that are significantly correlated to the process outcomes. For example, in \cite{chle10}, clustering of parameters was used to identify the representative parameter from each cluster and remove redundant parameters representing the same information. However, the process data usually show the temporal nature and time-dependency, so the feature selection approaches need to consider the sequence of events. The dynamic nature and time-series setting of process data is a considerable challenge for common feature selection methods because of the interdependence among parameter values across time \citep{fica16}. Statistical methods can be used in this context to assess the relative importance of temporal process variables in discriminating runs from various clusters \citep{chhu08}.

\subsubsection{Data integrity and quality}
Data integrity and quality are also some of the key factors which can affect the performance of ML models. For instance, missing values and incompleteness of datasets are the inevitable features of bioprocess data. There are several reasons for missing/incomplete information in bioprocess databases ranging from biomanufacturing settings such as a sensor breakdown during the operation, inconsistent sampling rates, the utilisation of incorrect format in data logging, malfunctions in data acquisition devices, errors in recording of real-time parameters to the change or update of instrumentation systems and a variety of human factors \citep{semo17, maga19, gase21}. 

The way of handling missing values accounts for one of the main challenges in mining manufacturing data of biologics because the missing value handling method has a strong influence on the performance of successive analyses of the bioprocess datasets \citep{maga19}. The high complexity of underlying mechanisms of the biological systems poses additional challenge to the efficient handling of missing values as the employed methods need to take into account the correlation of individual parameters and complex interactions among them when dealing with missing information \citep{deco19}. As presented in \cite{maga19}, the missing values of parameters in bioprocess data usually occur at random. Majority of these missing values are associated with parameters which are measured and fetched automatically. Therefore, there are three possible approaches to deal with such missing values \citep{lida12}, i.e., eliminating incomplete samples, imputing the missing data and using the learning algorithms and analysis methods not requiring a complete dataset such as the methods in \cite{ga02, foar17, khch21}. Discarding a large number of observations may introduce biases to analytic results, whereas the implementation of effective imputation methods, as claimed in \cite{maga19}, has not introduced any bias to the subsequent analyses of production datasets of mAbs.

In addition to missing information, the presence of outliers in bioprocess data, which can result from errors in recording and measurements, unknown data structure, or new phenomenon, can also negatively affect the performance of subsequent data analyses and ML models. Therefore, identifying and excluding outlier points in the dataset is also an important step prior to ML model building. The methods based on the standard deviation, statistical metrics and tests are usually used for outlier detection in bioprocess data \citep{kico07}.

\subsubsection{Lack of open datasets for reproducibility of ML solutions}
To assess the effectiveness of the ML solutions in the literature as well as performing comparative analyses between different algorithms, it is required to run algorithms on the same datasets. However, the reproducibility of ML solutions for bioprocess data is a significant challenge due to the lack of publicly available benchmark datasets. Bioprocess data are expensive to obtain as conducting  biological experiments is time-consuming, labor-intensive and requires expensive materials \citep{dubo22}. In addition, the bioprocess data are associated with strategic products and business secrets of pharmaceutical companies, thus it is unlikely that they are publicly shared with the research community. These facts prevent the explosive growth of ML solutions for biopharmaceuticals in comparison to other fields such as natural language processing and computer vision. 

\section{The applications of ML in early stages of biologics development}\label{section_early_stage}
Multiple applications of ML algorithms to the discovery and development of small chemical molecules over the past few decades have resulted in some outstanding outcomes. Several recent surveys such as \cite{scwa20, vacl19, chsh19} have summarised the algorithms, remarkable achievements, and challenges in applying ML to the discovery of small molecule drugs. In contrast to the small chemical molecules, the utilisation of ML for the early discovery and development stages of biologics is still limited. As mentioned in \cite{nadi21}, there are a number of reasons for this trend including the complexity of biological products and the sparse, heterogeneous, and smaller datasets of biomolecules in the available data warehouses of biopharma organisations in comparison to small molecules. This section reviews the up-to-date applications of ML to the early discovery and development stages of biologics.

\subsection{Discovering prospective biologic candidates and assessing their computational developability} \label{discover_mabs}

Evaluation of potential biological candidates in terms of safety, immunogenicity risks, and chemistry, manufacturing and control challenges including conformational stability, aggregation, high viscosity, and chemical degradation sites at the early stages of discovery and development is critical for risk mitigation of costly delays and failures during later product development stages, especially for antibody-based therapeutics \citep{kupl18} and the development of biosimilar products \citep{zu13}. Therefore, computational tools have been increasingly employed to assist the discovery of biologic drug candidates. To effectively apply computational algorithms, especially ML models for this purpose, it is necessary to build a library containing sequence information and structures of biologics together with their chemistry, manufacturing and control (CMC) attributes including expression levels, cell lines, physicochemical stability, and analytical characterization results as well as physicochemical properties. Such library can be explored by ML based methods to discover the insights regarding how molecular sequences and structural properties form attributes such as concentration-dependent viscosity behaviours, chemical degradation, solubility, and aggregation \citep{kupl18}. This knowledge enables to enhance the candidate selection process. \cite{khcu22} provided a list of useful databases and datasets which can be used to train, validate, and evaluate biopharmaceutical informatics tools for the discovery and development of antibody-based biologic drug candidates.

In the discovery phase of biological products, the protein sequence information is usually available quite early in the libraries of prospective binders to the given product targets \citep{kupl18}. This information in the libraries frequently results from hybridoma, phage display technique, next generation sequencing technology or other similar activities \citep{glda15}, and it can be applied to specifying aggregation prone regions, immune epitopes, or potential physicochemical degradation liabilities in biologic candidates using computational and ML tools \citep{kupl18, puda22}. In addition, combining protein structural information based molecular modelling with the analysis of sequence information can provide valuable insights regarding molecular origin to develop optimisation strategies to reduce safety issues for biologic candidates. For instance, \cite{kumi12, kuth16} presented the identification of common molecular regions underpinning both aggregation and immunogenicity of biologics. \cite{kupl18} stated that it is feasible to predict critical quality attributes of a biologic candidate such as chemical degradation sites based on its amino acid sequence by computational modelling. The integration of molecular modelling and data analytics can lead to novel, accelerated, and cost-effective bioprocesses and products \citep{zu13}.

According to \cite{zu13}, product design, lead selection, and manufacturing process development contain significant risks for biopharma industry due to their impacts on final product quality, costs, biological activities, and safety. As a result, the introduction of a developability assessment to the lead selection and optimisation phase of biologic candidates can facilitate tackling crucial quality aspects regarding the manufacturability, safety and pharmacology of novel biologic candidates. The concept of developability is defined as the probability for the successful translation of a lead candidate to a biologic drug which ensures stability, manufacturability, safety, and specificity during production and formulation \citep{jako15}. Therefore, developability is a multi objective optimisation issue, where ML can be integrated with high-throughput experimental approaches to form a potential screening method aiming to enhance efficiency and determine optimal properties at early stages of the biologics discovery process \citep{nadi21}. Unlike the QbD method, developability focuses mainly on the product design and discovery stages rather than on the control of the manufacturing processes \citep{jako15}. The developability assessments at the early development stages of biologics bring many benefits including significant reduction of risks at later development pipelines, saving time and resources, providing a chance for re-engineering of the molecule to identify and alleviate sequence or structural liabilities. It also contributes to the selection of better prospective
molecule candidates representing similar efficacy and more favorable development ability \citep{khcu22}.

In practice, the deployment of experimental methods for the production of a large number of candidates and the biophysical characterisation assays to assess biophysical properties is costly and time consuming \citep{hewa19}. Therefore, the \textit{in silico} candidate screening approaches are usually preferred to identifying high value lead candidates and novel engineering targets aiming at accelerating the biotherapeutic development process \citep{shmo18}. However, these approaches require the availability of experimental datasets for biophysical properties of approved biological products. Several such databases have been presented in the literature such as \cite{jasu17}'s dataset and Thera-SAbDab \citep{rama20}. Using these datasets of biophysical properties of approved mAbs, ML models can be developed to make prediction of biophysical properties and antibody developability based on the information of amino acid sequences. \cite{hewa19} developed ML models trained on the experimental dataset \citep{jasu17} for all the
12 biophysical properties to predict the performance in terms of biophysical characterisation based on the amino acid sequence features. This is a crucial step for evaluating the developability of antibody-based biotherapeutics. \cite{chdo20} demonstrated the success of ML models in predicting antibody developability from amino acid sequence information based on physicochemical and learned embedding features, which can avoid the determination of antibody structures experimentally or computationally. The authors used a dataset of over 2400 antibodies \citep{dukr14} to extract features based on antibody sequence data. Then, six ML models including Gaussian Bayes, Logistic Regression, Support Vector Machines (SVM), Random Forest, Gradient Boosting, and Multilayer Perceptron were trained on the physicochemical features and embedding features extracted from sequence and amino acid data to predict the developability of antibodies. Out of the six ML models, the SVM showed the best performance in capturing antibody developability from the sequence data. \cite{gero19} built the artificial neural networks to predict critical biophysical properties of therapeutic mAbs including aggregation onset temperature, melting temperature of unfolding, and  protein-protein interactions in diluted solutions using only the amino acid compositions of mAbs. The predictive model could be used to select the mAb candidates with good physicochemical properties among thousands of mAbs sequences before expression in cells and purification for the next step of the developability assessment. 

The main requirements for developability assessments are to identify critical quality attributes at early stages and to evaluate the influence of detected molecule liabilities for the selection of the best candidates \citep{loba21}. Typical liabilities in terms of chemical and physical stability attributes consist of methionine (Met) and tryptophan (Trp) oxidation, asparagine (Asn) deamidation, and aspartic (Asp) isomerisation \citep{xuwa19}. Chemical and physical stability is a key issue in the development of protein therapeutics because of its effects on both efficacy and safety of the product \citep{jisu17}. Therefore, stability studies have an important role to play in initial developability assessment of therapeutic proteins and their formulations \citep{puda22}. Several ML algorithms have been used as \textit{in silico} predictors of specific liabilities, which show the capability of virtual screening of developable biologic drug candidates. For example, \cite{fa20, cawa19, johe18} demonstrated the use of ML algorithms for prediction of the stability changes in proteins based on single or multiple mutations. However, \cite{fa20} claimed that these ML models have not yet achieved a sufficient performance level for practical use which could have resulted from the limited amount of experimental data and informative features. In terms of prediction of the chemical stability from amino acid residues to eliminate or reduce the liabilities as early as possible in the selection stage of biologic candidates, \cite{jisu17} evaluated the effectiveness of six ML models including SVM, random forests (RF), naive Bayes, K-nearest neighbor, artificial neural networks (ANN) and partial least squares to predict Asn deamidation based on the available experimental and structural data of deamidated proteins. Meanwhile, \cite{dewa19} introduced the use of random forest models to predict if an Asn is liable for deamidation and determine the deamidation rate. The random forest model was trained on the sequence data of mAb peptides. Likewise, SVM, RF, and ANN were also used to predict Met oxidation sites \citep{alca17} and risk \citep{saho18} based on the features extracted from the sequence and structure data of proteins and antibodies. \cite{yaja17} demonstrated the use of RF regression to estimate solvent-accessible surface area of Met residues from the antibody sequence data to be used as an alternative metric for prediction of oxidation propensity. In addition to chemical stability, biophysical stability is also commonly used to identify the stability of proteins and mAbs. \cite{kiwo11} used the SVM model for the prediction of the thermal and pH stability of antibodies based only on their sequence data. 

In recent years, there has been much emphasis on the development of high concentration antibody formulations for low volume, subcutaneous delivery, and moving to appropriate and patient-centric dosing schemes and patient self-administration \citep{laga22, puda22, whxi17}. The developability properties of mAbs including low viscosity, low aggregation propensity, and high solubility are important for the development of novel high-concentration mAb-based therapeutics \citep{laga22, puda22}. Nevertheless, the assessment of such properties of mAbs at high concentrations in early stages of discovery and screening of candidates is difficult because of the limitation in the sequence and biophysical property data of molecules \citep{laga22}. Therefore, it is highly desired for the construction of predictive models which can estimate the developability of high concentration antibody formulations at early development stages. \cite{lafe21} demonstrated the use of Decision Trees trained on features extracted from 27 FDA-approved mAbs including net charges and high viscosity index (HVI) to predict antibody viscosity at a high concentration (150 mg/ml). \cite{laga22} extended the research by combining datasets of 27 approved commercial antibodies with 20 preclinical and clinical stage mAbs to train logistic regression, SVM, KNN, and Decision Trees for prediction of low and high viscosity of antibodies. In another study, \cite{lasw21} showed that the combination of coarse-grained models with hydrodynamic calculations and HVI parameters derived by ML models can predict viscosity of mAb candidates at various concentrations. For aggregation, there are several \textit{in silico} models using ML algorithms for predicting the protein aggregation rates, which is a key aspect in the development of biologic drug candidates because protein aggregation can affect safety, efficacy, and immunogenicity of protein-based therapeutics \citep{hecr04}. \cite{lafe21b} deployed the k-nearest neighbors and support vector regression models trained on structural-based features extracted from molecular dynamics simulations of 21 mAbs to predict the antibody aggregation rates at 150 mg/ml concentration. In later study, \cite{laga22} evaluated the performance of ML regression models such as linear regression, support vector regression, and k-nearest neighbors regression in predicting antibody aggregation rates for 20 preclinical and clinical stage mAbs using features extracted from  molecular dynamics simulations of the full-length antibody. \citep{obar15} developed tree-based ensemble models to predict the intrinsic aggregation propensity of mAbs based on physicochemical properties derived from amino acid sequence data. \citep{gero20} applied ANN models to prediction of real-time aggregation of proteins under 24 different real storage conditions based on accelerated stability studies and commonly used biophysical properties such as the first apparent temperature of unfolding and onset temperature of aggregation. The proposed method can provide better insights regarding the long-term thermal stability of proteins without having to provide the primary sequences and protein structural inputs. For \textit{in silico} sequence-based protein solubility predictors, gradient-boosted machines \citep{rama18}, deep neural networks in combination with a novel data augmentation algorithm based on generative adversarial nets \citep{hazh19}, and convolutional neural networks (CNNs) \citep{khra18} are the current state-of-the-art models. Moreover, ML models such as random forests \citep{yani16}, can be deployed to predict the effects of mutations such as individual amino acid substitutions on the solubility of the target proteins, which can be used as a cost-effective method for screening of soluble mutational variants of therapeutic proteins. In spite of high potentials of ML on the development of \textit{in silico} tools, it is noted that one common drawback of these frameworks in the early stages of biologic drug discovery is the input of these models using only protein sequences or structure-based information without consideration of the effects of formulation conditions \citep{nadi21}. More detailed discussions regarding the application of ML algorithms in screening of molecule candidates based on biophysical and chemical properties at early stages of development and discovery of biologic drug candidates can be found in a recent review \citep{nadi21}.  

\subsection{Design of mAbs}
Section \ref{discover_mabs} presented the applications of ML models to addressing the issues regarding biophysical properties in the discovery of potential biotherapeutic drug candidates and their developability assessments including thermal stability and colloidal stability (viscosity, solubility, and aggregation). This section will review the potential of using ML algorithms to tackle the problems concerning the design of therapeutic mAbs. Together with the increasing availability of sequence databases (e.g., International Immunogenetics Information System, Abysis, DIGIT, iReceptor), structure databases (e.g., Protein Data Bank (PDB), Structural Antibody Database (SabDAb)) and experimental databases (e.g., PDBBind, Ab-Bind, or SKEMPI containing antibody–epitope interactions) \citep{noam20}, ML models have a great potential to be used as computational tools in supporting the design of mAbs \citep{nadi21}. As a result, we are moving from the random design era to the one where mAbs, their structure and underlying mechanisms related to how the mAbs bind to the target have been comprehended to the level that enables these mAbs to be designed \textit{de novo} \citep{de19}. As shown in \cite{akba22}, there are three main pillars for ML-based design of mAbs, i.e., the learning and understanding of rules underlying antibody–antigen interactions and binding, the capability of modular and non-linear optimisation of interdependent antibody design parameters, and the ability to synthesize novel antibodies beyond the training datasets (unconstrained generation). 

\subsubsection{Prediction of antibody-antigen binding}
For the investigation of antibody-antigen binding, there are three key prediction problems where ML models can be applied: a) prediction of the antibody-antigen binding interface including epitope, paratope, and epitope–paratope interactions; b) binding affinity prediction, and c) binding partner prediction including single and/or many-to-many epitope-paratope binding partners \citep{roak21}. The comprehension of rules of the epitope–paratope interactions at the atomic level is essential for antibody–antigen recognition in the effective design of therapeutics. However, experimental approaches used as a gold standard are time-consuming and expensive. Therefore, computational methods based on ML models which can predict antibody–antigen contact surfaces could be deployed as an alternative to assist rapidly in the design of therapeutic mAbs \citep{noam20}.

Epitope prediction can be grouped into two distinct application classes, i.e.,  antibody-agnostic epitope prediction and antibody-aware epitope prediction \citep{akba22}. While the former is used to identify the most desirable epitopes without prior insights of the corresponding antibodies, the latter is usually employed to determine the epitope to which a given antibody will bind.

There are several applications of ML models for the prediction of antibody-agnostic epitopes. \cite{sara06} evaluated the effectiveness of recurrent neural networks (RNN) for predicting linear B-cell epitopes on a antigenic sequence dataset including 700 B-cell epitopes taken from the BciPEP database and 700 non-epitope peptides taken randomly from the Swiss-Prot database. The input peptide length (sliding window for the input layer of the network) ranging from 10 to 20 amino acids was tested, and the best performance using 5-fold cross validation was about 66\% accuracy achieved at a window size of 16 amino acids. \cite{mago18} built a non-redundant dataset including 5,550 experimentally validated B-cell epitopes (BCEs) and 6,893 non-BCEs from the Immune Epitope Database. Then, the authors proposed a new ensemble method which is a fusion of extremely randomized tree (ERT) and gradient boosting (GB) classifiers to predict linear BCEs based on the features regarding the combination of physicochemical properties with amino acid composition, amino acid index, dipeptide composition, and chain-transition-distribution extracted from the constructed dataset. There are also other prediction tools for classification of linear BCEs and non-BCEs using SVM models such as BCPred \citep{eldo08}, BEST \citep{gafa12}, COBEpro \citep{swba09}, and AAP antigenicity scale approach \citep{chli07}. These approaches combined physicochemical properties associated with antigenicity and information generated from a sequence conservation, predicted 2D structural features, and related solvent accessibility. However, the vast majority of epitopes (about 90\%) are conformational \citep{elho10} and discontinuous in nature \citep{swba08}, and thus, linear epitopes are unlikely to capture discontinuous epitopes within antigen sequences \citep{akba22}.

The prediction approaches of discontinuous epitopes usually require the information regarding three-dimensional structure of the antigenic protein \citep{lizh10}. Compared to the prediction of linear epitopes, there have been limited numbers of methods for the effective discontinuous epitope prediction due to the limitation of available antigen-antibody complex structures. Existing methods for antibody-agnostic predictions of discontinuous epitopes train learning models on the antibody-antigen structures and make epitope prediction on the antigen structures alone \citep{akba22}. These models use amino-acid propensity scales together with geometric predictors or clusters with different shapes representing the spatial compactness of surface residues and secondary structure composition such as PEPITO \citep{swba08}, SEPPA \citep{suwu09}, DiscoTope \citep{krlu12}, and EPSVR \citep{lizh10}. Furthermore, in a recent study, \citep{luli22} proposed to use both local and global features for the prediction of BCEs. In their approach, Graph Convolutional Networks are used to capture local spatial neighborhood information in parallel with Attention-Based Bidirectional Long Short-Term Memory networks which are used to extract global information from the whole antigen sequence. The experimental results confirmed that global features play a critical role in the prediction of BCEs. However, these methods need the information of antigen structures because they used structure-based residue features as input. In contrast, the two-stage method proposed by \cite{reso17} can predict conformational B-cell epitopes using only antigen sequence information. In the first stage, sub-classifiers are deployed to learn general epitope patterns from each propensity type separately extracted from a large data set of antigen sequences containing computationally defined epitopes. In the second stage, a decision tree classifier trained on antigen sequences with diversified experimentally determined epitopes is used to identify the heterogenous complementarity of the propensities which constitute a basis for antibody-antigen interactions. Because the method only needs antigen sequence information, it can be deployed for large-scale predictions of discontinuous epitopes and broader applications, such as the discovery of novel epitopes and their corresponding antibodies,or development of new antigens for a given pathogen.

As argued in \cite{akba22}, antibody-agnostic prediction of epitopes is an ill-defined problem because an epitope only become useful in a specific context associated with an antibody. Therefore, antibody-aware epitope prediction approaches have been introduced to address this issue. \cite{sebe14} used a Random Forest model to predict the likelihood of a given residue in an antibody complementary determining region (CDR) to pair up a given residue on the antigen surface based on the features of these residues. For each antigen residue, the features comprise secondary structure, estimated disorder, surface accessibility, predicted interaction hotspots, the considered amino acid and its amino acids neighbors in antigen sequence. For each antibody residue, the extracted features include the residue position in the antibody, the considered amino acid and its amino acids neighboring in the sequence of antibody. Another study \citep{jema19} enhanced the prediction of antibody-specific epitopes by exploring features generated from structures of antigens and their associated antibodies and using them for statistical and ML algorithms. The study identified the correlation of geometric and physicochemical features to the interacting paratope and epitope patches. The inclusion of these features in the training set for feed-forward neural networks can increase the prediction accuracy of the related antigen target for a given antibody and the antibody target for a given antigen.

Although the prediction of epitopes provides useful information related to the antigen regions which can be bounded by the antibody, this prediction does not directly give the insights into the specific antibody residues which can be mutated to change their functionality \citep{krli14, noam20}. This issue may be addressed by the antibody–antigen docking technique, which provides a list of potential orientations and poses binding two molecules to each other given the structure of the antibody and the antigen. Therefore, there were several studies aiming to use the structure-based antibody-specific epitope predictions to enhance the quality of antibody-antigen docking algorithms (using mathematical and mechanistic models) by including geometrical information of both antibody and antigen to re-rank the list of possible poses provided by docking algorithms. For example, the EpiPred method \citep{krli14} combined the conformational matching of the antibody-antigen interfaces and a knowledge-based specific antibody-antigen score to return a ranked list of likely epitope regions for a given antibody. These predicted epitope regions can facilitate docking procedures when they can increase the quality of raw antibody–antigen poses provided by docking frameworks. \cite{scbu22} built 3D antibody structures from
their sequence and generated rigid-body docking poses from this modelling in combination with given antigen epitopes. This information was then used to train a convolutional neural network. The trained model can be used to rerank docks generated by the docking algorithms.

Together with antigenic epitope prediction, accurate prediction of antibody paratope can provide useful knowledge for the investigation of antibody-antigen interaction mechanisms and antibody design \citep{luli21}. Similarly to the prediction of epitopes, paratope prediction approaches can be divided into antigen-agnostic and antigen-specific groups \citep{akba22}. In each group, the structure or sequence data of antibodies and antigens can be used to train paratope predictors.

There have been several ML models proposed for the antigen-agnostic paratope prediction task which are based on a sequence or structure information of the antibody. \cite{live18} proposed the Parapred method, which uses a combination of CNN and long-short term memory networks to predict the paratopes based on only the amino acid sequence of CDRs. Amino acid sequences were encoded as tensors using the one-hot encoding method prior to being passed through the deep learning models. Experimental results showed that Parapred outperformed the proABC method \citep{olch13}, which is also a paratope prediction model using a random forest classifier trained on features extracted from amino acid residue sequences of an antibody. In contrast to Parapred, the proABC method requires additional features on the entire antibody sequence such as the predicted canonical structure associated with each CDR and hypervariable loop length, and germline family together with antigen volume apart from the heavy and light chain sequences of the antibody. In an enhanced version, proABC-2 \citep{amol20} was introduced, in which the random forest classifier is replaced with a CNN model. Otherwise, the data processing method and set of features were kept the same as the proABC. However, the experimental outcomes indicated that proABC-2 outperformed Parapred \citep{live18}. In addition to the prediction of paratope residues, proABC-2 is able to predict types of interactions including general, hydrogen bond and hydrophobic interactions of the paratope. Beside the antibody sequence information, a structure of the antibody molecule can also be used for the paratope prediction. \cite{dafe19} proposed a novel approach for the paratope prediction from antibody structures using 3D Zernike descriptors and the SVM model. The proposed method first extracted geometrical representation together with physicochemical and biochemical properties of the residues on the antibody surface from the given experimentally obtained 3D structures of the antibodies. For each spherical patch uniformly sampled from the antibody surface, the rotationally invariant local features based on 3D Zernike moments are computed. The randomized logistic regression is used within the 3D Zernike descriptors for feature selection. From the obtained features, SVM models are trained to distinguish interface surface patches belonging to the paratope from non-interface ones. To identify paratope residues, the predicted interface surface patches are then mapped on the underlying residues. Each residue in the antibody is given a score representing its probability of belonging to the paratope based on the minimum distance from its atoms to the centre of predicted paratope surface patches. Next, each residue in the antibody may be predicted as antigen binding or non-antigen binding by determining the threshold value for its final score. The experimental results showed that this proposed method outperformed other antigen-binding interface prediction tools such as Parapred \citep{live18}. Moreover, the predicted local surface patches can help docking algorithms to narrow down the conformational search space to only surface patches belonging to the paratope.

Similarly to the case of antibody-agnostic epitope prediction, the paratope prediction without considering the context of a specific antigen may not provide much insightful knowledge for the antibody design \citep{akba22}. Therefore, antigen-aware paratope prediction methods have been introduced to take benefits from the knowledge of the specific epitope residues to refine the paratope prediction. \cite{deve19} proposed a method to expand the Parapred \citep{live18} using the information about residue sequences on the target antigen apart from the features extracted from the residue in the CDR including amino acid type, six heavy and light chain sequences and seven additional features describing physical, chemical, and structural properties of the given amino acid type. These features are then used to train a deep learning architecture combining dilated convolutional layers and a self-attention layer. The proposed method slightly improved the paratope prediction accuracy compared to the original Parapred version. In addition to sequence information, several antigen-specific paratope prediction approaches have used the features extracted from the structures of antibody and antigen as input. \cite{piba20} introduced a model based on the Graph Convolution Attention Network, called PECAN, which can predict binding interfaces for both paratope and epitope. Graph convolutions are used to capture the local spatial connections of the interfaces, while an attention layer is deployed to learn distant information, and a base network trained on general protein–protein interaction data is used as a transfer learning technique for fine-tuning of hyper-parameters for epitope and paratope prediction networks. In another study, \cite{dede21} proposed a novel architecture for the deep learning model using epitope-paratope message passing (EPMP) for the prediction of both paratopes and epitopes. The proposed architecture for the paratope model sequentially handles the input antibody residue sequences in lower layers by CNNs followed by graph structure features of residues in higher message passing neural network layers. In contrast, the epitope model uses purely structural features extracted from antigen residues to train graph neural network layers in combination with contextual cues from the given antibody. Recently, a geometric deep learning has appeared as one of the most promising approaches for the generation of representations of protein structure and the molecular surface \citep{atgr21} which can be employed to predict antibody–antigen interacting interfaces \citep{gasv20}. This method expands the neural networks to integrate the geometric knowledge regarding structure and symmetry of the antigens and antibodies to refine the quality of the predictive models. 

Besides the prediction of the paratope or epitope for the known binding pairs, the extraction of paratope-epitope matching rules and generation of all possible binding partners for a given antibody or antigen are still a challenging issue. This is due to the lack of large antibody-antigen structure and affinity datasets to build the many-to-many binding partner prediction models \citep{akba22}. Therefore, \cite{akro21} performed some initial work on taking advantage of antibody-antigen complexes in the public antibody database, graph theory and deep learning to formulate a set of antibody-antigen structural
interaction motifs. These motifs have illustrated the potential predictability of antibody–antigen interactions in general and paratope-epitope pairs in particular. The generated structural interaction motifs enable to predict paratope-epitope interfaces of unrelated antibody-antigen complexes. The initial results in this study indicated that the availability of general interaction vocabulary of antibody–antigen interfaces may facilitate learning and extraction of antibody–antigen interaction rules.

Another key problem in the prediction of antibody-antigen binding is the prediction of binding affinity changes of the antibody toward a given antigen based on the mutations in the amino acid sequence of the antibody. The improvement of binding affinity is essential for increasing the efficacy and reducing the amount of antibody per dose to lower price \citep{kusa20}. ML models have been used to predict the antibody-antigen interaction binding affinity changes based on mutations. \cite{pias16} developed a web server, namely mCSM-AB, to predict the effect of amino-acid substitutions on the antibody–antigen affinity changes of an antibody for a given antigen based on graph-based signatures. For each single-point mutation on the antibody-antigen complex, the proposed method extracts the wild-type residue environment. The wild-type residue environment is then used to generate a structural signature and the pharmacophore count difference between wild-type and mutant residues. These features along with experimental antibody-antigen affinity difference from the literature are used to train K-Nearest Neighbours, Regression Trees, and Gaussian Processes models for prediction. However, mCSM-AB has used only structural information for the prediction of binding affinity changes upon mutations, while evolutionary information and energetic terms which have shown significant impacts of the mutation on the antibody binding affinity has not been used for the prediction. Therefore, in an updated and refined version of mCS-AB, namely mCS-AB2 \citep{myro20}, the learning models have been trained on a larger and more comprehensive dataset, which utilises not only graph-based signatures (pharmacophore and distance pattern) but also evolutionary and energy-based features as well as interatomic interactions to better capture structural and sequence-based information for an improved predictive performance of antibody-antigen binding affinity changes due to mutations. The learning models in the mCSM-AB2 are popular classifiers available in the scikit-learn library including Extra Trees, Random Forest, Gradient Boosting and XGBoost regression. \cite{kusa20} also proposed to use ensemble models to predict the antibody affinity changes upon single point mutations. The authors used the predicted outcomes of antibody affinity changes from 11 molecular mechanics predictors and the mCSM-AB model as input features to train Gaussian process regression and random forest regressor to provide the final prediction for the antibody affinity changes upon mutations. The obtained results showed better classification performance than molecular mechanics-based affinity predictors. While these studies predict the antibody-antigen binding affinity changes using a single point mutation at a time, \cite{mypi20} proposed an approach, called mmCSM-AB, with the ability to predict the binding affinity changes upon multiple point mutations using graph-based signatures, sequence and structure based information of antibody-antigen complexes. A wide range of ML models including Random Forest, Extra Trees, Gradient Boosting, XGBoost, SVM and Gaussian Process was considered in the mmCSM-AB tool.

\begin{figure}[!ht]
    \centering
    \includegraphics[width=1\textwidth]{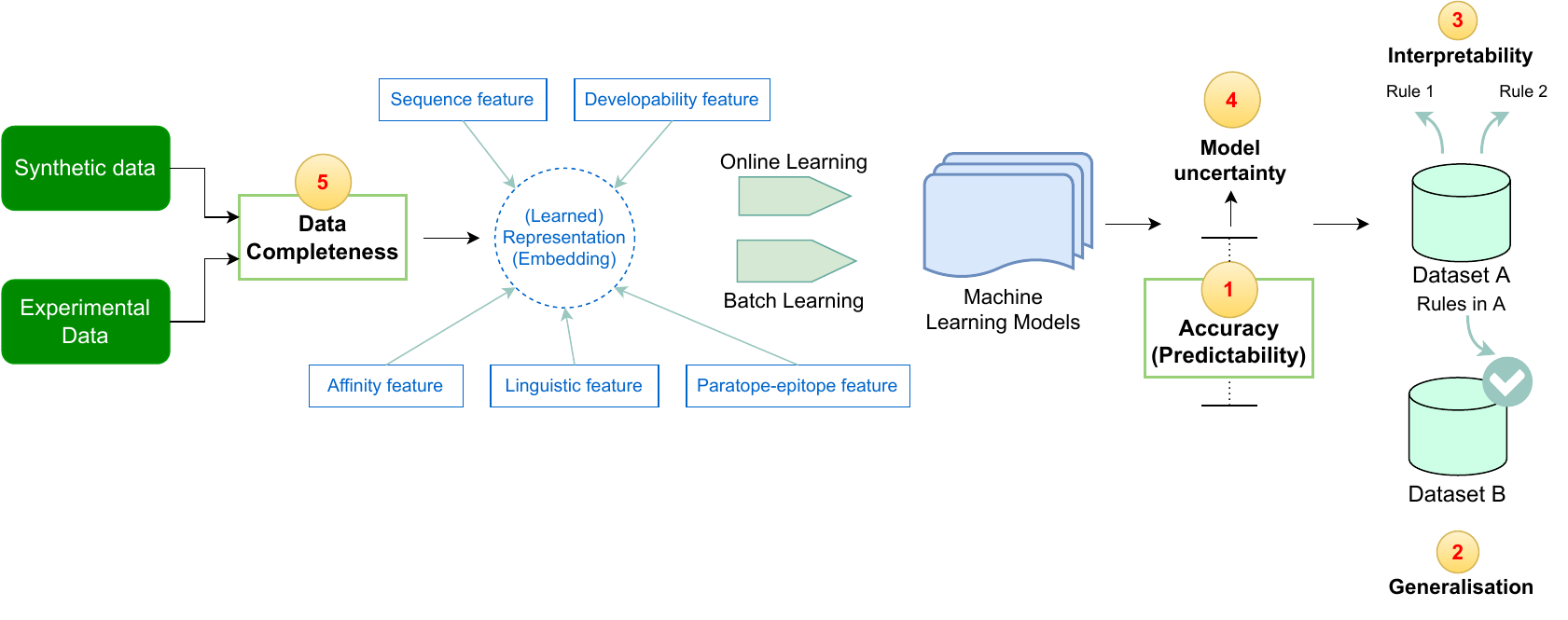}
    \caption{Five key challenges for the applications of ML to the antibody–antigen binding prediction (adapted from \cite{akba22})}
    \label{aa_binding_challenges}
\end{figure}

Although the applications of ML models for the prediction of antibody and antigen binding have made considerable progress in recent years, as presented above, there are five key challenges still pertaining to the learning models for this topic including predictability, generalisation, interpretability, uncertainty, and data completeness \citep{akba22}. These challenges are summarised in Fig. \ref{aa_binding_challenges} within a typical ML workflow. Predictability is the capacity of ML models to accurately predict properties of antibody-antigen binding, which is essential to assist the design of antibodies. However, the accuracy of predictive models is usually impacted by the biological complexity and the restricted information of the considered datasets. The generalisation of learning models refers to the performance of the trained models on different datasets without the need for additional training. The generalisation is important for the ML-based mAbs design. To achieve good generalisation performance, a learning model needs to consider the similarity of antibodies in terms of both sequence similarity and binding behaviors because sequence-similar antibodies may bind to different antigens. Another challenge for the ML models to be used for mAbs design in practice is the ability to explain the reasons behind their predicted outcomes, i.e., showing the rules for prediction and generalisation. The interpretability of ML models would reduce the risks of results generated from data-dependent properties and provide insights for development of novel antibody sequences. This characteristic is still a challenge of ML models for antibody-antigen binding prediction as most of the predictive approaches are deep learning models which are inherently difficult to extract interpretable knowledge from. Model uncertainty is also another challenge for the application of ML models to this theme. It is because training data are often noisy and sparse, so it is difficult for learning models to exhaustively learn the underlying rules for the prediction. As a result, the learning models could converge to an approximate set of biased rules or short-cut solutions. The completeness of a training dataset is the amount of information contained in this dataset to assist ML models in inferring predictive results. It is still challenging to assess the completeness of training sets for the prediction of antibody-antigen binding.

\subsubsection{Learning and optimisation of interdependent antibody design parameters}
In order for the potential candidates of therapeutic mAbs to be manufactured at commercial scales and successfully go through clinical procedures, they need to overcome several developability problems \citep{rama19}. The developability of an antibody is evaluated based on a number of biophysical and biochemical properties including the propensity to stability, aggregation, immunogenicity, plasma half-life and clearance \citep{bami20}. Early screening of developability with properties satisfying manufacturing requirements is essential to reduce the time and cost for the choice of lead mAb candidates. The development of \textit{in silico} tools equipped with ML algorithms for the prediction of developability parameters could help to enhance the quality of mAb designs. According to \cite{akba22}, the design process needs to take into account the interdependence and non-linearity of antibody design parameters. 

The applications of ML models for the prediction of antibody design parameters in terms of developability properties based on aggregation and stability including thermal and pH stability, solubility, and concentration-dependent viscosity were presented in Section \ref{discover_mabs}. This section will discuss the applications of ML algorithms to the estimation and identification of antibody design parameters in terms of the plasma half-life, clearance, and immunogenicity.

The serum half-life and clearance are important factors influencing the efficacy and optimal dosing regimen of mAbs \citep{gowa18}. The half-life of Immunoglobulin G (IgG) of mAbs can be adjusted by optimising several physicochemical parameters such as the Neonatal Fc-receptor (FcRn) affinity, the isoelectric point (pI), binding poly-specificity reagent (PSR) and thermal stability \citep{akba22}. These key parameters have been identified through several studies with the support of ML algorithms. \cite{gowa18} deployed the least absolute shrinkage and selection operator regression (LASSO) to determine the key combination of biophysical parameters for highly accurate prediction of half-life and clearance with the mAbs clinical half-life and clearance data. The analytical results showed that FcRn binding at pH 7.0 and thermodynamic stability (particularly the change in heat capacity upon unfolding) result in the most accurate prediction of half-life and clearance for mAbs. In a recent study, \cite{grth21} analysed the impact of 40 biophysical and sequence-based parameters (12 measurements \textit{in vitro} and 28 computed in \textit{in silico}) on mAb clearance over 48 IgG1 antibodies. The authors used the Random Forest model for the classification of the mAb linear clearance (normal clearance and fast clearance). The ranking analysis of biophysical variables over 10,000 loops of the RF model showed that \textit{in silico} computed sequence-based pI and PSR are the most important parameters influencing the human clearance. These results aligned with the findings in previous studies on the impacts of pI and PSR on the pre-clinical clearance  \citep{igts10, kesu15, dalu15}.

Antibodies developed from non-human sources such as B-cell repertoires of animals is more unlikely to be tolerated by humans and so it can lead to immunogenic responses. This phenomenon is known as immunogenicity. These immunogenic responses can result in negative effects on safety and pharmacokinetic properties of therapeutic mAbs and the generation of neutralising antibodies causing the loss of efficacy \citep{luhw20}. Therefore, the evaluation of immunogenicity of mAbs is a critical step in the design of mAbs. Humanization is one of the methods to reduce immunogenicity while still preserving the mAb efficacy \citep{mahu21}. Humanization approach of a mAb involves the partial substitution of non-human CDR sequences using a referenced human scaffold. An efficient humanization method needs the ability to accurately identify the humanness of a sequence. Higher humanness scores are usually associated with lower immunogenicity levels. Several studies have applied ML models to the assessment of antibody humanness and acceleration of mAb humanization via \textit{in silico} solutions. \cite{cldi18} proposed to use the multivariate Gaussian model to estimate the degree of humanness in consideration of the correlations between pairs of residues within and among amino acids at different positions. Although the proposed method may accurately classify human and murine sequences, the humanness score showed only a weak negative correlation to experimental immunogenicity levels. In another study, \cite{woxu19} trained a bi-directional long short-term memory network on human antibody sequences and the model showed the better performance in discriminating between the antibody sequences of human and other species than the method proposed in \cite{cldi18}. The trained model can be used for selecting suitable scaffolds and mutations to assist in humanization of antibodies. \cite{mahu21} built RF classifiers to discriminate each human V gene type from non-human sequences. The model is trained on large-scale repertoire sequence data including human and murine precursor sequences taken from  the Observed Antibody Space (OAS) database \citep{kole18}. The trained RF model is then used to develop a humanization tool, called Hu-mAb, that can humanize both the VH and VL sequences of potential mAbs by proposing mutations increasing humanness. The Hu-mAb tool could suggest optimal solutions which can minimise the number of mutations made in the sequence to reduce the effect on efficacy of mAbs. The experimental results showed that the humanness scores provided by the RF model represent a negative correlation with experimentally observed immunogenicity. It means that the sequences associated with a higher humanness score could have lower immunogenicity levels. As a result, the mutations proposed by the Hu-mAb are similar to those provided by experimental humanization studies. However, the non-human sequences in the OAS database are almost all murine sources, and thus the Hu-mAb can only perform effective humanization on murine precursor sequences. In addition, the tool only supports V-typed antibody formats. The OAS database was also used in the study of \cite{prma22}. The authors developed an open-source platform for humanization (Sapiens) and humanness evaluation (OASis). The OASis can provide granular and interpretable humanness scores using 9-mer peptide search in the OAS which is able to precisely separate human and non-human sequences with results correlated with clinical immunogenicity. The Sapiens is a deep learning humanization technique, which train a deep neural network with the Transformer encoder architecture using language modeling. The input sentences of the model are amino acid sequences of antibody variable regions, while the word tokens are individual amino acid residues. The experimental outcomes showed that the Sapiens can achieve a higher humanness enhancement than Hu-mAb and comparable to experimental
methods. The tool can provide viable point mutations that can be experimentally validated, while preserving specificity and binding affinity of mAbs.

Another approach to estimate the immunogenicity of mAbs is based on the adaptive immune system activation mechanism by  analysing the peptide presentation on the Major Histocompatibility Complex class II (MHC II) molecules recognised by T cells. This characteristic is called immunopeptidome, which can provide valuable knowledge for immunogenicity estimation \citep{sori16b, ducr16}. With the popularity of immunopeptidome public databases such as  Immune Epitope Database~\footnote{www.iedb.org} (IEDB) and ones presented in \cite{dodo21}, various \textit{in silico} tools have been developed to predict protein immunogenicity using immunogenic peptides. Different algorithms including the use of ML models such as ANNs and SVM, discussed in a recent review of \cite{dodo21}, could utilise the sequence or structural data of proteins to estimate their efficacy to stimulate a T-cell response.

In summary, although there have been many advanced algorithms proposed and improvements made in efficiency and accuracy of \textit{in silico} predictive approaches of developability parameters for potential mAb candidates, several challenges
remain unsolved \citep{akba22}. For example, computational immunogenicity predictions have not yet completely substituted the \textit{in vitro} testing in animals because of the safety requirements accompanying the specific parameters \citep{dodo21}. \cite{kuts20} argued that it is likely to be tough to base only on \textit{in silico} approaches for the assessment of immunogenicity due to the limitation in the understanding of biological rules for immunogenicity. Moreover, finding a multi-objective global optimisation for all developability parameters for a given antibody is still a challenging requirement because the enhancement of a developability measure could lead to the sacrifice of another. For instance, \cite{kuts20} indicated that the improvement of antibody stability might lead to the increase of aggregation. A promising solution to learn and optimise multiple design parameters for a given mAb is the combination of many ML models trained on data from various experimental campaigns \citep{lize20}.

\subsubsection{Generation of in silico antibody sequences}
\begin{figure}[!ht]
    \centering
    \includegraphics[width=1\textwidth]{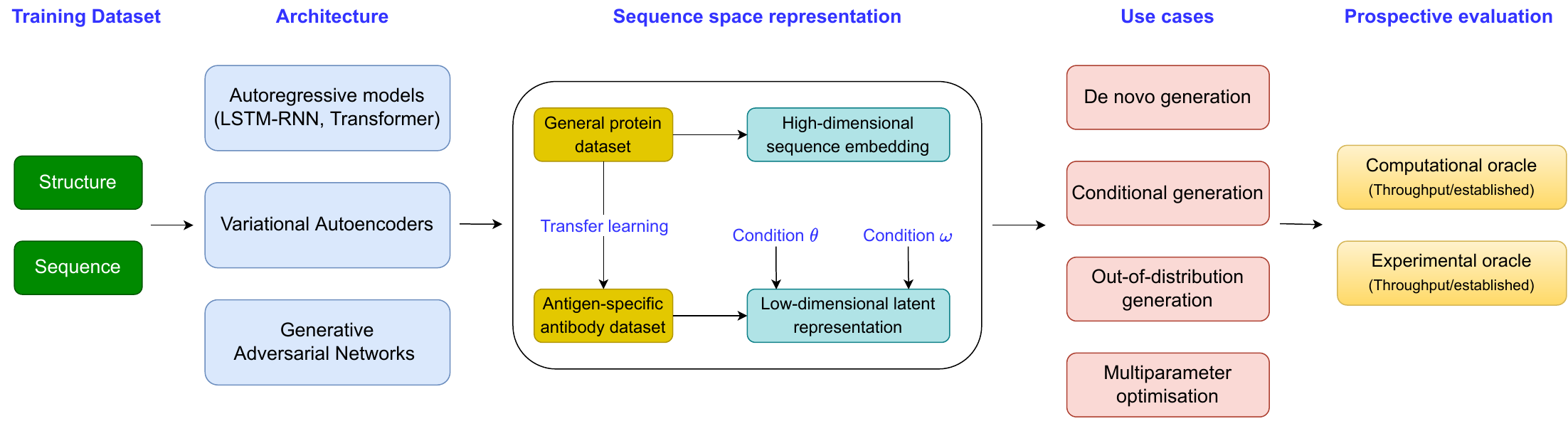}
    \caption{Main steps of an in silico generation of potential candidates in the mAb design process (adapted from \cite{akba22})}
    \label{mab_generation}
\end{figure}

As presented in a recent survey \citep{akba22}, the \textit{in silico} generation of antibody sequences has been dominated by the deep learning models with three typical architectures namely Variational Autoencoders (VAE), Generative Adversarial Networks (GAN), and Autoregressive (AR) models including long short-term memory recurrent neural
network (LSTM-RNN) and a transformer. These generative models can be trained on general or customised datasets to learn sequence space representations and synthesise novel sequences. The autoregressive models are able to generate highly diverse mAb sequences and meaningful sequence embeddings reducing the demand for manual features. In the similar manner, the VAEs and GANs have been used to generate biologically meaningful latent representations and condition them on extra features (e.g., viscosity, solubility). As a result, these models could be deployed for \textit{de novo} sequence generation, conditional or out-of-distribution generation, and optimisation of multiple mAb design parameters. Transfer learning approaches can be used to infer higher-order and function-specific interactions from a small number of available sequences based on the insights learned from general large datasets. To assess the effectiveness of generative models, it is necessary to have external computational or experimental validators (oracles) for resulting antibody sequences after generation. Experimental evaluation methods may be time-consuming and costly, so it is desirable to develop \textit{in silico} oracles. This generation process of \textit{in silico} antibody sequences is illustrated in Fig. \ref{mab_generation}. 

GAN models are usually used to explore the prospective distribution of an actual dataset by deploying a generator and a discriminator in a zero-sum game. The best performance will be obtained as the discriminator no longer has the ability to distinguish the outputs of the generator from the actual distribution. Therefore, GANs posses a great potential for generation of mAb sequences with a particular property or novel sequences (out-of-distribution sampling) from existing distributions of available sequence spaces. \cite{amsh20} illustrated the use of the GAN model trained on 400,000 light and heavy chain human-repertoire sequences randomly taken from the OAS database for generation of novel antibody therapeutics. The resulting sequences generated by the GAN model can mimic typical human repertoire features
like diversity and immunogenicity. Specifically, the authors showed that the proposed method can bias the model to achieve the improved developability and chemical and biophysical properties of interest by continued training of the learned model with a data subset of specific desirable properties. For example, as shown in the outcomes, this transfer learning approach can reduce negative surface area patches of antibodies (recognised as a potential aggregation source, thermal instability, and possible half-life decreasing), bias of a higher pI to lower aggregation, lower binding affinity toward MHCII, or going towards longer CDR3 lengths (identified as a method to increase diversity and formulate more effective therapeutics for a given target). The authors experimentally validated their method \textit{in vivo} by expressing GAN-generated antibody sequences via phage display and in stable Chinese hamster ovary (CHO) cells. The experimental results showed that the generated antibodies exhibit multiple desirable biophysical properties.

In the mAb design, VAEs could be employed to lower the dimensionality of a dataset, formulate biologically meaningful representations and clusters, generate \textit{de novo} sequences, and create the mAb sequences with the expected function by interpolating in the latent space. \cite{egch22} proposed a method for novel and high-quality 3D backbone structures of antibodies using a VAE trained on the structural data collected from the antibody structure database AbDb/abYbank \citep{fema18}. The experimental results showed that the backbones generated by the VAE model are new and chemically accurate and able to be used in other design protocols to achieve biochemically realistic sequences. For example, the use of a constrained optimisation and simulation framework such as Rosetta \citep{lety11} does not result in a large change in the overall structure of antibodies generated by the VAE model but there is a refinement of chemical validity. In addition, by constraining and optimising a structure in the latent space of the VAE model, it can generate structures satisfying user-defined features. The authors demonstrated these characteristics by integrating the proposed method with Rosetta to generate a single-domain-binder to the ACE2 epitope of the SARS-CoV2 receptor binding domain via discrete sampling and constrained optimisation of the mAb structures in the latent space. The outcomes showed that the latent space of VAEs can be used to generate new binding proteins and provide rational mAb candidates for further experimental validation. In another study, \cite{frne20} also introduced a VAE model trained on antibody repertoires from the bone marrow of antigen-immunized mice to model the underlying B-cell receptor recombination process, in which a Gaussian mixture model is applied to latent space to cluster potentially antigen-specific antibody sequences. This method can provide both the latent embedding and cluster labels for similar sequences. Hence, the proposed VAE model could synthesise new sequences and group input sequences to distinct clusters based on the learned latent space embedding. The authors experimentally verified the proposed method by recombinantly expressing twelve antibody sequence variants from one cluster mapped to one of the confirmed binders. The results confirmed that all of these 12 variants were antigen-specific.

Due to inherently sequential nature of autoregressive models, they are usually suitable for modeling biological sequences, where the order of amino acids is fixed and sequential (one amino acid after the other). In a benchmarking research regarding the antibody sequence design, \cite{meda21} indicated that AR models are appropriate choices for the specific antibody design tasks including grafting specific regions like a short CDR3 within an input sequence and generation of diverse sequences with broader coverage while still preserving the consistency with a given input sequence. AR models can be used to generate antibody sequences with the optimised binding affinity. \cite{saka21} demonstrated the use of a LSTM neural network for the generation of high antibody against kynurenine, which is a metabolite concerning the niacin synthesis pathway. The LSTM model is trained on enriched antibody sequences by next-generation sequencing techniques. Then, virtual sequences are generated from the trained LSTM model by mimicking enriched sequences, and the priority is put into the promising sequences with high affinity based on the likelihood scores of binding affinity computed by the trained model. The outcomes confirmed that the likelihood values of antibody sequences generated from the proposed model are highly correlated with the quantitative binding affinity. Compared to frequency based screening techniques, the generated sequences by LSTM models also have a greater affinity. The LSTM model can also be used to learn distributions of antibody sequences together with a wide range of design parameters including paratope, epitope, affinity, and developability. This characteristic was illustrated in the study of \cite{akba22b}. The authors generated \textit{in silico} training dataset of 1D native antibody sequences (CDR-H3, the third CDR of the antibody heavy chain) (taken from \cite{grme17}) annotated with their corresponding paratope, epitope, and affinity to each 3D antigen (from the Antibody Database \citep{fema18}) using the antibody-antigen binding simulation framework, called \textit{Absolut!} \citep{roak21}. Then, these 1D antigen-aware CDR-H3 sequences were used to train an LSTM generative model. Note that the LSTM model was not trained on any explicit 3D information on the paratope, epitope, affinity, nor the developability parameters of a given sequence. The trained model was then deployed to generate novel CDR-H3 sequences, which were assessed for their antigen-specificity including affinity, paratope, epitope associated with developability parameters using \textit{Absolut!}. The obtained results showed that the generated CDR-H3 sequences had the developability parameters and affinity values similar or better than those of training sequences. This research confirmed that the AR models are capable of learning 3D-affinity, paratope, and epitope information from 1D antibody sequence data. In another study, \cite{shri21} showed the efficiency of the AR model in designing new single-domain antibody sequences (also known as nanobodies) from existing natural nanobody repertoires. The authors employed dilated convolutional neural networks trained on around 1.2 million llama nanobody sequences from the immune repertoires to learn the constraints characterising functional nanobodies. The model learned conditional distributions over CDR-H3 sequences conditioned on the CDR-H1 and CDR-H2 sequences to eliminate CDR-H3 sequences which are not chemically compatible with the other CDR-Hs. The experimental results confirmed that the new CDR-H3 sequences generated by the causal CNNs have similar distributions of hydrophobicity and isoelectric point in comparison to the reference natural CDR-H3 sequences.

In conclusion, the current studies on \textit{in silico} generative modeling of antibodies have mainly focused on the generation of a new 3D backbone structure of antibodies with expected properties \citep{egch22}, sampling antibody sequences with high affinity \citep{saka21}, clustering and interpolation of the latent space for discovering new antibodies \citep{egch22, frne20}, and the generation of new and diverse antibody sequences satisfying the desirable developability parameters \citep{amsh20, akba22b}. However, the synthesis of new antibodies using generative models also faces many challenges. The first one is the lack of effective external (computational or experimental) oracles to assess the outputs of generative models. The second issue is the difficulty in the simultaneous optimisation of multiple developability parameters and binding affinity during the generation of new antibodies as the label (e.g., binding affinity, developability, and plasma half-life ) of new antibodies is not known \textit{a priori}. Therefore, it is highly desirable to develop high-throughput experimental oracles.

\subsection{Design of other therapeutic proteins}
Protein engineering is a progressive process of designing or developing proteins which meet a set of valuable properties to be used in scientific or medical applications \citep{yawu19}. The discovery of proteins with expected properties and functions is usually resolved through three approaches of protein engineering or design including rational design, directed evolution, or \textit{de novo} design \citep{bofu20, nadi21}. This section will review the typical applications of ML to protein engineering and design of general proteins through these three main routes. While the rational design of proteins focuses on building predictive models from structure or sequence data to predict the properties (e.g., affinity, stability) of a given input protein, the directed evolution workflow aims to combine the predictive modelling with optimisation algorithms to generate new proteins with pre-defined desired properties. Meanwhile, \textit{de novo} protein design is directed towards building generative models to explore the input space together with perturbing the input data for generation of new resulting proteins unrelated to those in nature on the basis of physical principles of intra- and inter-molecular interactions \citep{hubo16}.

Rational protein engineering methods aim to develop quantitative models for protein properties, and deploy them to more efficiently discover the protein fitness landscape (the relationship between a given set of mutations and the protein function of interest) to bypass the directed evolution challenges \citep{alkh19, roch17, hubo16, rokr13}. Such rational design needs a comprehensive and predictive understanding of structural stability and quantitative molecular function of the proteins \citep{alkh19}. As a result, the quantitative structure-activity relationship (QSAR) modeling using ML algorithms have been introduced for virtual screening of prospective biological active molecules from millions of candidate compounds cheaply and rapidly \citep{wawu15}. The predictive models usually use the sequence or structure data of proteins represented under the form of input feature vectors or matrices and predict the target protein properties. The details of the use of traditional ML models and deep learning for QSAR approaches can be found in several surveys such as \cite{zhta17, xuve20, maak21}.

Directed evolution techniques are used to repeatedly enhance an expected function of a given protein through mutation and selection steps of an initial protein sequence. Machine-learning-assisted directed evolution is a branch of protein engineering aiming at optimisation of protein functions \citep{yawu19}. Data driven ML models can be used to predict the way of mapping protein sequences to corresponding functions without the demand for a detailed approach of the underlying physics or biological pathways. Such approaches enable to accelerate the directed evolution by learning from the properties of given variants and utilising this learned knowledge to choose sequences likely to express the enhanced properties. As shown in \cite{yawu19}, machine learning has been highlighted as a potential tool in the directed evolution workflow. By reducing the number of variants which need to be verified and learning from each directed evolution round \citep{bofu20}, ML-guided directed evolution has resulted in new approaches for optogenetics \citep{beya19}, the engineering of reference green fluorescent proteins \citep{saoi18}, protein thermostability \citep{rokr13}, and demonstrated some potential to address the problem of epistatic effects derived from combinatorial mutations \citep{wuka19}.

\textit{De novo} methods aim to take advantage of the protein folding principals to design new polypeptide sequences which can fold into a stable 3D structure and are associated with an existing or new desirable functionality.  Similarly to the design of mAbs, generative DL models including GAN, VAE, and Autoregressive models have been used for \textit{de novo} design of protein sequences with expected functionalities like solubility, stability, and affinity \citep{kazh20}. \cite{reja21} proposed a new GAN architecture using temporal convolutional networks and a self-attention layer to learn the meaningful sequence motifs and evolutionary relationships directly from the complex multidimensional amino-acid sequences. The trained model then can generate novel and highly diverse sequence variants exhibiting natural-like physical properties within the given biological constraints. \cite{coma19} proposed a VAE-based protein sequence generation method using residual networks, dilated convolutions, and an autoregressive module on top of the decoder trained on the protein sequence and phenotype data. The latent space of the proposed architecture is able to accurately reconstruct sequences with a set of desirable functions and phenotypes. The research is a step forward to build a model able to learn general structures of protein sequences and design general-purpose structure free proteins. \cite{rime21} proposed to use a Transformer network and \cite{alkh19} used LSTM-RNNs trained on large unlabelled sequence databases to generate sequence embeddings which are able to capture the secondary and tertiary structures and funtionalities of a protein as well as amino acid biochemical properties. \cite{bikh21} demonstrated how such sequence embeddings can be used for \textit{de novo} protein evolution. First, the general embeddings were generated by an LSTM-RNN network trained on 20 million general protein sequences. Then, these embeddings were fine-tuned on sequences of proteins of engineering interest, followed by the sampling procedure of a small number of mutants, quantifying their particular functional properties and developing a linear regression model with inputs being embedding representations and outputs being quantified protein properties. The trained model is then used to propose the optimal designs evolved from a starting protein sequence by mutating, embedding, and prediction of the desirable properties. This process is repeated until the expected functionality of the protein is achieved. There have been several studies demonstrating the capability of conditional transformer models, which have been successfully used for text generation, for protein engineering problems, i.e., \textit{de novo} protein design. \cite{inga19} proposed a conditional transformer model for designing protein sequences using a graph representation of their 3D structures. The proposed architecture can effectively employ the spatial locality of dependencies in molecular structures for the generation of new protein sequences. Another study performed by \cite{mamc20} can create new functional protein sequences with existing folding motifs by training a Transformer network on diversified sequences with conditioning tags which encode annotations like organism, function, location, and process information. The proposed model is able to learn these conditional probabilities over amino acids and conditioning tags of interest. The experimental results indicated that the protein sequences derived from conditional
labels are different from the similar sequences in the training data, and to some extent the model can generate out-of-distribution sequences.

\subsection{Cell-line selection}
Cell-line selection is a gene transfection process of protein of interest into host cells to build a heterogeneous cell pool. Cells are then sorted into single-cell cultures, and the cell-line leading with regard to the highest quantity, quality and stability across sequentially scaled-up cultures is chosen for the master cell bank, which are usually utilised for manufacturing phases in clinical trials and commercial use later on \citep{nadi21}. Traditional approaches of cell-line selection are time and labour consuming and dramatically restricted by the number of clones which can be screened to obtain high-production clones \citep{bral07}. Therefore, the high-throughput automation tools and image analysis techniques have been developed to quickly screen high quality clones from a large population of heterogeneous cells.

\cite{gogu16} compiled a bibliography of more than 1000 articles on CHO cells and analysed the data using multivariate statistical methods. They found significant differences among cell lines regarding specific productivity, growth rate, and viable cell density. Despite this variation, they were able to draw some general conclusions that were independent of the particular cell line. For example, they observed a correlation between titer and cell viability, as well as a reverse correlation between growth rate and both specific productivity and titer. Therefore, the identification of cell-line and clone specific characteristics is crucial for the design of integrated bioprocesses \citep{basu20}. Changes in these characteristics can have a significant impact on upstream processes and, consequently, downstream process design.

As reported in the literature, ML models have been deployed for image analysis in fluorescence-based cell sorting aiming to separate the heterogeneous pool of cells into single-cell cultures. \cite{guzh19} developed a real-time image guided cell sorting and classification system using image processing methods and an SVM model. The constructed system can generate bright field and fluorescence cell images under the form of time domain waveforms. Cell images are then rebuilt from such time domain waveforms. After that, the system extracts image features and derives image-based gating criteria. Finally, cell images are handled, and the cell sorting process is performed in real time using the SVM model based on the image-guided gating criteria. \cite{nisu18} introduced an intelligent image-activated cell sorter method based on image analysis and classification using CNN models. The system combined high-throughput cell microscopy, focusing, and sorting within a hybrid software and hardware data management framework, which allows real-time automated execution from data acquisition, processing, decision making to actuation. The images of sorted cells can be employed to further train the deep CNN models for improving sort decisions and optimisation of the recognition of extremely rare cells, e.g., circulating tumor cells, antigen-specific T cells and cancer stem cells.

\section{The applications of ML in understanding, prediction, optimisation, monitoring and control of upstream processes} \label{upstream}
In biopharmaceutical process development manufacturing, the upstream process refers to the initial stages of production including the cultivation and harvesting of cells that produce the active pharmaceutical ingredient \citep{paqu17}. The main role of the upstream process is to ensure the growth and reproduction of the cells producing the active pharmaceutical ingredients in large enough quantities to meet the production demand \citep{sa22}. Data analytics and modelling techniques are frequently deployed in the upstream process development mainly because it plays an important role in achieving the maximum quality and quantity of products and a large amount of valuable data is usually derived during this stage \citep{sori17, rile20}. Therefore, this section will present a comprehensive review of how ML models applied to the selection and optimisation of cell culture media, the determination of the variations in the upstream process performance and product quality with regard to the changes in operational conditions and CPPs of bioreactors as well as understanding and extracting insights from upstream processes based on the collected data. This section also reviews the applications of ML models to real-time monitoring and controlling of the upstream process as well as transferring knowledge across different scales of cell cultivation experiments.

\subsection{Efficient screening of cell culture media supplements}
Cell culture media optimisation is an effective way to enhance the quality of recombinant proteins produced in cell culture in addition to the development of cell lines and culture conditions \citep{reda15, ivvi14}. By optimising the media, it is possible to fine-tune the glycosylation of the protein within the capabilities of a particular cell line \citep{brjo15, thwu15}. This approach can also reduce the complexity and timeline of product development, as it avoids the need to redevelop the cell line \citep{brso17}. High-throughput cell culture systems allow for the simultaneous testing of many different conditions \citep{bhku11}, enabling the use of statistical tools and rational experimental design to assess simultaneously multiple quality attributes in media optimisation \citep{sori16}. This can improve the efficiency of process development by providing valuable insights at an early stage and allowing for a more focused approach to scaling up the process, resulting in reduced experimental efforts.

The culture medium's composition, which can include sugars, amino acids, vitamins, and salts \citep{riwu18}, needs to be carefully chosen to support cell growth and production \citep{lusu05, sach22}. A comprehensive review conducted by \cite{riwu18} on the culture media developments, the effects of
critical media components on cell growth, viability, product
titer and quality emphasised the importance of maintaining a strict balance in media formulation. For example, trace
elements can be crucial to obtain higher productivity, but at high concentrations can result in toxic effects. However, understanding how each component of the medium affects cellular metabolism can be complicated \citep{azed20}, making it challenging to optimise the medium \citep{haoz22}. The statistical and mathematical methods are usually used for cell culture medium optimisation \citep{siha17, aldu07}. Nonetheless, the use of statistical method, i.e., design of experiments (DoE), for media optimisation is effective when there are fewer than ten medium components that need to be adjusted \citep{siha17}. When using the mathematical method such as response surface methodology (RSM), which deploys a quadratic polynomial approximation, the complex interactions between the medium and cells may not be accurately represented by simple equations \citep{sima09}. To address these limitations, machine learning methods have been proposed as an alternative approach for medium development \citep{grwa21}, and they have been shown to perform better than DoE and RSM methods \citep{sima09, cobl21}. \cite{brso17} proposed to use a combination method of high-throughput optimisation and multivariate analysis including principal component analysis and decision trees to rapidly screen potentially influential cell culture media supplements aiming to enhance the quality of biosimilar products through the selection of the best performing glycosylation profiles of the target product. \cite{xuya14} used a Plackett-Burman statistical design and a support vector regression (SVR) to efficiently screen a large number of combinations of medium supplements and develop a serum-free medium. The SVR model is deployed to predict the growth rate and identify the key factors in medium supplements that impact the growth rate of Chinese hamster ovary cells. The experimental outcomes indicated that the combination of Plackett-Burman design and support vector regression is a powerful technique for optimising cell growth and analysing the interactive effects of the most influential medium supplements. In a recent study, \cite{haoz22} presented the first study using active learning combining gradient-boosting decision trees with experimental validation, i.e., the high-throughput assays to fine-tune medium components aiming to achieve improved cell culture media. The experimental result showed that there was a significant decrease in fetal bovine serum and a differentiation in vitamins and amino acids when the media was fine-tuned.

\subsection{Design of experiments}
Within the QbD framework, the design of experiments (DoE) provides a systematic way for designing and analysing experiments in order to simultaneously assess multiple input variables of interest and represent their relationships and impacts on the output variables such as product yield, CQAs and cell growth \citep{magr09, moku19, wamy22}. Typically, a screening design step is deployed initially to identify critical process variables, such as medium compositions and process parameters. Next, the experimental space, defined by the process variables together with the variations in their ranges of values, is performed to figure out their impacts on targeted outcomes. The results of these experiments are then used to develop predictive models for the relationships between process variables and the output variables of interest. The goal of DoE is to find a stable set point for the process variables (e.g., medium concentrations), which ensures process stability and defines a ``design space" for stable operation \citep{moku19}.

One major disadvantage of DoE methods in bioprocess development is the demand for high number of costly and time-consuming experiments \citep{magr09, mopo17}. Additionally, these methods usually rely on user-defined selections of the experimental design and value ranges of process parameters \citep{mabr08, magr09}. This requires a significant amount of expert knowledge to define appropriate boundary values for the development and optimisation of cell culture processes. To partly address these drawbacks, mathematical process models (e.g., mechanistic models) and ML algorithms can be integrated into DoE approaches to leverage  the data from previous cultivation experiments for the development of predictive models for target variables based on the values of input variables to reduce the number of laboratory experiments. The details of the way of combining mathematical process models with DoE to build the simulated DoE via predictive models were presented in \cite{moku19}. Meanwhile, a review of the application of advanced ML algorithms and hybrid ML-mechanistic modeling methods to improve the accuracy of model predictions, identify the relationships between CPPs and CQAs, and decrease the number of experiments required in the DoE methods for the study of changes in process conditions and/or products can be found in \cite{wamy22}. In the later subsections, we will present a comprehensive review of how ML approaches applied to prediction, monitoring, and understanding of the impacts of CPPs on product quality and CQAs are used in the upstream process development. To reduce the number of experiments required in the DoE approaches and accelerate the design screening step, transferring process knowledge collected from small scale experiments to validation experiments at larger scales is highly desirable to save valuable resources and costs. \cite{badu21} investigated the transferability of a hybrid modelling (ANNs and mechanistic models) for CHO cell cultivation from a 300 ml bolus feeding shaker to 15L continuous feeding stirred tank bioreactor with the same DoE settings. The obtained results showed low errors when transferring the knowledge learned in prediction models from the shaker to bioreactors. The research confirmed the potential in reducing the number of large-scale practical experiments required for DoE approaches.

Another limitation of the DoE method is that it only considers factors that do not change over time \citep{ge13}. In batch processes, the change in operating conditions, such as temperature, over time can greatly affect the outcome of the process. The traditional DoE method does not provide a thorough and systematic way to assess the effect of time-varying operating conditions on the process results. To address this limitation, \cite{ge13} proposed a generalisation of the conventional DoE, called the design of dynamic
experiments (DoDE), which allows to use time-varying inputs, optimise the process and guide the design of the next experiments. Based on the data collected at the end of each batch, the proposed method construct a data-driven response surface methodology model for the optimisation of the process. As shown in \cite{wage13}, the DoDE approach is greatly influenced by the range of input domain, so to enhance the performance of the DoDE method \cite{wage17} introduced the way of incorporating prior knowledge of the process to assist in the design of the time-varying input profiles and their corresponding input ranges. The \textit{in silico} experiments in both multi-feed Hybridoma cell culture and secondary metabolite fermentation confirmed that prior knowledge, such as the necessary amount of nutrient required to sustain the production of the product at the end of the batch and the sequential scheduling of substrate feeds, can help to improve the process performance of the DoDE approaches. The authors proposed a method to reduce the number of experiments in the DoDE by using an evolutionary modeling and optimisation approach to minimise the initial number of experiments according to time and budgetary constraints. It can be seen that the optimisation methods for DoDE in the literature have used mechanistic and dynamic models together with the reduction method of parametric uncertainties, while ML algorithms have not been integrated into the DoDE.

\subsection{Early prediction of key performance indicators and critical quality attributes}
The production of biopharmaceutical products is a complex process \citep{shbr11} associated with many CQAs regarding the final products as presented in Section \ref{biopharma_process} and Table \ref{cqa_table}. In the upstream process, bioreactors are often utilised to cultivate mammalian cells (e.g. CHO), microbial cells and yeast or small plant cells. These cells serve as factories for the manufacturing of desired compounds. Through fermentation, they convert nutrients such as glucose and amino acids into valuable biopharmaceutical products such as vaccines, mAbs, and other therapeutic proteins. The variability in the operational conditions of bioreactors is more likely to result in the changes in CQAs of the final products. However, CQAs are hard to measure directly in production, so in practice it is common to monitor and predict physical and chemical critical process parameters and key performance indicators (KPIs) related to the quality attributes along the upstream and downstream portions. The common CPPs which have the potential to affect the CQAs of the final products usually collected from the cell culture at bioreactors include pH, dissolved oxygen (DO or pO\textsubscript{2}), dissolved CO\textsubscript{2} (DCO\textsubscript{2} or pCO\textsubscript{2}), temperature, pressure, osmolality, nutrients (e.g., glucose, glycerol, glutamine, and amino acids) and secondary metabolites (e.g., lactate, acetate and ammonium) concentrations \citep{mi13}. The commonly used KPIs at the bioreactor for biopharmaceutical products are product titer, cell growth rates, total and viable cell density, while bioprocess CQAs are typically associated with the quality of the protein such as the glycosylation, protein glycan distribution (e.g., afucosylated glycan), charge-variant distribution (e.g., glycated, deamidated forms), or molecular-size distribution (e.g., aggregate forms) \citep{ha13}.

ML predictors and multivariate data analysis methods have a great potential to be used for prediction of KPIs and CQAs based on the historical data of CPPs. There are two typical prediction types for predicting KPIs and quality of biologics in a cell culture process in bioreactors. These include real-time prediction at the time data is collected by monitoring instruments and the prediction of future values using data currently collected from the bioreactor \citep{wamy22}. Real-time prediction of quality attributes and performance factors to assist in monitoring the cell culture process in bioreactors will be presented in subsection \ref{monitoring_cell}. This subsection focuses on learning the relationship between the quality attributes of the product and the CPPs using ML models. Table \ref{cell_forecasting} summarises typical studies using ML and mutivariate data analysis algorithms to predict the quality attributes and key performance factors of the final products during the cell cultivation process in upstream bioreactors. As argued in \cite{wamy22}, although early prediction of KPIs and CQAs during the cell cultivation is crucial for manufacturing as it allows proactive measures to be implemented when the predicted outcome is undesirable, this prediction task has not been given much attention until now in the literature. There have been several recent review papers which also classify and discuss the applications of ML algorithms to the prediction problem of CQAs and KPIs, e.g. \cite{baal21} and \cite{wamy22}. The main role and limitations of data analysis methods used as PAT tools for upstream cultivation processes was discussed in detail in \cite{medi14}. The main purpose is to utilise existing datasets, typically collected from a pilot or manufacturing experiments, to identify correlations and comprehend the causes of variation in bioprocesses. Data analytical methods such as ML models are effective tools for gaining insights into cell cultivation operations after batches have been completed. These insights can be used to early predict the CQAs and KPIs of the products in bioreactors in the next runs. However, only off-line measurements of process responses measured over a certain period of time are not sufficient to rapidly predict the impact of CPPs on the cell culture process performance and product quality. Therefore, it is necessary to combine the off-line measurements with real-time values collected from real-time monitoring devices and instruments, e.g., spectroscopes and sensors \citep{redu22}. These issues will be discussed in subsection \ref{monitoring_cell}.

To develop predictive modelling, historical cell culture datasets are usually organised into three types of information matrices \citep{th06}. The first one is the matrix of static controlled process parameters (e.g., set points of pH and dissolved oxygen) and other parameters such as cell age, which are kept unchanged during the cell culture process. This is a two-dimensional array, referred to as \textbf{Z}, where its rows contain the various experiments and columns are different variables of interest. The second matrix (called \textbf{X}) consists of dynamic process variables such as viable cell density (VCD), nutrient and metabolite concentrations (e.g., glucose, glutamine, lactate, ammonium) measured periodically (e.g., daily). Unlike the \textbf{Z} matrix, process variables in the \textbf{X} matrix are not controlled and thus they reflect the dynamic changes of the cell cultivation process over time. Hence, to facilitate the analysis, the three-dimensional \textbf{X} matrix is usually unfolded batch-wise into a two-dimensional matrix, where the rows represent the different batches and the columns are values of different process variables taken at different time points (i.e., variables of the same batch at two different time points are represented by two separate columns). The last information matrix, denoted as \textbf{Y}, contains the values of the product quality variables, e.g., product titer, aggregates, and charge variants. The \textbf{X} and \textbf{Z} matrices are used as inputs to learning algorithms, while the \textbf{Y} matrix is the targeted outcome of predictors.

\begin{center}
\scriptsize{
\begin{longtable}[t]{cC{2cm}C{1.5cm}C{1.2cm}L{3.4cm}L{2.95cm}L{2cm}L{2.2cm}}
    \caption{Summarising typical applications of ML models to predicting KPIs and CQAs of the products at the bioreactors} \label{cell_forecasting} \\
    \toprule
       \textbf{ID} & \textbf{Scale} & \textbf{Type} & \textbf{Cell} & \textbf{Input} & \textbf{Output} & \textbf{Algorithm}  & \textbf{Reference}  \\
      \midrule
      1 & 5L to 500L bioreactors & fed-batch & CHO & Gln, K, Osm, VCD, Temperature, EGN, Lac, Glu, ACV, HCO\textsubscript{3}, Na, CPDL, ACC, pH, NH\textsubscript{3}, pO\textsubscript{2}, Glc & Product concentration & A two dimensional multi-model system using SVR & \cite{gase21}\\
      \midrule
      2 & T25 cell culture flasks & fed-batch & L1 & Cell seeding density, media supplement percentage, media exchange volume during routine feeding, and cell line & cell doublings (computed from total viable cells at havest and seed) & ANN and SLSR & \cite{rojo21}\\
      \midrule
      3 & 3.5L bioreactor & fed-batch & CHO & pH\textsubscript{set}, DO\textsubscript{set}, amplification duration, amplification VCD, amplification cell age, VCD, Viability, Glc, Lac, Gln, Glu, NH\textsubscript{4}, Osm, pH, pCO\textsubscript{2} , DO & Titer & Combined DT and PLS & \cite{naso19} \\
      \midrule
      4 & AMBR 15ml & fed-batch & N/A & Clone type, Mn, galactose, Fe, CuZn, Asn, cobalt, spermine, gallic acid, sucrose, temperature after shift, feed regime, nitrogen flow, VCD, Viability, Glc, Lac, NH\textsubscript{4}, pH, pCO\textsubscript{2}, DO & Titer, low-molecular weight, neutral charge variant (C3) & Combined DT and PLSR & \cite{naso19}\\
      \midrule
      5 & 3L bioreactor & fed-batch & N/A & pH, DO, temperature, pCO\textsubscript{2}, Glc, Gln, Glu, Lac, NH\textsubscript{4}, Na, K, Ca, Osm, TCD, VCD & Titer & Adaptive model selection algorithm of SVM, GPR, PLSR, regression trees, and ensemble trees & \cite{bawa18} \\
      \midrule
      6 & 0.5ml DWP, 10-15ml AMBR, 3L, 280L bioreactors & fed-batch & CHO & Temperature, pH, pCO2, DO, VCD, viability, Glc, Lac, NH\textsubscript{4} & Titer, LMW, charge variants, glycoforms & combined PCA, DT, PLSR, and Genetic Algorithms & \cite{somo18}\\
      \midrule
      7 & 15ml AMBR & fed-batch & N/A & 9 supplemented media factors (e.g., Manganese, Galactose, Iron, Copper zinc, etc), temperature shift, nitrogen flow, feeding regime, pH, DO, pCO\textsubscript{2}, VCD, Viability, Glc, Lac, NH\textsubscript{4} & Titer, glycoforms, charge variants, aggregates, LMW forms & PCA, PLSR, Genetic algorithms & \cite{sori17}\\
      \midrule
      8 & 5L bioreactors & fed-batch & CHO & pH, temperature, DO, pCO\textsubscript{2}, Glu, Gln, Lac, Glc, NH\textsubscript{4}, VCD, Viability, Osm & Product concentration, charge variants, glycan variants, product monomers, aggregated product & PLSR, RF, RBF-ANN, SVR & \cite{scpo15}\\
      \midrule
      9 & 3.5L stirred tank reactors & fed-batch & CHO & DO, pH, pCO\textsubscript{2}, stress, Osm, Glc, Lac, NH\textsubscript{4}, Gln, Glu, VCD, Viability & Titer & PCA, PLSR & \cite{soso15} \\
      \midrule
      10 & 2000L bioreactors & fed-batch & N/A & 20 process parameters & G0 product quality, final titer, DNA, HCP & PLS, Elastic net with Monte Carlo & \cite{seva15}\\
      \midrule
      11 & shaking 96-DWP & fed-batch & CHO & 47 components of media (e.g., amino acids, sodium selenite, Vitamin B12, cupric sulfate, etc.) & Titer, VCD, Viability, PDL, IVC & PLSR, ANOVA & \cite{rope13} \\
      \midrule
      12 & 80L, 400L, 2000L, and 12000L bioreactors & batch and fed-bach & CHO & 35 online, offline and derived parameters (e.g., pH, DO, pCO\textsubscript{2}, Osm, VCD, Viability, Glc, Lac, Na, NH\textsubscript{4}, air sparge rate) & Product concentration, final Lac concentration & PLSR and SVR & \cite{leka12} \\
      \midrule
      13 & T-flask, 2L and 50L wage bags, 30L and 60L bioreactors, 3.5L production bioreactor & fed-batch & CHO & VCD at seeding, pH, DO, and cell culture duration & Titer, bioactivity, product-related impurities (HMW species, clipped forms), levels of different product variants (glycoforms, oxidized forms), process-related impurities (HCP, DNA) & Regression models and ANOVA & \cite{roso12} \\
      \midrule
      14 & 5L bioreactor & fed-batch & Murine hybridoma & amino acids, glucose, glutamine, and oxygen consumptions & cell growth rate and Lac, NH\textsubscript{3}, and mAb productivity & PCA, PLSR & \cite{seki10}\\
      \midrule
      15 & 80L, 400L, 2000L, 12000L bioreactors & fed-batch & CHO & Off-line physical and chemical parameters (e.g., pCO\textsubscript{2}, DO, pH, Lac, Glc, Na, NH\textsubscript{4}, Osm, VCD, Viability, etc.), online control parameters (e.g., DO, pH, temperature, air sparge rate, etc.) & Titer & SVR & \cite{chle10}\\
      \bottomrule
       \multicolumn{8}{L{17cm}}{\scriptsize \textbf{ACC}: Average Cell Compactness, \textbf{ACV}: Average cell volume, \textbf{AMBR}: Advanced micro-bioreactor, \textbf{ANN}: Artificial neural network, \textbf{ANOVA}: Analysis of variance, \textbf{Ca}: Calcium, \textbf{CHO}: Chinese hamster ovary, \textbf{CPDL}: Cumulative population doubling level, \textbf{DT}: Decision trees, \textbf{DWP}: Deepwell plates, \textbf{EGN}: Elapsed generation number, \textbf{Glc}: Glucose, \textbf{Gln}: Glutamine, \textbf{Glu}: Glutamate, \textbf{HCO}\textsubscript{3}: bicarbonate, \textbf{HCP}: Host Cell Proteins, \textbf{HMW}: High molecular weight, \textbf{IVC}: Integral viable cell density, \textbf{K}: potassium, \textbf{Lac}: Lactate, \textbf{LMW}: Low molecular weight, \textbf{Na}: Sodium, \textbf{NH}\textsubscript{4}: Ammonium, \textbf{Osm}: Osmolality, \textbf{PCA}: Principal component analysis, \textbf{PDL}: Population doubling level, \textbf{PLSR}: Partial least square regression, \textbf{RBF-ANN}: Radial basis function neural network, \textbf{RF}: Random forest, \textbf{SLSR}: Standard least squares regression, \textbf{SVR}: Support vector regression, \textbf{TCD}: Total cell density, \textbf{VCD}: Viable cell density}
\end{longtable}
}
\end{center}

One of the main factors affecting the performance of ML models for early prediction of CQAs and KPIs of the final product based on the historical data of CPPs is the small number and limited range of training data, also known as the `Low-N' issue. For new biopharmaceutical products, there are limited number of bioreactor runs because of the high cost. Each cell culture experiment usually runs over a two weeks period, where off-line measurements such as nutrient and metabolite concentrations are daily measured. In practice, there is often a limited number of historic experiments available for training learning models, and it is common that the number of samples is smaller than the number of considered variables \citep{tuga18}. As a result, the ML models built from these datasets may be at risk of overfitting \citep{vefr05}, especially when the ratio of a number of variables to samples is high. To cope with this issue, \cite{tuga18, tuga18a} proposed to use probablistic models such as Gaussian process models to generate synthetic data based on available small data sets. The proposed method assumes that the initial small samples are representative of the entire process, but this assumption may be incorrect. \cite{seva15} proposed to use the regularisation method such as the elastic net in combination with repeated sampling techniques (i.e., Monte Carlo sampling) and validation approaches to estimate parameters of the model so that they can accurately capture the distribution of training and validation sets over many iterative times of sampling and validation. Another approach is to use simulators developed from mechanistic and first-principle equations to create an unlimited amount of simulated data, which can be employed to generate and test ML models for process optimisation to accelerate the implementation of ML algorithms while waiting for the availability of experimental data for practical performance verification \citep{godu19}. Another challenge to the performance of predictive models for CQAs and KPIs of the product during the cell culture process is the sparsity of high-dimensional upstream process datasets, which means that each individual variable does not include sufficient information for the construction of ML models. \cite{mosa21} proposed to use rigorous first-principle models to tackle the knowledge gaps rather than using data augmentation approaches. The hybrid models of mechanistic and data-driven algorithms are also effective ways to deal with this sparsity issue \citep{nabo22, tsva21}.

\subsection{Understanding of upstream process data}
In addition to early prediction of KPIs and CQAs of the final product, understanding historical manufacturing data of cell cultivation is crucial for the development, optimisation, and enhancement of efficiency and robustness of biological production processes and safety of the final products \citep{thsp10}. The analysis of upstream bioprocess data can provide better insights to enrich the process knowledge, enhance the process understanding, visibility and agility of the process development, ultimately leading to increased productivity \citep{pogu22}. By analysing the data of cell cultivation, one can explore the interactions among process parameters, identify key variables affecting process performance and product quality, and detect the CPP or a set of CPPs responsible for the deviations of a given CQA \citep{bosu18}. These insights are crucial for robust process scale-up and meeting the requirements of quality by design \citep{brfr17}. For example, by analysing the upstream process data, \cite{gahi11} realised that when the glucose concentration in a culture falls below a threshold (generally below 1 mM), cells start absorbing lactic acid from the medium, leading to an increase in pH. Hence, the authors proposed a control strategy of lactate accumulation during the cultivation by triggering a nutrient feeding procedure with a concentrated glucose solution based on the increase of pH values. Additionally, knowledge extracted from upstream process data can be used to detect and diagnose any issues which may arise during the next cell culture processes as well as identifying risks involved in the manufacturing process and how best to cope with those risks. As argued in \cite{thsp10}, a reliable risk assessment can only be effectively conducted when a thorough understanding of the impact of changes on the process or product quality is achieved. Different data analysis techniques and ML models have a great potential to assist in analysing and extracting useful knowledge from upstream process data. This subsection will review typical applications of data analytics to this theme.

The data generated during biopharmaceutical manufacturing is complex and multi-faceted, relying heavily on the relationships between different process parameters. As a result, univariate or bivariate analysis alone can be ineffective in understanding these complexities and may result in inaccurate conclusions \citep{ko04}. Therefore, it is crucial to take into account the relationships between multiple process parameters to gain a comprehensive understanding of the bioprocess and avoid missing important information. As a result, multivariate data analysis (MVDA) using projection methods such as principal component analysis (PCA) and partial least squares (PLS) regression is frequently applied to analysing and extracting valuable insights from the data of cell culture processes in the literature \citep{thsp10}. Projection methods can effectively address the challenges presented by experimental data sets such as multidimensionality, missing values, and variation caused by factors like experimental error and noise \citep{mamo02}. PCA can be used to identify patterns and trends in process data, such as clusters and outliers, for process diagnosis, and subsequently, PLS can be employed to explore the relationships between process parameters and product quality attributes.

\cite{kico07} examined the feasibility of various MVDA diagnostic plots derived from principal components of PCA and PLSR models in evaluating scale-up and comparability of the cell culture process, detecting outliers, and identifing critical indicators affecting the quality attributes and performance of prediction models. The authors indicated that the score plot is appropriate for determining clusters, trends, outliers and similarity within the dataset, while the batch control chart is suitable for detecting outlying samples with regard to control limits. The loading plot and the variable importance for the projection plot can be used to identify the variables which have the most influence on the performance of the cell culture process. Similar approaches have been used by \cite{kigr08} for identifying the underlying causes of variations in scale-up experiments (i.e., small scale, pilot-scale and commercial-scale runs) and exploring interactions among parameters that negatively affect the performance of cell culture process and product attributes. These visual analytics approaches have also been used in many later studies to analyse and understand the upstream process data such as \cite{thsp10, medi13, rasi15, sori17, duhe18, bosu18} and \cite{rile20}.

The latent variable multivariate analysis methods have been widely used to analyse bioprocess data in both industry and academia to generate process knowledge \citep{zuso20}. As mentioned in \cite{paca14}, these MVDA methods are not easy to use for real-time monitoring applications, but they still play important roles in extracting insights to understand, optimise, and control the upstream processes. These methods can be used to uncover the controllability of CQAs and examine the relationship between CPPs and CQAs. Additionally, the knowledge extracted from process data can be employed to establish the optimal medium composition, feeding strategies, and harvest times, as well as define the limits of a process failure and determine an optimal and robust design space for the next experiments.

\cite{medi13} illustrated the use of PCA as a quick and systematic method to reveal the technical procedures affecting the process behaviour. Particularly, it can identify the previously unknown impact of changes in preparation procedures such as medium preparation, probes calibration, and bioreactor inoculation on the deviations between different batches. The authors showed that the process behavior was significantly affected by variations in the concentrations of additives in the feed medium during cultivation, an inoculation at twice the intended cell density, a change in the medium preparation procedure, and a deviation in the calibration of the dissolved oxygen probe. This knowledge was generated by analysis of historical cell culture datasets.

The latent variable multivariate analysis approaches can be used to identify and understand the interrelationship of process and product quality variables \citep{sori17}. \cite{aldu07} used the PCA and PLSR to analyse the exponential phase fluxes in batch culture aiming to identify factors which can increase the monoclonal antibody yield. The study determined and classified amino acids into different groups which have positive, negative, low, and negligible correlation to mAb production, cellular growth, glucose consumption, and ammonia and lactate production. \cite{zuso20} presented the use of PCA and PLSR with the analysis of variable importance in the projection to identify important amino acids and metabolites among hundreds of intracellular and extracellular variables which make impacts on the dynamic glycosylation behaviors of the expressed mAbs. \cite{zako15} applied metabolic flux analysis in combination with PLSR using analyses of variable importance in the projection to investigate the impact of pH and temperature shift timing on the lactate metabolism and cell growth during the fed-batch process of the CHO culture. The analytical outcomes indicated that the onset of lactate consumption was significantly affected by the timing of a pH shift, while temperature shift only had a significant effect when combined with a pH shift. Based on the finding, the authors proposed that a control strategy targeting an early shift to lactate uptake must perform the changes in both temperature and pH set points. These studies confirmed that the comprehensive understanding of the influence of process parameters on metabolism and product quality can be used for the development of strategies to control metabolism aiming to guide cell physiology towards a high producing phenotype \citep{zako15}.

Additionally, understanding the consumption of various nutrients for various cell activities, such as growth, metabolism, and protein production, is crucial for the design of optimal medium components and processes. By analysing the cell culture data, \cite{fava14} showed that the balance of glucose and amino acid concentration in the culture is vital for the cell growth, IgG titer, and N-glycosylation. An unbalanced supply of nutrients for cell activities results in the accumulation of by-products such as ammonia and lactate, which inhibit the cell growth. Similarly, \citep{seki10} illustrated the use of PCA and PLSR methods to identify key medium nutrient components, such as amino acids and carbon sources and their impact on cell growth, mAb synthesis, and generation of inhibitory by-products (e.g., lactate and ammonia). Based on this information, the concentrations of essential amino acids in the feed medium can be adjusted to enhance cell viability and productivity as well as reducing toxic waste production. As a result, multivariate statistical analysis approaches can be utilised as a tool to develop rational medium design strategies \citep{seki10}. In another study, \cite{rasi15} showed how PLS analysis together with MVDA diagnostic plots were used to assess the effect of media supplements, feeding strategies, and process conditions on the product quality attribute such as cell growth, productivity, and mAb glycosylation pattern, and then this knowledge was applied to process optimisation and medium design. One key finding indicated that threonine was a key component having a significant impact on the yield and glycosylation profile of the resulting mAbs generated from the murine hybridoma culture. Moreover, the sparging of culture could change the overall nutrient uptake and metabolism dynamics. \cite{rope13} also presented the usefulness of PLSR in identifying crucial media components that can be examined more closely, potentially leading to the generation of new media formulations with better performance. Apart from PLS modelling, \cite{zozh20} proposed a novel non-parametric regression with Gaussian kernel method to investigate the impact of media components on the mAb titer. The experimental outcomes indicated that the proposed method can identify the key components that were overlooked by the traditional PLS model.

Having knowledge of how different process parameters interact with each other is particularly beneficial when scaling up a process of cell cultivation, where the changes in pH, DO, and pCO\textsubscript{2} are more likely to take place \citep{xike09}. \cite{brfr17} conducted a thorough study of the interactions among key parameters such as pH, DO, and pCO\textsubscript{2}, that are relevant for the scale-up, on the physiology of CHO cells, process performance and CQAs. The analytical outcomes indicated that the variations in pH caused by CO\textsubscript{2} accumulation or the addition of base during large-scale operations are likely to be the primary factor influencing cell growth, glucose uptake, lactate production, and amino acid metabolism. The changes in DO and pCO2 had an effect on both cell growth and specific productivity. The study also showed that the interactions between pH and pCO\textsubscript{2} had also impact on cell growth and productivity. Additionally, several effects of interactions between DO and pCO\textsubscript{2} with pH on amino acid metabolism, as well as interactions between pCO\textsubscript{2} and pH on mAb charge variants and N-glycosylation variants were discovered. The analysis of N-glycosylation revealed a positive correlation between mAb sialylation and high pH values and a relationship between high mannose variants and the pH of the process. However, there are no correlations of mAb aggregation and fragmentation with pH, DO, or pCO\textsubscript{2}. Based on the obtained results, the authors proposed to consider the interactions of the scale-up relevant process parameters on cell physiology, overall performance and product quality when scaling up cell cultures apart from assessing individual parameter impacts.

\subsection{Handling missing information in upstream related datasets}
The collection of bioprocess data over time and across various projects introduces technical variations and inconsistencies, including changes in instrumentation systems, inconsistent sampling caused by limitations of instruments, and human error. These challenges result in missing or incomplete information in databases. Therefore, it is essential to select an appropriate method for dealing with missing values to refine the accuracy of the predictive models and interpretations performed in the secondary analysis. According to \cite{maga19}, the biologics datasets often have missing patterns for different days of the bioprocess, primarily for parameters monitored by instruments and automatically retrieved. Some missing values are due to offline measurements. As a result, the biologics datasets exhibit a characteristic of missing values at random. Therefore, \cite{maga19} performed an empirical assessment of the effectiveness of mean imputation and multivariate regression in filling the missing values in historical bio-manufacturing datasets. The performance of these methods was evaluated using symbolic regression models and Bayesian non-parametric models in data processing. Mean imputation was found to be a straightforward and effective approach for quite smooth, non-dynamic datasets. Whereas, regression imputation was effective in preserving the standard deviation and distribution shape of dynamic datasets with less than 30\% missing data.

According to \cite{gase21}, in addition to the missing at random characteristic, upstream bioprocess development and production datasets are characterised by high heterogeneity due to inter-batch variability, varying production protocols, and diverse data collection methods. They typically comprise a mix of time-dependent and time-independent parameters. Time-dependent parameters represent temporal changes over time, while time-independent parameters do not exhibit predictable fluctuations. Hence, data imputation methods should take into account the multivariate time-series characteristic of bioprocess data. As a result, the authors developed a new dual-hybrid methodology to handle missing information in bioprocess datasets. It uses a combination of Fuzzy C-Means clustering, Support Vector Regression (SVR), and Genetic Algorithm (GA) to address time-independent parameters and a combination of Stineman interpolation and SVR to handle time-dependent parameters. The dual-hybrid methodology showed acceptable performance without relying on prior knowledge of the dataset or the physical and biological meaning of the parameters. The method can be applied for any datasets with heterogeneous time-dependency observed in the parameters.

\subsection{Cross-scale prediction of nutrient, metabolite, and cell growth parameters and knowledge transferability within bioreactors}
Developing effective cell culture processes is intricately complicated and requires numerous experiments of varying scales to determine the final process design space and meet regulatory standards \citep{somo18}. Therefore, the bioprocess development stage is usually performed in small-scale bioreactors to understand the complex dynamic bioprocess \citep{nalu19}. Small-scale experiments are cost-effective in terms of time (due to the potential for parallelization) and expenses (as a result of reduced material and labor costs) \citep{somo18}. For a given cost, testing with smaller bioreactors allows for evaluating more parameters, but larger bioreactors tend to more closely mimic commercial production processes \citep{bemo14}. Hence, it is essential to have a balance between flexibility and reliability. As argued in \cite{necr13}, it is crucial to take the industrial scale into account from the early stages of bioprocess development because variations in physical and chemical parameters between small and industrial scales can result in changes in the final outcome of the process. \cite{badu21} demonstrated distinct trends for key process variables within cell culture processes in a 300ml shaker and a 15L bioreactor. These variations may stem from differences in the bioreactor design or feeding strategies, such as a bolus feed in shakers and continuous linear feeding in bioreactors. Hence, the knowledge transferability among cell culture processes at different scales is not straightforward and still a challenging problem. According to \cite{tela15}, a critical factor for effectively characterising a bioprocess in the pharmaceutical industry is having a representative scale-down model of the manufacturing process. This enables the direct transfer of extensive knowledge gained at a small scale to the full-scale commercial operations \citep{rile20}. A key goal is to conduct scale-up characterisations of processes and technologies, with the aim of predicting commercial manufacturing performance based on data from scaled-down bioreactor models \citep{bemo14}.

Different analytical techniques can be performed in the process scale-up studies to provide the insight into the cell behavior consistency across different scales \citep{rile20}. Although there is a great potential for a comprehensive characterisation of dynamic processes, current approaches like evaluating mass transfer and hydrodynamic parameters with Monod kinetic models and semi-empirical logistic equations, conducting computational fluid dynamic simulations, and metabolic flux analyses may not fully cover all factors that may cause cross-scale inconsistencies, due to the limitations of the deterministic modeling framework \citep{nalu19}. As a result, to achieve a more comprehensive evaluation of the process fingerprint, including design variables such as media components and metabolic profiles, data-driven approaches were applied \citep{rami14}. However, there have been only few studies in the literature aiming to investigate knowledge transferability of data-driven multivariate data analytics and predictive modelling techniques across scales of cell culture processes. This section will review the typical studies in this theme.

\cite{bemo14} conducted experiments to assess the scalability and transferability of multivariate PLS models in real-time predictive modelling based on the Raman and offline analytical data collected at small (3L bioreactors), pilot (200L bioreactors), and manufacturing (2000L bioreactors) scales of the CHO cell culture processes. The outcomes indicated that the PLS models are well scaled up for modelling glucose, lactate, and osmolality based on the performance similarity across single scale and combination scale models. The PLS models for real-time prediction of glutamate and ammonium concentrations which were built from small scale data did not perform well on manufacturing scale batches. The PLS models for VCD and TCD based on Raman spectra exhibited some scale dependencies which can result in the reduction in the performance for cross-scale predictions. It is also worth noting that VCD and TCD measurements from the cell counting instrument are notoriously bad, particularly towards the end of a run, so it can also affect the accuracy of offline measurements among different scales and the transferability of the models.

\cite{ahja15} used multivariate analysis approaches such as PCA, PLS, orthogonal partial least squares, and discriminant analysis to analyse cell culture data from shake flasks, 3L and 15000L bioreactors. The obtained results showed that the predictive model constructed from the data of shake flasks was a better representation of the large-scale predictive performance in 15000L bioreactors (using one-sided pH control) than the 3L bioreactor modelling. The analysis showed that the differences in performance between the large and small scale models were due to pCO2 and pH values. By reducing the initial sparge rate, the performance of models built from the 3-L bioreactor data moved closer to that of the large scale. The results also confirmed that multivariate analysis is an efficient tool for assessing the goodness of a scale-down model, identifying input parameters responsible for performance differences across different scales, and developing hypotheses for refining the scale-down model which can then be applied for quality prediction of products at the manufacturing scale. In another study, \cite{somo18} built predictive models using PCA, decision trees, and PLS combined with genetic algorithms to estimate the mAb titer, charge variants and glycoform profiles from CHO-derived cell culture data collected from different scales, i.e., 96-deep-well plates (0.5ml working volume), ambr15 microbioreactors (10-15ml working volume), lab scale (3-3.5L working volume), and pilot scale (maximum 280L working volume).  The obtained results confirmed that the process knowledge generated from small-scale experiments, e.g., process variables with high importance, can be used to develop effective predictive models for large-scale experiments. In recent study, \cite{badu21} illustrated that a hybrid model combining ANNs with mechanistic equations trained on CHO cell culture data including CPPs and amino acid consumption patterns in 300ml bolus feeding shake flasks could be deployed to accurately predict mAb titer and viable cell concentration in a 15L continuous-feed stirred-tank bioreactor, within the same design space and with a minimal need for recalibration.

\subsection{Real-time monitoring of the nutrient, metabolite, and cell growth parameters}\label{monitoring_cell}

As argued in \cite{brfr17}, variations in the upstream process can result in the changes in the physiochemical, biological, and immunogenic properties of therapeutic proteins, which are intricate and fragile molecules. The manufacturing of biotherapeutics through the cell culture processes usually results in the generation of harmful substances such as lactate and ammonia, negatively impacting cell productivity. Hence, it is highly desirable to maintain a balance between supplying essential metabolites and limiting toxic compound accumulation for a significant enhancement in culture performance \citep{saca96}. In addition, controlling the cell culture's micro-environment within bioreactors is crucial in keeping the product heterogeneity within the set of specifications defined by commercial good manufacturing practices (GMP). To achieve this, it is essential to monitor and control the cell culture's processes using automated sampling techniques and advanced sensing approaches \citep{shbr11, ha13}. The PAT guidance of FDA \citep{shbr11} suggested moving away from a fixed cell culture process that may cause the deviations in product quality, towards an adaptive process that uses sensors and advanced control strategies to maintain consistent product quality. Despite significant advancements, commercial manufacturing still heavily relies on time-consuming offline analytics and manual control strategies \citep{goum20}.

Apart from few real-time measurements such as pH, DO, and temperature using electrochemical and optical sensors, most of the cell culture parameters like metabolite and product concentrations, cell growth, and other CQAs are monitored through offline approaches \citep{sake18}. Typically, these offline methods, which entail removing cells from the bioreactor, are usually time-consuming and labor-intensive, and result in waste due to the use of costly and potentially toxic reagents and samples \citep{tuwa19, weha18}. Moreover, offline measurements are conducted infrequently (e.g., every 12-24 hours), leading to low process monitoring resolution and the possibility of missing metabolic shifts in cells that could signal process changes or operational issues, so establishing control strategies based on offline analytical measurements is often ineffective \citep{tukh21}. In addition, every sample taken from the bioreactor poses a risk of contamination or batch loss \citep{tuwa19}. Hence, it is desirable to develop inline/online monitoring technologies of important bioreactor variables to support effective in-process control strategies \citep{krho17a}. The deployment of techniques enabling online monitoring of CPPs related to quality or CQAs allows for real-time data analysis and process modeling. This facilitates early detection of faults, allowing for corrective actions to be taken promptly, thereby maximising the chances of both successful process outcomes and obtaining the expected product quality \citep{medi14}. When an undesired deviation in a process is detected based on the outcomes of online monitoring, a control system can be triggered to adjust the process parameters and keep them within their desired range to maintain consistent product quality \citep{paca14}, e.g, performing a bolus feeding process based on real-time glucose concentration \citep{dofr20}.

\begin{center}
\scriptsize{
\begin{longtable}[t]{cC{1.8cm}C{1.3cm}C{1cm}C{1.8cm}L{2.6cm}L{1.8cm}L{2.2cm}L{2.2cm}}
    \caption{Summarising typical applications of ML models in monitoring process parameters in bioreactors} \label{table_cell_monitoring} \\
    \toprule
       \textbf{ID} & \textbf{Scale} & \textbf{Type} & \textbf{Cell} & \textbf{Offline sampling} & \textbf{Monitoring parameters} & \textbf{PAT Tool} & \textbf{Algorithm}  & \textbf{Reference}  \\
      \midrule
      1 & 2-12000L bioreactors & fed-batch & CHO & N/A & Glc, Lac, NH\textsubscript{4}, Na, amino acids, mAb titer, Osm, TCC, TCV, VCC, VCV & Raman & 18 algorithms including ANNs, tree-based models, linear models, SVR, AutoGluon, AutoKeras & \cite{poma22}\\
      \midrule
      2 & N/A & fed-batch & CHO & 1-2 samples/day & Glc, Lac & Raman & Spatio-temporal JITL GP & \cite{tukh21}\\
      \midrule
      3 & AMBR 15ml & fed-batch & CHO & every two days & Glc, Lac, VCD, mAb concentrations & Raman & PLSR & \cite{goum20}\\
      \midrule
      4 & 5L bioreactor & perfusion & CHO & every 24-36 h & etanercept concentration & non-linear fluorescence & MLP & \cite{chte20}\\
      \midrule
      5 & 4L bioreactor & batch, fed-batch & CHO DG44 & every 4 hours to once a day & amino acids & mass spectrometry & PCA & \cite{powa19}\\
      \midrule
      6 & N/A & batch, fed-batch, perfusion & modified CHO & 1-2 samples/day & Glc, Glu, Gln, Lac, Na, K, NH\textsubscript{4}, VCD, Viability & Raman & Real-time JITL GP & \cite{tuwa19}\\
      \midrule
      7 & N/A & batch, fed-batch, perfusion & modified CHO & 1-2 samples/day & Glc, Glu, Gln, Lac, Na, NH\textsubscript{4}, Ca, VCD, Viability & Raman & JITL PLSR, JITL GP & \cite{tusc19}\\
      \midrule
      8 & 5L bioreactor & fed-batch & CHO & once per day & pH & Raman & PLSR & \cite{raot19}\\
      \midrule
      9 & 2L, 7L, 15L, 10000L bioreactors & fed-batch, perfusion & CHO & once per day & Glc, Lac, Protein titer & Raman & PCA, PLSR & \cite{saka19}\\
      \midrule
      10 & 5 L glass bioreactors & fed-batch & CHO DG44 & twice a day (before and after adding daily nutrient feed) & Glc concentration & Raman & PLSR & \cite{masm18}\\
      \midrule
      11 & 5L, 10L bioreactors & fed-batch & CHOK1SV GS-KO & twice a day & Glc, Glu, Lac, NH\textsubscript{4}, mAb concentrations, VCC, TCC & Raman & PLSR & \cite{weha18}\\
      \midrule
      12 & 2L, 30L, 300L bioreactors & fed-batch & E. coli & 30 min & Amino acids, B vitamin, Glc, and optical density & NIR & PCA and PLSR & \cite{vash17}\\
      \midrule
      13 & 7L bioreactor & fed-batch & CHO (GS) & 14 day & Abnormal batches & Raman & MPCA with COW &  \cite{lian17}\\
      \midrule
      14 & 3L bioreactor & fed-batch & CHO & once per day & tyrosine, tryptophan, phenylalanine and methionine & Raman & PLSR & \cite{bhme17}\\
      \midrule
      15 & 4L bioreactor & fed-batch & CHO (GS) & 5-6 times/day & mAb concentration & Raman & PLSR & \cite{ancr15}\\
      \midrule
      16 & 3L, 5L, and 500L bioreactors & fed-batch & CHO & every 30 min & Glc, Lac, VCD & Raman & PLSR & \cite{mela15}\\
      \midrule
      17 & flask & batch & CHO & once per day & Glc, Lac & Raman & CLS & \cite{sigo15}\\
      \midrule
      18 & 15L bioreactor & fed-batch & CHO 320 & once per day & Glc, Gln, Lac, NH\textsubscript{4} & Raman & PLSR and Monod-type kinetics & \cite{crwh14}\\
      \midrule
      19 & 2L, 100L, 200L, 1000L, 5000L bioreactors & fed-batch & CHO & 12 times/run & glycoprotein yield & Raman & PLSR with CoAdReS and ACO & \cite{lira13}\\
      \midrule
      20 & 500ml shake flasks & fed-batch & CHO & N/A & mAb fermentation media quality & NIR, 2D-Fluorescence & PLSR, PCA & \cite{hast13}\\
      \midrule
      21 & 3L, 15L bioreactors & fed-batch & CHO 320 & once per day & Glc, Gln, Glu, Lac, NH\textsubscript{4}, TCD, VCD & Raman & PLSR & \cite{whcr12}\\
      \midrule
      22 & 500L bioreactor & fed-batch & CHO DG44 & once per day & Glc, Gln, Glu, Lac, NH\textsubscript{4}, VCD, TCD & Raman & PLSR & \cite{abke10}\\
      \midrule
      23 & 7L bioreactor & fed-batch & B. pertussis & every hour & biomass concentration & NIR & PLSR & \cite{sost08}\\
      \midrule
      24 & 2.5L stirred tank bioreactor & fed-batch & E. coli & 30 min to a few hours & Glc, Lac, Phe, HCO2, Acetate concentrations & Raman & least-squares regression & \cite{lebo04}\\
      \bottomrule
      \multicolumn{9}{L{18cm}}{\scriptsize \textbf{ACO}: Ant colony optimisation, \textbf{AMBR}: Advanced micro-bioreactor, \textbf{ANN}: Artificial neural network, \textbf{CHO}: Chinese hamster ovary, \textbf{CLS}: Classical least squares, \textbf{COW} : Correlation optimised warping, \textbf{CoAdReS}: Competitive adaptive reweighted sampling, \textbf{Glc}: Glucose, \textbf{Gln}: Glutamine, \textbf{Glu}: Glutamate, \textbf{GP}: Gaussian Process, \textbf{HCO2}: formate, \textbf{JITL}: Just-in-time-learning, \textbf{K}: potassium, \textbf{Lac}: Lactate, \textbf{MLP}: Multilayer perceptron artificial neural network, \textbf{MPCA}: Multi-way principal component analyses, \textbf{Na}: Sodium, \textbf{NH}\textsubscript{4}: Ammonium, \textbf{NIR}: Near‑infrared spectroscopy, \textbf{NMPC}: Non-linear model predictive controller, \textbf{Osm}: Osmolality, \textbf{PCA}: Principal component analysis, \textbf{Phe}: Phenylalanine, \textbf{PLSR}: Partial least square regression, \textbf{RF}: Random forest, \textbf{SVR}: Support vector regression, \textbf{TCC}: total cell concentration, \textbf{TCD}: Total cell density, \textbf{TCV}: total cell volume, \textbf{VCC}: viable cell concentration, \textbf{VCD}: Viable cell density, \textbf{VCV}: viable cell volume}
\end{longtable}
}
\end{center}

Because of the need for quick, ongoing measurements and the cost savings associated with several analytical techniques, soft-sensors have been used as a cost-effective solution for real-time monitoring in process industries \citep{kaga09}. A soft-sensor is a computational algorithm that estimates specific state variables that cannot be obtained promptly, or is used to predict future values of the process variables \citep{gust19}. The first applications of soft-sensors in bioprocessing were for real-time estimations of process variables, including biomass, product and substrate concentrations, and their dynamically changing rates as reviewed in \cite{rama17}. The advancement of soft-sensors for the real-time monitoring of biopharmaceutical processes is concurrent with the growth of real-time PAT sensors, such as real-time spectroscopic methods and deterministic models \citep{lubr12}. Several inline spectroscopic techniques including Raman, Near Infrared (NIR), two-dimensional (2D) Fluorescence have been developed for continuous cell culture monitoring \citep{weha18}. The historical spectroscopic measurements combined with advanced data analysis methods and signal processing techniques have been used to build soft-sensors for real-time monitoring of CPPs, KPIs, and CQAs of an upstream cell culture process. Table \ref{table_cell_monitoring} summarises typical applications of ML models together with real-time PAT sensors in monitoring upstream process parameters. There have been other review papers presenting the use of ML models in combination with PAT tools to construct real-time monitoring solutions for the various process parameters and product's CQAs, e.g., \cite{paca14, bury17, krho17a, rama17, papa21, escu21, redu22}.

It can be observed that the Raman spectroscopy combined with PLSR models have been the leading approaches for real-time monitoring of critical cell culture performance parameters including amino acids, glucose, glutamate, glutamine, lactate, ammonia, VCD, and product concentration. As explained in \cite{goum20}, Raman spectroscopy has become increasingly popular in recent years as it is well-suited for analysing aqueous samples because of its low water interference and high specificity compared to other spectroscopic techniques. Additionally, Raman spectra offer both qualitative and quantitative information, including details about the composition, chemical environment, and structure of samples. Spectroscopic data usually contains thousands of variables (wave numbers) and measurements (observations), so as to extract useful information and find the relationship between spectroscopic signals and the process parameters of particular interest and it requires multivariate analysis approaches to be used. The latent variable multivariate analysis methods such as PLSR and PCA have been widely used in analysing spectroscopic data due to their capability to analyse extensive spectral distribution and effectively distinguish between spectra of samples that only have slight variations \citep{ocry10}. Furthermore, by analysing the dominant regression coefficients of PLSR models, we can identify and assess which characteristic peaks associated with particular wavenumbers have significant impacts on the capabilities of detecting the variations in the variable of interest \citep{bhme17, goum20}. Despite the popularity of PLSR models in analysing Raman spectra, a comprehensive experiment in recent study \citep{poma22} showed that neural networks and random forest regression outperformed PLSR in real-time prediction performance of CPPs and KPIs during the upstream cell culture process. However, one of the existing problems in the studies regarding the development of Raman-based real-time predictive models is the method of evaluating their performance. Typically, the number of data points used for training and testing the models ranges from tens to possibly low hundreds of offline measurements. Therefore, careful consideration must be taken when selecting the training and testing datasets, and there is likely to be a significant amount of stochasticity when assessing the goodness of fit related to the train-test datasets splitting. Furthermore, there does not seem to be much consideration of the `noisiness' of predictions between offline time points. Instead, most studies simply report several statistical error metrics, such as root mean squared error, mean absolute percentage error, or coefficient of determination ($R^2$), based on the limited number of offline test points. These issues need to be addressed in order to develop effective Raman-based real-time predictive models in practical applications.

The main drawback in using Raman spectroscopy for monitoring of cell culture process parameters is the impact of the strong intrinsic fluorescence from many biomolecules and background signals \citep{gava15}, making it difficult to obtain accurate results. This is because the weak Raman signals are often overpowered by the much more intense fluorescence background. Moreover, the Raman spectra as well as other spectroscopic signals are often impacted by noise from the detector and fluctuations in the intensity of the radiation source. Hence, we need to perform pre-processing steps to eliminate the unwanted signals, including fluorescence, Mie scattering, calibration errors, cosmic rays, detector noise, laser power fluctuations, and signals from cell media or glass substrate, from Raman spectra before extracting the information from them \citep{bowa11}. The comprehensive pre-processing techniques for Raman and infrared spectroscopy, such as Cosmic ray removal, background correction, smoothing, and outlier removal can be found in \cite{gava15}. As shown in the empirical outcomes in \cite{poma22}, the pre-processing techniques of Raman data have significantly impacted on the performance of all considered ML models.

As confirmed in \cite{baal21} after reviewing 60 journal articles regarding the PAT applications, the authors concluded that the use of PAT within the biopharmaceutical manufacturing is still in the early development stages as most of the studies focus on the process monitoring, while only few studies presents the demonstrations of control associated with monitoring activities. To build a comprehensive soft-sensor system for online monitoring and control of the cell culture manufacturing process, it is essential to combine information from real-time process control measurements of pH, DO, temperatures, pressure, etc. with online measurements collected from analytical technology such as spectroscopy of nutrient, metabolites, and product concentrations and cell growth rates \citep{th06}.

\subsection{The combination of mechanistic models with data-driven models for process monitoring and control}\label{hybrid_upstream}
Models for monitoring CPPs and predicting relationships between process parameters and process outputs or product CQAs can be developed based purely on knowledge or data. Data-driven models are preferred when little understanding of the input-output relationship exists and rely on statistical methods or machine learning algorithms to determine relationships from data. In contrast, knowledge-based models like mechanistic models are usually used when a clear understanding of the relationship between inputs and outputs is available \citep{sohi17}.

In industry, the data-driven modelling approach is prevalent due to its systematic nature and clear results, as well as its connection to the QbD paradigm \citep{sost21}. Over the past 20 years, the use of data-driven modelling approaches including statistical models via multivariate data analysis or ML algorithms for bioprocess data has steadily increased \citep{gugl18}. There have been many typical applications reported using data-driven modelling approaches as presented in this Section \ref{upstream} and \citep{gugl18} such as real-time estimation of CPPs/CQAs (soft sensors), extraction of knowledge in multidimensional spectra data, prediction of the KPIs and CQAs based on CPPs data, upstream process data understanding using MVDA, process control, and dynamic upstream process simulations. However, this approach has several main drawbacks. First, the number of experiments (samples) needed to understand the effect of process parameters on the response variables may increase disproportionately or even exponentially as the number of parameters increases, a phenomenon known as the curse of dimensionality \citep{sost21}. The performance of the models is poor when adequate data is not provided \citep{nalu22}. Second, the understanding gained is often specific to the analysed data, which limits its knowledge transferability \citep{sost21}. The data-driven models lack the ability to offer insights regarding the causal relationships between bioprocess inputs and outputs \citep{tsva21}. Additionally, data-driven models are often less robust when built from raw data, and thus they can require extensive data pre-processing.

In contrast, the knowledge-driven models such as mechanistic models can provide a deep understanding of the system and are typically based on differential equations and phenomenological assumptions that have been used and modified over decades. Therefore, this knowledge can be easily transferred between projects as physical laws remain constant across projects \citep{sost21}. These models have been applied in the development of enhanced strategies for DoE, monitoring and control strategies \citep{fami17}. However, the development of mechanistic models demands a high level of understanding and expertise, making the process of developing and using these models challenging. This becomes even more difficult with a large number of process parameters, because attempting to accurately capture interactions between them can make models very complex \citep{sost21}. Moreover, the knowledge-driven modelling lacks sufficient model resolution to predict biopharmaceutical CQAs \citep{mest17} and the ability to represent dynamic (scale and time-dependent) parameters \citep{gugl18}.

Hybrid models can integrate the strengths of both the knowledge-driven and data-driven approaches. This approach balances the understanding and knowledge from the knowledge-driven method with the simplicity of the data-driven approach \citep{sost21}. Hybrid modeling allows for incorporating of existing knowledge while the existing knowledge gaps may be filled through the use of data-driven sub-models. The model's mechanistic part can be continually enhanced through the accumulation of knowledge during operational processes \citep{sohi17}. The advantages and disadvantages of mechanistic, statistical and ML, and hybrid modelling are summarised in Table \ref{pro_con_methods}.

In recent years, hybrid models have been considered as a key enabler towards the development of digital twins \citep{sost21} and achieving the goals of Biopharma 4.0 \citep{tsva21} by allowing model-based control \citep{pako20} and optimisation to increase yield \citep{naso19a, krho17a} and quality \citep{koko20} of biologics. However, the number of applications of hybrid models in monitoring and control of upstream bioprocesses has not received much attention in the literature yet. Typical studies regarding the use of hybrid models for bioprocess monitoring, prediction and control are summarised in Table \ref{hybrid_predict}.

\begin{table}[H]
    \centering
    \scriptsize{
    \caption{The pros and cons of upstream bioprocess modelling (adapted from \cite{tsva21, wamy22})} \label{pro_con_methods}
    \renewcommand{\arraystretch}{0.25}
    \begin{tabular}{L{1.2cm}L{6.5cm}L{7.5cm}}
         \textbf{Method} & \textbf{Pros} & \textbf{Cons} \\
        \midrule
     Mechanistic & \begin{itemize}[noitemsep]
         \item Using mechanistic knowledge (physics, chemistry, biology) leads to improved process understanding
         \item Can be employed for model-based design for optimally informative experiments
         \item Can be applied to control and optimise bioprocesses
     \end{itemize}
     & \begin{itemize}[noitemsep]
         \item Complex and require a deep understanding of the underlying biological mechanisms, especially the dynamic nature of biological systems and the uncertainties associated with the model parameters
         \item Rely on simplifying assumptions which can oversimplify the real-world biological processes
         \item Lack of flexibility
         \item Automating/formalising model assembly is challenging
         \item Deployment within industry is difficult due to the required expertise
     \end{itemize} \\
     \midrule
     MVDA & \begin{itemize}[noitemsep]
         \item Easy to set up with readily available computational tools featuring user-friendly graphical interfaces
         \item Fast to optimise and reduced computational overhead
         \item Models are frequently represented as understandable linear equations
         \item Effective when number of CPPs are small
         \item Useful for data visualisation
         \item Can be deployed for real-time monitoring bioprocesses and fault detection
     \end{itemize} & 
        \begin{itemize}[noitemsep]
            \item Predictive abilities can be restrained to validation space
            \item Linear equations-based algorithms such as PCA/ PLSR can lose information
            \item Inability to model complex relationships between CPPs and CQAs when data is noisy and contains non-linear relationships
            \item Requiring easily accessible, representative, and reliable training datasets
            \item Limited capacity for bioprocess control and optimisation
            \item Data may need preprocessing
     \end{itemize}\\
     \midrule
     Machine learning & \begin{itemize}[noitemsep]
         \item Capable of capturing complex relationships, including non-linear relationships, which results in modelling the underlying process more effectively
         \item Ability to process large datasets obtained from various sources, such as multi-omics, in-situ spectra, and conventional analytical techniques
         \item Feature selection algorithms can uncover new CPPs in high-dimensional data
         \item Able to be effectively deployed for real-time bioprocess monitoring
         \item Automatic model assembly
     \end{itemize} & 
     \begin{itemize}[noitemsep]
         \item Efficient model training typically requires large amounts of data
         \item Optimisation may be slow and require high computational power
         \item Difficult to set up, which increases the likelihood of incorrect design
         \item Data may need preprocessing
     \end{itemize}\\
     \midrule
     Hybrid & \begin{itemize}[noitemsep]
         \item Constrains model outputs to physically feasible values through the use of mechanistic relationships
         \item Automatic model assembly
         \item Able to be used for model-based design of optimally informative experiments
         \item Ability to deploy for bioprocess control and optimisation
     \end{itemize} & 
     \begin{itemize}[noitemsep]
         \item Needs readily available, representative, and reliable training datasets
         \item Demands resource-intensive training procedure to attain sufficient predictive abilities
         \item Data may need preprocessing
     \end{itemize}\\
     \bottomrule
    \end{tabular}
    }
\end{table}

\begin{center}
\scriptsize{
\begin{longtable}[t]{cC{2cm}C{1.5cm}C{1cm}L{3.5cm}L{2cm}L{2cm}L{2.5cm}}
    \caption{Summarising typical applications of hybrid modelling to prediction and monitoring of CPPs, KPIs, and CQAs at the bioreactors} \label{hybrid_predict} \\
    \toprule
       \textbf{ID} & \textbf{Scale} & \textbf{Type} & \textbf{Cell} & \textbf{Input} & \textbf{Output} & \textbf{Algorithm}  & \textbf{Reference}  \\
      \midrule
      1 & 3.5L bioreactor & fed-batch & CHO & Amplification duration, Amplification VCD, Amplification cell age, DO\textsubscript{set}, pH\textsubscript{set}, pH, DO, pCO\textsubscript{2}, Glc, Lac, Glu, Gln, NH\textsubscript{4}, Osm, VCD, Titer & Titer & Monod kinetics + ANN & \cite{nalu22} \\ 
      \midrule
      2 & N/A & fed-batch & mammalian & VCD, Glc, Gln, Lac, NH\textsubscript{4}, Titer & VCD, Glc, Gln, Lac, NH\textsubscript{4}, Titer & GPSSM + polynomial derivatives & \cite{crna22}\\
      \midrule
      3 & 3.5L bioreactor & fed-batch & CHO & Amplification duration, Amplification VCD, Amplification cell age, DO\textsubscript{set}, pH\textsubscript{set}, pH, DO, pCO\textsubscript{2}, Glc, Lac, Glu, Gln, NH\textsubscript{4}, Osm, VCD & VCD, Titer, Glc, Lac & ANNs + mass balance equations + EKF & \cite{nabe20} \\ 
      \midrule
      4 & 20L bioreactor & fed-batch & E. coli & 27 CPP conditions & biomass concentration, soluble product titer & biomass and product equations + ANN & \cite{bast20}\\
       \midrule
      5 & 20L bioreactor & fed-batch & E. coli & 277 CPPs & biomass concentration, product titer & Material balance equations + ANN & \cite{bast20a}\\  
      \midrule
      6 & 3.5L bioreactor & fed-batch & CHO & Amplification duration, Amplification VCD, Amplification cell age, DO\textsubscript{set}, pH\textsubscript{set}, pH, DO, pCO\textsubscript{2}, Glc, Lac, Glu, Gln, NH\textsubscript{4}, Osm, VCD & Titer, VCD, Lac, Glu & ANNs + mass balance equations & \cite{naso19a}\\
       \midrule
      7 & N/A & batch, fed-batch & E. coli & Online measurements including temperature, pH, DO, agitation rate, air-flow rate, vessel pressure, feed and base balance readings & base addition, biomass and product concentrations & biomass and product equations + ANN & \cite{voha16}\\
       \midrule
      8 & 1L bioreactor & fed-bacth & CHO-K1 & Online measurements: pH, DO and temperature, Offline measurement: pH, VCD, Viability, Glc, Glu, Gln, Lac, NH\textsubscript{4}, amino acids & Titer, IVCD, Lac & MFA + PLSR & \cite{zako15}\\
       \midrule
      9 & 500ml shake-flasks & batch & CHO & Fluorescence Spectra; Offline measurements: Cell count, Glc, NH\textsubscript{4}, protein concentrations & Viable and dead cells, Glc, NH\textsubscript{4}, protein concentrations & MFA + EKF + PLSR & \cite{ohle14}\\
       \midrule
      10 & 7L bioreactor & fed-batch & B. pertussis & Online measurements: NIR, pH, DO, temperature, Offline measurements: biomass, Gln, Lac concentrations & biomass, Gln, Lac & PLSR + Macroscopic material balance equations & \cite{stol11}\\
      \bottomrule
       \multicolumn{8}{L{17cm}}{\scriptsize \textbf{ANN}: Artificial neural network, \textbf{CHO}: Chinese hamster ovary, \textbf{DO}: dissolved oxygen, \textbf{EKF}: Extended Kalman filter, \textbf{GPSSM}: Gaussian Process State Spaces Models, \textbf{Glc}: Glucose, \textbf{Gln}: Glutamine, \textbf{Glu}: Glutamate, \textbf{IVCD}: Integral viable cell density, \textbf{Lac}: Lactate, \textbf{MFA}: Metabolic flux analysis, \textbf{NH}\textsubscript{4}: Ammonium, \textbf{NIR}: Near‑infrared spectroscopy, \textbf{Osm}: Osmolality, \textbf{PLSR}: Partial least square regression, \textbf{VCD}: Viable cell density}
\end{longtable}
}
\end{center}

It can be seen that the ANN models are usually combined with mechanistic models in the development of hybrid models, where the ANN models are used to estimate the values for parameters of the mechanistic equations based on the empirical data. \cite{naso19a} introduced a serial hybrid model, where an ANN was deployed to estimate the unknown-specific uptake rates of a bolus feeding within cell culture equations based on mass balances. The hybrid model showed a better accuracy in the prediction of key nutrient and metabolite concentrations as well as product titer than statistical approaches such as PLSR. Additionally, hybrid models also demonstrated better robustness and extrapolation
abilities than statistical models which struggle with time alignment issues and have limitations in predicting beyond their training regions. The extrapolation ability is crucial not only for optimising processes, but also for scaling processes up, transferring knowledge, and making informed decisions in process development. In a later study, \cite{nabe20} combined the hybrid model proposed in \cite{naso19a} with an extended Kalman filter (EKF) for real-time monitoring, control, and automatic decision making in mammalian cell culture processing. The empirical outcomes showed that the hybrid‐EKF model led to an improvement of at least 35\% in predictive monitoring accuracy, compared to industrial benchmark tools based on PLS models. Using an industrial use case, the authors demonstrated that the soft sensor for process monitoring using the hybrid-EKF showed a 50\%
improvement in prediction accuracy of titer in comparison with a benchmark soft sensor tool. The EKF was also used in combination with dynamic mechanistic models and PLSR for real-time monitoring of viable and dead cells, protein, glucose, and ammonia concentrations based on the fluorescence spectroscopic data \citep{ohle14}. In this study, the PLSR was used to identify the components of the fluorescence spectra related to key process variables, and then these outputs were provided as inputs for a mechanistic model to enhance the predictive performance. Combining material balance equations with statistical models within the Kalman filter can create a direct connection between data and physical bioprocess systems. It is because the Kalman filter utilises the predictions from the mechanistic model and the data provided by the data-driven model to continually update the state estimators to synthesize the information from both models. In addition to ANNs, the PLSR can also be used to estimate the specific reactions rates of macroscopic material balance equations, especially in case of a high number of parameters in spectral data and a demand for extracting knowledge from highly correlated data as presented in \cite{stol11}.

\cite{nalu22} demonstrated that hybrid models are a superior choice over purely data-driven or mechanistic models for cell culture applications. The authors proposed that the degree of hybridization (levels of process knowledge embedded) in the hybrid model must be chosen based on the data available and the targeted application. In cell culture applications, where the underlying processes are poorly understood and a large number of process parameters are involved, hybridization approach can be used to test mechanistic hypotheses and increase our understanding of key process parameters and characteristics. The empirical results showed that the performance of the models can be improved as knowledge is added, as long as the knowledge is not biased. The selection of hybrid models will depend on the goal of the model development. For instance, hybrid models that incorporate mass balances for each species are better for transferring models across different operating modes. On the other hand, models with a higher degree of hybridization allow for more process interpretation.

Hybrid modeling, when combined with the design of experiments, has also the potential to speed up process development \citep{bast20a, sohi17, moku19} by reducing the number of required experiments through increased process knowledge, thus potentially lowering product/process development costs. Another potential approach is to use mechanistic models to generate training datasets for statistical or machine learning models. This approach may not improve process understanding, but has potential applications in developing bioreactor soft sensors, where the number of available training data is limited \citep{sohi17}.

\subsection{Control and simulation of the upstream processes}
The consistent quality of biopharmaceutical products relies heavily on effective control of bioprocesses. Within a biopharmaceutical plant, the bioreactor is a crucial component, and one of the key process parameters to be optimised is the nutrient and metabolite concentrations in the bioreactor. Therefore, achieving accurate control over the cell culture process within the bioreactors is highly desirable \citep{crwh14}. The most difficult aspect of controlling bioreactors is managing the cell growth, due to the sensitive and complicated nature of the cells that are responsible for converting raw materials into new cells and products \citep{vash17}. Despite advancements in monitoring capabilities as shown in subsection \ref{monitoring_cell}, the primary challenge now lies in effectively utilising the data generated by these monitoring systems to design effective control strategies. \cite{mest17} divided the process control into two groups. The first group includes offline control strategies which use model simulations or laboratory-scale studies to test and develop appropriate process operating conditions (e.g., temperature, pH, pressure, aeration rate, and stirrer speed), initial conditions, safety constraints, and control strategies. The second group is online process control which uses real-time process monitoring tools as an integrated part of developing control strategies and operating conditions.

It is obvious that the simulation of performance and behaviors of a control strategy can be conducted more rapidly than executing it in real-time before being applied in industrial-scale manufacturing. It is because the dynamics of bioprocesses are slow and complex. For instance, a typical mammalian cell culture fed-batch process can take from 1 to 3 weeks, while a perfusion process can take up to 3 months \citep{crwh14}. By simulating the control strategies, errors in configuration can be identified and corrected before practical implementation, which saves time and minimises the risk of control failures. Simulations can also be used to test different `what-if' scenarios and aid in developing a realistic process model, which can then be used to determine optimal operational conditions. In this way, a mechanistic model has more advantages than a data-driven model \citep{dosc17, kocl17} because it is dynamic and predictive as well as able to represent the non-linear behaviors of fermentation systems without the need for a large number of training samples \citep{mest17}. As a result, mechanistic models are suitable for development of simulation systems and verifying various control strategies. 

\begin{figure}
    \centering
    \includegraphics[width=0.6\textwidth]{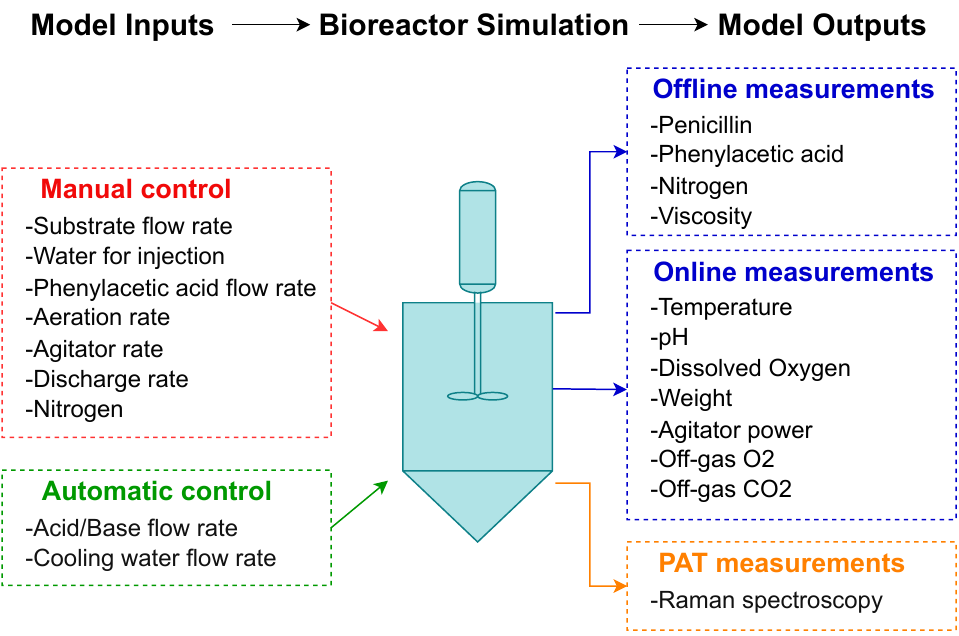}
    \caption{An overview of inputs and outputs of an industrial-scale penicillin fermentation simulation (adapted from \citep{godu19}).}
    \label{pennicillin_sim}
\end{figure}

\cite{gost15} developed a semi-batch bioreactor simulation model for an industrial-scale penicillin product, with a total size of 100,000L. This model was validated using historical data collected from a real plant. The process model employed a first principles mathematical model to represent the mechanisms of cell growth, substrate intake, and penicillin production. The simulation successfully captured the complex process dynamics and accounted for common faults associated with modern biotechnology facilities. The model also incorporated a time-varying structured kinetic model that accurately represented the biomass and penicillin, as well as the primary process variables of an industrial-scale fermentation process. The simulation accounted for various environmental factors, such as dissolved oxygen, viscosity, temperature, pH, dissolved carbon dioxide, nitrogen, and phenylacetic acid, which affect industrial-scale penicillin production. As a result, the simulation model is particularly valuable in testing different control strategies before implementation. In a later study, \cite{godu19} demonstrated the capability of this simulation as a bench-marking tool for the development and testing of new control strategies which can be applied in biopharmaceutical processes. The authors extended the proposed tool by adding a simulated Raman spectroscopy device into the framework. The overview of an industrial-scale penicillin fermentation simulation can be summarised in Fig. \ref{pennicillin_sim}. The simulation tool has the capability to operate in either fixed or operator-controlled mode, and it produces all the on-line, off-line, and Raman spectra that are available for every batch. By using this simulation, the authors designed a control strategy aimed at maximising the annual production of penicillin while also reducing the variability in batch yields. The research also illustrated the identification of CPPs and CQAs that have an impact on penicillin production through simulated scenarios. The authors showed the improvement of the control strategy for pH and temperature by minimising their variations in comparison to the current PID control loops. By using the spectra recorded by the Raman spectroscopy device, the study demonstrated the ability to develop a soft-sensor that can predict phenylacetic acid, biomass, or penicillin concentration in real-time. Additionally, the authors developed a control strategy that determines the optimal harvest time for each batch to maximise the annual penicillin yields produced during a yearly campaign. In another study, \cite{ohpa22} built a virtual plant for penicillin fermentation by slightly modifying the simulation model proposed in \cite{gost15}. Based on the built simulation, \cite{ohpa22} developed integrated approach between the double deep Q-network and model predictive control to achieve the substrate feeding strategy of a semi-batch bioreactor. Through simulation of a penicillin fermentation bioreactor, \cite{kipa21} designed a two-stage optimal control framework for a fed-batch bioreactor using reinforcement learning, which can be readily implemented in industrial scale manufacturing by addressing issues regarding a model-plant mismatch and real-time disturbances. Compared to the production of penicillin, the process of mammalian cell cultures is even more complex. \cite{crwh14} demonstrated the successful use of nonlinear, first principle, mechanistic mathematical models to simulate a non-linear model predictive control that effectively control the bioreactor environment in a Chinese hamster ovary mammalian cell fed-batch process. The simulator was tuned based on historical offline measurements of CPPs and KPIs. Then, the offline simulator was combined with real-time monitoring data of nutrient and metabolite concentrations values to develop a control strategy by dynamically adjusting the nutrient feed rate to the bioreactor to maintain the glucose concentration above a set-point.

Unlike offline process control, where the process models are built from mechanistic models, advanced online process control can take benefits from real-time process monitoring, which can be obtained by using soft-sensors. In case of any unusual process behavior, real-time fault detection enables to conduct corrective feedback or feed-forward control mechanisms to ensure final product quality \citep{wadi20, rowo18}. For example, multivariate spectroscopic models can enable real-time control of process variables such as glucose and lactate concentrations in bioreactors. However, a significant amount of process information that can serve as the basis for control has still been collected through offline measurements, leading to a time delay in control strategies \citep{nalu19}. In addition to DO, pH, and temperature, the concentrations of glucose, ammonia, and osmolality are crucial parameters for determining the glycosylation patterns of monoclonal antibodies \citep{robi16}. To enable the development of more advanced control strategies, spectroscopic soft sensors based on techniques like Raman spectroscopy can be used to quantify these additional variables in real-time \citep{dofr20, mabe16, bedo15}. For example, a feedback control mechanism based on real-time monitoring of glucose using Raman spectroscopy has resulted in a reduction of glycation variability by over 50\% \citep{bedo15}. Hence, process monitoring and advance process control can provide a deeper understanding of process behaviors, leading to consistent maintenance of product quality.

\begin{figure}[!ht]
    \centering
    \includegraphics[width=1\textwidth]{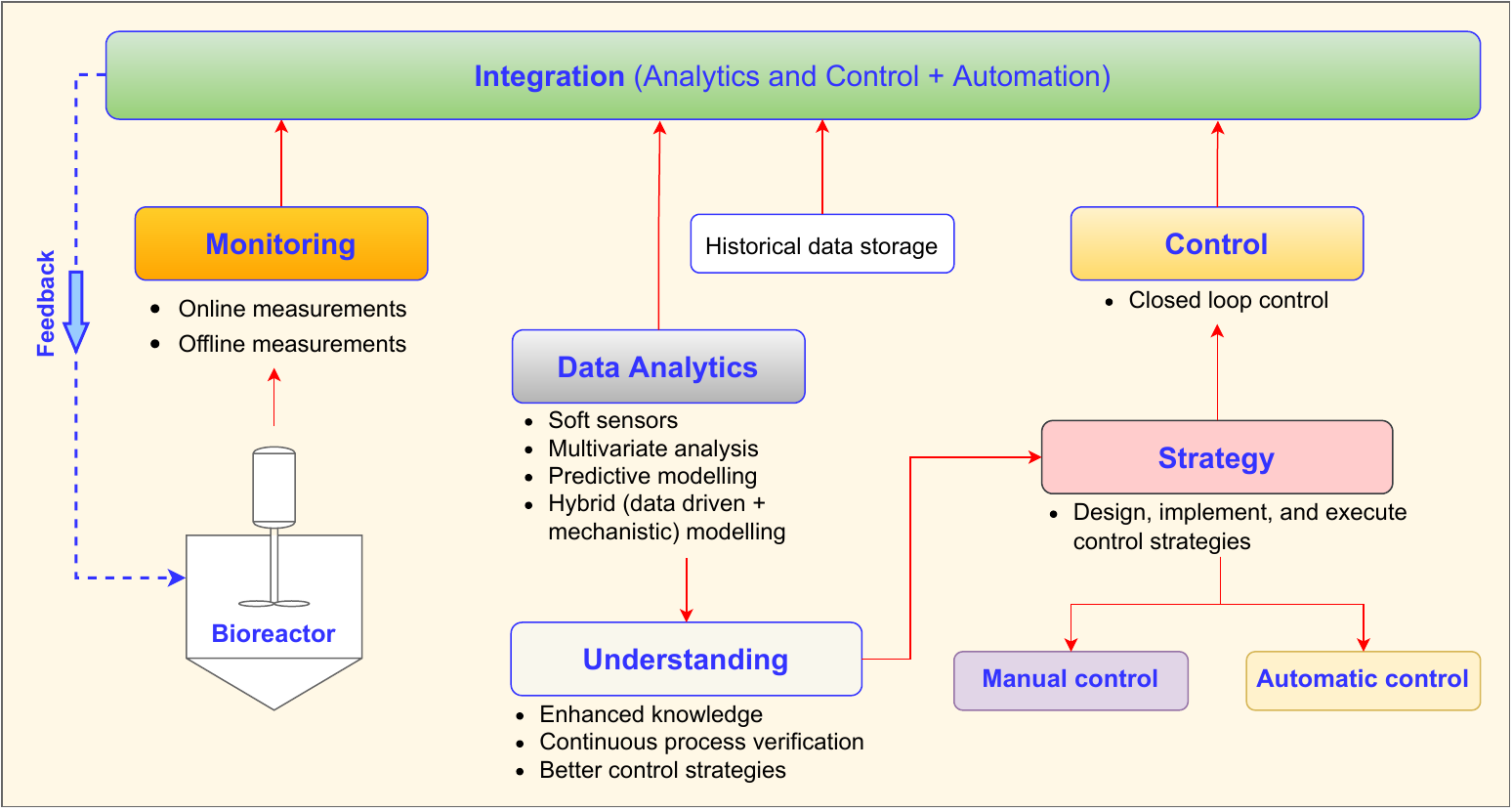}
    \caption{An integrated framework for monitoring, data analytics, understanding, strategy design and control (adapted from \cite{vash17}).}
    \label{monitor_control_framework}
\end{figure}

In manufacturing, a substantial challenge is how to use advanced monitoring tools effectively. To address this challenge, it is essential to have an integrated approach that combines monitoring hardware tools to data-analytic modeling approaches. The combination of advanced monitoring tools with multivariate data analytics is a basis to build soft sensors. In the literature, there have been many review papers regarding the critical roles of soft sensors for process monitoring and control, such as \cite{redu22, gust19, rama17, sosi17, magu14, medi14, lubr12, kagr11, kaga09}. The use of soft-sensors enables timely real-time process data, which is crucial not only for generating process understanding during the process development stages but also for enabling advanced control during manufacturing \citep{luto13}. \cite{vash17} proposed an integrated framework of monitoring, data analytics, understanding, strategy design and control as presented in Fig. \ref{monitor_control_framework}. Within this framework, it becomes possible to perform data analytics to generate process understanding, enabling the integration of soft sensors or predictive models with other monitoring information to design strategies for advanced control. \cite{vash17} demonstrated the effectiveness of the integrated framework through practical implementation. The framework helped to increase batch-to-batch consistency in final product titer, reducing the coefficient of variability from 8.49 to 1.16\%, enabling the early prediction of possible failures within the process and triggering of essential actions to prevent their occurrence, thereby avoiding lost batches.

In current industry practices for large-scale mammalian cell cultures, a standard platform fed-batch process with fixed volume bolus feeding is typically used \citep{luto13}. However, bolus feeding can result in wide glucose concentration ranges, which can affect process performance and product quality attributes, e.g., leading to the accumulation of lactate and ammonia, which can increase the cell death rate. A lower glucose concentration can decrease the glucose consumption rate, lower lactate production rate, and increase the process pH, leading to lower base addition \citep{bedo15}. Consequently, changes to metabolite feeding and control can have significant effects on product quality attributes such as peak cell count, final titer, and specific glucose and lactic acid consumption \citep{kocl17}. One strategy is to establish bolus feeding profiles that are administered routinely as part of daily sampling. Another strategy is to set a defined, continuous feed rate regardless of culture feedback. Though these strategies have been widely used, they cannot meet actual nutrient consumption demands of cells, which may result in the accumulation of by-products and depletion of certain nutrients \citep{luto13}. Adaptive feeding strategies can be implemented to address these issues. To perform an adaptive control strategy for feed rate, it is desirable to acquire the real-time data from PAT tools. Table \ref{table_control} summarises the typical applications of ML in designing control strategies, mainly focusing on adaptive feeding strategies.

\setlength\LTleft{-2cm}
{
\scriptsize{
\begin{longtable}[t]{cC{1.5cm}C{1.2cm}C{1cm}C{1.8cm}L{2.8cm}L{1.8cm}L{2.5cm}L{1.5cm}L{1.8cm}}
    \caption{Summarising typical applications of ML models in bioprocess control} \label{table_control} \\
    \toprule
       \textbf{ID} & \textbf{Scale} & \textbf{Type} & \textbf{Cell} & \textbf{Offline sampling} & \textbf{Input parameters} & \textbf{Control parameters} & \textbf{Control strategy} & \textbf{Algorithm}  & \textbf{Reference}  \\
       \midrule
      1 & 100,000L bioreactor & semi-batch & Penicillin chrysogenum & every 24h & Online measurements: Temperature, pH, DO, pCO\textsubscript{2}, off-gas; Offline measurements: substrate, phenylacetic acid, biomass, penicillin concentrations, nitrogen and viscosity & Substrate rate, acid/base flow, coolant flow, heating flow, water flow & Optimal substrate feeding strategy according to the stage-wise cost, terminal cost, and constraints & Reinforcement learning with NMPC & \cite{ohpa22}\\
      \midrule
      2 & 2L, 3L bioreactors & fed-batch & N/A & once a day & Offline measurements: VCD, Viability, Glc, Lac, Gln, Glu, NH\textsubscript{4}, Na, K, Osm, Titer & Glc and nutrient feed volume & optimal feed strategy at a specified time point to maximise VCD while minimising byproducts & Hybrid FPM, ANN, and LR &  \cite{rakh22}\\
      \midrule
      3 & N/A & fed-batch & Penicillin & 0.5h & Real-time temperature, pH, agitation; Offline measurements: biomass, substrate, penicillin, hydrogen ion, DO, CO\textsubscript{2} concentrations, Culture volume & feed flow rates of substrate, cooling water, acid, and base & optimal feed trajectory that maximises the final time productivity and yield & model-based reinforcement learning & \cite{kipa21}\\
      \midrule
      4 & 2L bioreactor & fed-batch & CHO DG44 & 1-2 times/day & Raman; Offline measurements: VCD, pH, Glc, Lac, Glu, Gln, NH\textsubscript{4} & Glc feeding volume and times & dynamic feeding strategies based on real-time glucose concentration & PLSR & \cite{dofr20}\\
      \midrule
      5 & 3.5L bioreactor & fed-batch & CHO & once per day & Amplification duration, Amplification VCD, Amplification cell age, DO\textsubscript{set}, pH\textsubscript{set}, pH, DO, pCO\textsubscript{2}, Glc, Lac, Glu, Gln, NH\textsubscript{4}, Osm, VCD & Glc feeding rate & when the Glc concentration reaches below 20 mM, feeding is triggered to achieve a Glc concentration after feeding equal to 30 mM & ANNs + mass balance equations + EKF & \cite{nabe20} \\
      \midrule
      6 & 250ml Erlenmeyer flasks & fed-batch & Plectonema & once a day & biomass, nitrate consumption, and C-PC concentration & nutrient addition on daily basis  & maximise the yield of C-PC at the end of the batch & deep reinforcement learning & \cite{mano20}\\
      \midrule
      7 & N/A & fed-batch & baker's yeast & every seconds & medium's biomass and substrate concentrations & feed substrate concentration & achieve and maintain a substrate concentration of 7.5 g/l & partially supervised reinforcement learning, NMPC & \cite{pano18}\\
      \midrule
      8 & AMBR-15ml, 7L, 50L bioreactors & fed-batch & CHO & once per day & Online measurements: agitator speed, DO, DO at maximum saturation, inlet gas flow rate; Offline measurements: VCC, Viability, Glc, Lac, titer & Glc feeding volume & Online glucose advanced process control based on online oxygen transfer rate to maintain Glc at an adjustable set-point & Mathematical equations & \cite{gole18}\\
      \midrule
      9 & 2L, 30L, 300L bioreactors & fed-batch & E. coli & 30 min & NIR, inline DO, pH, agitation, airflow, temperature; Offline measurements: amino acids, Glc, B vitamins, cell density & induction, gas flow, condensate valve & trigger induction at the optimal Glc concentration and cell density; reduce gas flow and open a condensate valve to prevent a filter clog based on the results of the clog predictor & PCA, PLSR, PID & \cite{vash17} \\
      \midrule
      10 & 1L bioreactor & fed-batch & GS-NS0 & every 6h & Offline measurements: Glc, Glu,Arginine, Aspartate, Asparagine concentrations; final mAb concentration at the end of batch run & feed flow rates of Glc, Glu, Arginine, Aspartate, Asparagine & Medium is introduced in pulses of 2h duration to maximise the final mAb concentration & Monodkinetics model & \cite{paqu17}\\
      \midrule
      11 & 6.6L bioreactor & batch & Saccharomyces Cerevisiae & - & Biomass (using NIR), DO, pH, Temperature, Off-gas, acid/base concentrations, heating/cooling rate (every minute) & antibiotics addition, gassing rate, stirring rate, nitrogen addition & using out-of-control event signaled by multivariate control chart to trigger expert systems for control & PCA, multi-block PLSR, rule-based expert systems & \cite{besc16}\\
      \midrule
      12 & 2L, 80L bioreactors & fed-batch & CHO DG44 & once per day & Online measurements: pH, Glc, Lac; Offline measurements: Total cells, Viable cells, viability, IgG, Glc, Lac & Glc feed volume and time point & adaptive feeding strategy based on real-time signals, capacitance and pH & PLSR & \cite{kocl16}\\
      \midrule
      13 & 5L, 200L, 315L bioreactors & fed-batch & HEK 293 & once per day & Raman; Offline measurements: Glc, Lac, NH\textsubscript{4}, VCD, Viability, product titer & Glc feed volume and time point & adaptive feeding strategy based on online measurement of Glc and Lac estimated from Raman spectra & PLSR & \cite{mabe16}\\  
      \midrule
      14 & 500ml, 1L, 3L shake flasks & fed-batch & CHO DG44 & 2 to 4 times/day & Raman; offline measurement: Glc, Lac, Osm, VCD, TCD, Viability, IgG titer & Glc feed setpoint & When the online Glc concentration dropped below the lower deadband (0.25 g/L < setpoint), pump 200 g/L glucose stock solution & PLSR & \cite{bedo15}\\
      \midrule
      15 & 500L bioreactor & perfusion & CHO IGF-1 & every 4h; once a day & real-time gas chromatography-mass spectroscopy for Glu and mannose; Offline measurements: titer, hexose, and product glycan (every 4h); growth, Viability, Osm, Lac (daily) & \% High mannose & add or remove mannose to media with the mannose feed rate computed from NMPC & process mathematical model, least squares regression & \cite{zubr15}\\
      \midrule
      16 & 5L bioreactor & fed-batch & once per day & CHO, HEK & Online measurements: Glc, Lac (using BioPAT Trace system); Offline measurements: VCD, viability, pH, DO, pCO\textsubscript{2}, Glc, Gln, Lac, NH\textsubscript{4}, K, Na, Osm, mAb concentration & complex feed addition, Glc stock solution additions & dynamic feed strategies including a set point control (Glu < set point, trigger a feed pump to add complex nutrient feed containing Glu), fully automated feedback control, and an adaptive feeding based on real-time Glc and Lac & Feedback control &  \cite{zhts15}\\
      \midrule
      17 & 15L bioreactor & fed-batch & CHO 320 & once per day & Raman; offline measurement: Glc, Gln, Lac, NH\textsubscript{4}, VCD, Viability & Glc concentration fixed set-point & Continuously adjust feed rate to keep the Glc concentration at a set-point of 11 mM throughout the culture using a NMPC and Raman-based Glc measurements & Mechanistic models, PLS & \cite{crwh14}\\
      \midrule
      18 & 3L bioreactor & fed-batch & CHO & once per day & Online capacitance; Offline measurements: VCC, Viability, pH, NH\textsubscript{4}, Osm, DO, Ca, K, Na, Glu, Gln, Glc, Lac, amino acid, vitamins & media feed volume & dynamic feed based on predictive values of VCD or Glc concentration feedback together with feed media optimisation & mathematical equations & \cite{luto13}\\
      \midrule
      19 & 1L, 2L, 180L, 500, 2500L bioreactors & fed-batch & CHO K1 & once per day & Offline measurements: Glc, Lac concentrations, Osm, gas, pH, IVCD, titer & Glc feed volume & high-end pH-controlled delivery of glucose leading to a doubling of the final titers & simple threshold & \cite{gahi11}\\ 
      \midrule
      20 & 2L, 8L, 24L bioreactors & batch, fed-batch & BHK-21A & 6-12h & Offline measurements: viable cells, Glc, Lac, Gln, NH\textsubscript{4}, alanine, IgG concentrations & Glc and Gln feeding volume & optimal feeding strategies of Glc and Gln & kinetics models + ANN & \cite{teal07}\\
      \bottomrule
      \multicolumn{10}{L{18cm}}{\scriptsize \textbf{ACO}: Ant colony optimisation, \textbf{AMBR}: Advanced micro-bioreactor, \textbf{ANN}: Artificial neural network, \textbf{CHO}: Chinese hamster ovary, \textbf{C-PC}: cyanobacterial-phycocyanin, \textbf{EKF}: Extended Kalman filter, \textbf{Glc}: Glucose, \textbf{Gln}: Glutamine, \textbf{Glu}: Glutamate, \textbf{GP}: Gaussian Process, \textbf{HEK}: human embryonic kidney, \textbf{K}: potassium, \textbf{Lac}: Lactate, \textbf{LR}: Linear regression, \textbf{MPC}: Model Predictive Control, \textbf{Na}: Sodium, \textbf{NH}\textsubscript{4}: Ammonium, \textbf{NIR}: Near‑infrared spectroscopy, \textbf{NMPC}: Non-linear model predictive controller, \textbf{Osm}: Osmolality, \textbf{PID}: proportional-integral-derivative controller, \textbf{PCA}: Principal component analysis, \textbf{PLSR}: Partial least square regression, \textbf{FPM}: first principle model, \textbf{TCC}: total cell concentration, \textbf{TCD}: Total cell density, \textbf{VCC}: viable cell concentration, \textbf{VCD}: Viable cell density}
\end{longtable}
}}

\cite{bedo15} presented the design of a process automation system that allows for glucose setpoint control in real-time. A glucose-free nutrient feed was administered on a daily basis, while a glucose stock solution was supplied as required based on online Raman measurements. Two feedback control conditions were implemented, one keeping glucose at a constant low concentration, and another reducing it in a step-wise manner. Results showed that glycation levels decreased from 9\% to 4\% when using the low target concentration, but not in the step-wise condition, compared to the bolus glucose feeding method. \cite{dofr20} demonstrated that the dynamic feeding control approach is an efficient way to maintain the concentration of nutrients, such as glucose and glutamate, within a narrow and low range. The authors showed that an automated feeding control system using Raman spectroscopy can manipulate the cell metabolism to use lactate as the primary carbon source, resulting in moderate toxic by-products accumulation and a favorable environment for the cells. As a result, the VCD increased in the two cultivations with dynamic feeding compared to the bolus fed experiment, and prolonged cell viability was achieved, leading to an increased harvest titer in the dynamic fed experiments.

Despite initial positive results in developing control strategies at laboratory scales, transferring them to cell culture manufacturing processes faces several challenges as summarised in \cite{dosc17}, including the complex, interconnected nature of biological systems, difficulty in measuring critical quality attributes, and the need for complex models to capture the nonlinear dynamics of these processes. To address these challenges, researchers can leverage insights into cellular metabolism to develop effective process control strategies. For instance, pH rise can signal lactate consumption, which can trigger glucose feed \citep{hiov17}. To support closed-loop control, soft sensors integrated with PAT tools such as Raman spectroscopy can provide real-time measurements of quality attributes and performance indicators. Notably, Raman spectroscopy has been successfully used for glucose feed-back control in fed-batch cultures such as in \cite{escu21, dofr20, bedo15}. According to \cite{nalu19}, the definition of adaptive dynamic control algorithms for real-time prediction and control of CPPs and CQAs is important to ensure the quality criteria and the realization of Pharma 4.0. Moreover, it is essential to have a supervisory control system in place that takes into account all process units and their associated devices. In that direction, \cite{labe13, labe14, lupa16} explored various promising methods for integrating different units and achieving continuous biomanufacturing. It is worth emphasising that as the industry continues to trend towards integrated and continuous operations, a reliable supervisory control and data acquisition system is crucial to serve as an interface for process control that is based on advanced process models, data analytics, and real-time PAT tools, in order to fully leverage their impact on process operation and automation. These integrated frameworks are still not popular in monitoring and control of development and manufacturing processes of biologics.

\section{The applications of ML in prediction, monitoring, control, and optimisation of downstream processes} \label{downstream}
Downstream processing employs multiple purification techniques to capture the desired protein while eliminating impurities associated with various process elements (such as antifoam and Protein A leachate), host cell materials (including host cell proteins and DNA), and product-related impurities (such as aggregates and fragments) \citep{raka14}. The specific unit operations used in downstream processing can vary depending on the product being purified, but typically include a combination of the following steps: chromatography, polishing, viral filtration/clearance, and ultrafiltration/diafiltration. This section will review typical applications of ML in various stages of a downstream process.

\subsection{Monitoring and prediction problems in the capture chromatography phase}
Chromatography has a vital role to play in the purification of biopharmaceutical products. Currently, a common purification method for biopharmaceuticals involves using two or more single-column chromatographic steps in succession \citep{defe20}. The initial step, known as the capture step, is used to eliminate any non-product-related impurities, such as host-cell proteins, DNA, and lipids. Following the capture step, multiple polishing steps are employed to achieve the desired level of purity for the target molecule by removing impurities that are product-related. These impurities include species produced during synthesis that have very similar chemical characteristics to the target compound, such as truncated or deamidated species and aggregates. This section will cover the typical applications of ML in addressing the monitoring and prediction problems in a capture chromatography process. The general steps involved in the capture chromatography are \citep{bi01, nalu21}:
\begin{itemize}
    \item \textit{Column preparation}: The chromatography column is packed with a resin which has a high binding capacity for the target molecule. The resin is typically pre-treated to optimise its performance and minimise non-specific binding.
    \item \textit{Sample loading}: The sample is loaded onto the column, usually using a peristaltic pump. The sample is enabled to flow through the resin bed, and the target molecule binds to the resin while other impurities pass through.
    \item \textit{Washing}: After the sample has been loaded, the column is washed with a buffer solution to eliminate any unbound impurities.
    \item \textit{Elution}: The target molecule is then eluted from the column using a specific elution buffer disrupting the interaction between the target molecule and the resin. The eluted material is collected in a separate container for further processing.
    \item \textit{Column regeneration}: After the elution, the column is typically regenerated using a cleaning solution to eliminate any remaining impurities and prepare it for the next cycle of purification.
\end{itemize}

Machine learning can be used to develop predictive models that can forecast the behavior of the capture chromatography process based on historical data, e.g., fouling columns and breakthrough profiles. This can help to identify potential issues before they occur, allowing for proactive corrective action. Machine learning can also be used to monitor the capture chromatography process in real-time, optimising operating parameters and enhancing the efficiency of the overall process.

During the capture phase, a protein mixture is passed through a column containing a stationary phase (which is typically a resin with specific ligands) designed to selectively bind the target protein, while other impurities and proteins will pass through. The relationship between the concentration of a target protein (e.g., mAbs) in the column effluent and the volume of the sample that has been passed through the column over time is graphically represented by breakthrough curves. Initially, the effluent contains a low concentration of the target protein, but as more sample is passed through the column, the concentration of the target protein in the effluent increases until it reaches a plateau. Breakthrough curves are useful for assessing the efficiency of the chromatography column, identifying the maximum loading capacity, and determining the point of breakthrough, which is when the column is no longer able to capture the target protein effectively. The shape and characteristics of the breakthrough curves provide valuable information regarding the behavior of the target protein in the chromatography process. For instance, a steep breakthrough profile can indicate that the binding capacity of the chromatography resin is being exceeded, and additional steps may be required to obtain the desired purity \citep{sush21}. On the other hand, a more flat breakthrough profile can indicate that the chromatography resin is underused and that higher yields could be obtained with a higher loading density. Therefore, the accurate prediction of the breakthrough curves is vital to optimise various parameters of the capture chromatography process, such as the loading density onto the column, flow rate, and elution conditions \citep{nalu21}. In addition, by accurately predicting the breakthrough curves, one can reduce time and cost while maximising the yield and purity of the target protein.

\cite{fega19a} presented a hybrid model combining a deterministic chromatographic model with online information generated from Raman-based PLS estimates to estimate the mAb concentration at the outlet of a column aiming to monitor the chromatographic breakthrough curves. Experimental results indicated that Raman-based predictive PLS models could capture the general shape of the breakthrough curves, but their results were too noisy to be used in practical applications such as process control. This noise was caused by the high levels of impurities in the harvested sample, resulting in overlapping spectral features of different species such as the target monoclonal antibody, media components, host cell proteins, DNA, and high molecular weight compounds. As a result, broad bands and indistinct peak profiles were observed within single spectra. On the other hand, the mechanistic model, i.e., the lumped kinetic model (LKM), which was appropriately calibrated with an external dataset, was able to capture the qualitative shape of the breakthrough curves. However, it showed deviations from the off-line reference measurements that were too large. Therefore, the authors proposed an extended Kalman filter approach, which combines the contribution of the Raman-PLS and the LKM predictions in the final estimated values of the filter. In the hybrid model, the LKM provides robust predictions, while the real-time information from Raman-PLS can be used to update the state estimates and effectively minimise the LKM offset. As a result, the proposed EKF provides superior results in comparison to individual Raman-PLS and LKM models. In a later study, \cite{nalu22} confirmed that both data-driven and mechanistic models are prone to significant biases. The data-driven models lack generalisability, while the mechanistic models are overly constrained and may not adequately represent the underlying phenomena. To overcome these issues, the authors proposed to use different degree of hybridisation between an LKM model and neural networks to formulate various hybrid models. The empirical outcomes showed that hybrid models provided accurate predictions of breakthroughs and internal column profiles of solid and liquid-phase concentration. This information is critical for process control and optimisation because to optimise resin utilisation and maintain high yield during the capture step, it is crucial to load the column to its maximum capacity while preventing any breakthrough. The experimental outcomes confirmed that hybrid models exhibit very little bias and they can adapt to various systems with little variability because of the
mechanistic part supporting robust extrapolations. \cite{nase21} proposed a novel hybrid modeling approach for predicting breakthrough curves in the capture chromatography procedure, which combines neural networks and mechanistic models to learn the features of the process in an unbiased way. The performance and potential of the hybrid model are investigated on both in-silico and experimental datasets, and the results show that the hybrid model outperformed the LKM in terms of prediction accuracy and robustness in extrapolating across process conditions.

Compared to upstream processing, online or at-line monitoring tools are not widely used in downstream processing, especially in chromatography, and only a few examples have been presented in the literature \citep{raka14, fega19}. The implementation of online monitoring techniques in downstream is challenging due to stringent requirements such as high sensitivity, robustness, fast response time, high accuracy, wide dynamic range, and low limit of detection (LOD) with minimal recalibration needs \citep{roma16}. pH, conductivity, pressure, mass flow, optical density, and single wavelength UV spectroscopy sensors are commonly used in chromatography. However, the data they provide is limited and cannot be used to accurately determine product concentration or other CQAs such as product-related impurities (e.g., aggregates, fragments, isoforms) or process-related impurities (e.g., HCPs, DNA, leached resin ligands) \citep{fega19}. Currently, the most widely used method for quality control in the downstream processing of therapeutic proteins is high performance liquid chromatography (HPLC) \citep{fega19, tika18}, which generally operates in off-line mode but in few cases also at-line. However, the HPLC has certain limitations. For instance, it requires either manual sample handling or a costly autosampler but still exhibiting non-negligible time delays \citep{rubr17}. Furthermore, sample preparation and analysis can be time-consuming, and thus real-time decisions are not possible. In contrast, the spectroscopy is an emerging technique that shows a great potential for in-line implementation. It offers a short measurement time ranging from seconds to minutes, and it is a noninvasive and nondestructive in its operation. Additionally, the spectroscopy enables a simultaneous quantification of multiple variables, making it a promising alternative to HPLC for quality control and real-time monitoring in downstream processing \citep{rubr17}, where fast measurement times are vital for the chromatography steps. To take advantage of strong points of spectroscopy for real-time monitoring of downstream processes, ML algorithms and multivariate
data analysis approaches are usually applied to extract information from spectroscopic measurements. A critical review on recent trends and a potential of spectroscopy techniques as PAT tools for biopharmaceutical downstream processing can be found in \cite{roru20}.

In another review paper discussing popular spectroscopic tools for PAT applied in downstream processing, \citep{rubr17} concluded that UV/vis spectroscopy is primarily used to measure the content of aromatic amino acids, providing information on the primary structure of proteins. Vibrational spectroscopy, such as Fourier-Transform Infrared (FTIR) and Raman spectroscopy, is frequently used to measure the concentration and secondary structure of proteins. These methods enable to measure the vibrational modes of the polypeptide backbone. The fluorescence of aromatic amino acids allows for the assessment of the tertiary structure of proteins. Finally, the quaternary structure of proteins, including the assembly of multiple subunits or native aggregation of protein monomers, can be assessed using quasi-elastic light scattering techniques such as static light scattering (SLS) and dynamic light scattering (DLS), which provide information on protein size. This information is essential to separate the target protein from other impurities.

\cite{rubr16} demonstrated that PLS modeling on UV/vis absorption spectra can be used to quantify the concentration of mAb in the effluent of a Protein A capture step during the load phase, despite the presence of protein and non-protein-based contaminants. The PLS model was successfully calibrated based on the corresponding absorption spectra of the effluent in several breakthrough experiments, and it accurately predicted the mAb concentrations in the effluent during a validation experiment. This study highlights the potential of UV/Vis spectroscopy for real-time monitoring and control of the loading phase in protein A capture chromatography. In practice, capture chromatographic procedures are typically operated at high loading densities to achieve efficient processes \citep{caju20}. However, high loading densities usually result in a wide range of product and contaminant concentrations that coelute from the column. Therefore, monitoring methods must have an ability to quantify the broad dynamic range of protein concentrations. \cite{brru18} proposed a method using variable pathlength UV/Vis spectroscopy in conjunction with PLS modelling to selectively quantify coeluting protein species with highly loaded columns. The method was demonstrated on the separation of lysozyme (considered as a product) and cytochrome c (contaminant) at high loading densities, with a broad dynamic range of protein concentrations. The concentration of the product peak ranged from 30 g/l to 80 g/l, while the concentration of the contaminant peak was only 4 g/l to 20 g/l. The separation outcomes based on the proposed approach predicted a purity of 99\%, while off-line analytics showed a 99.7\% purity. The authors concluded that their approach has potential for in-line monitoring and control of the capture chromatography.

Protein A capture is widely used in monoclonal antibody purification processes due to its high selectivity \citep{tave12, tswa14}. However, the resin's capacity decreases over time due to leaching and degradation of the Protein A ligands, as well as blockage by impurities or leftover product \citep{jili09}. The resin needs to be substituted after 80 to 200 cycles depending on
the purity required for the sample \citep{brpa15}. This leads to increased raw material costs on an industrial scale, as replacing fouled Protein A resin incurs a significant expense \citep{boka16}. As a result, cleaning-in-place (CIP) procedures have been developed to extend the resin's lifespan \citep{grer11} by clearing non-eluting proteins from the resin, but it is difficult to quantify any remaining contaminants reliably through chromatograms over repeated cycles. \cite{boka16} used the FTIR spectroscopy to monitor the resin fouling \textit{in situ} during an affinity chromatography by embedding an attenuated total reflection (ATR) sensor inside a micro-scale column. The PLS model was used to estimate the total protein content of resin over multiple process steps in contact with the ATR crystal. The results revealed that Protein A ligand leached during CIP procedures and that host cell proteins bound more strongly to the resin than mAbs. The study also found that typical CIP conditions did not remove all fouling contaminants, highlighting the need for more effective CIP protocols to optimise the resin lifespan. The monitoring of in-column binding behaviour of mAb during a Protein A capture could help optimise the CIP protocol, track mass balance, and monitor protein contaminant build-up for quality assurance purposes. The study suggested that even minor alterations in purification and cleaning protocols could result in significant cost savings in the production of therapeutic mAbs. \cite{grru18} also showed that FTIR in conjunction with the PLS modeling is capable of distinguishing and quantifying proteins in real-time based on their secondary structure, as well as monitoring process-related chemical contaminants like Triton X-100. FTIR provides additional information compared to UV/Vis spectra and could be useful for monitoring process attributes that were previously not visible.

According to \cite{fega19}, UV/Vis spectroscopy primarily assesses the amino acid composition of a protein, while the application of FTIR spectroscopy is restricted due to its sensitivity to water interactions. In contrast, Raman spectroscopy can assess a protein's secondary and tertiary structure without being significantly affected by water interactions. Therefore, \cite{fega19} conducted a study to evaluate the ability of Raman spectroscopy in online monitoring of the mAb concentration in the eluate stream after the protein A capture step. The authors developed an in-line flow cell to increase signal intensity, optimised measurement settings, performed off-line experiments to enhance model robustness, and developed a chemometric modeling pipeline including wavelength selection, PLS model calibration, and cross-validation. The empirical results demonstrated good estimation performance with an average root mean square error of 0.12 mg/mL and an average LOD of 0.24 mg/mL. 

Other spectroscopic methods such as near infrared (NIR) spectroscopy in combination with PLS modelling have also been used to real-time monitor and quantify mAb concentration and changes in the column binding capacity \citep{thhe19}. This can provide early warning of resin degradation or column quality problems and enable to activate the dynamic loading mechanisms of protein A capture chromatography. In this way, resin associated costs can be reduced while still ensuring consistent elution time and quality. Furthermore, efforts have also been made to overcome the limitations of using a single spectral technique by incorporating multiple inputs. In a study conducted by \cite{wasc19}, conventional detectors such as UV, pH, and conductivity were combined with additional techniques including fluorescence spectroscopy, mid infrared, light scattering, and refractive index measurement to monitor protein A capture chromatography. These inputs were then used in PLS regression to create predictive models for mAb concentration, monomer purity, aggregate content, and host-related impurities such as HCPs and DNA. While accurate predictions were made for titer and monomer purity, the results were less reliable for HCPs, DNA, and aggregate percentage, particularly when the sample matrix was altered.

The process variations, e.g., aging of column or process errors, in the capture chromatography can cause deviations in product and impurity levels, which can in turn lead to decreased purification performance. Therefore, \cite{wabr17} proposed a novel method that combines mechanistic chromatography models and artificial neural networks to identify the root cause of these deviations. The method is demonstrated in a case study, where it successfully predicts deviations in column capacity and elution gradient length with maximal errors of 1.5\% and 4.9\%, respectively. The study focuses on the ionic capacity of the column and the salt gradient length as these parameters can have a significant impact on the chromatogram, but their actual values are hard to measure quantitatively. The ionic capacity of a column can be affected by daily use, aging, or column exchange, and any deviation in capacity can lead to variations in the elution behavior of both the product and impurities. The proposed method uses \textit{in silico} experimentation to generate information about the interrelationships between the chromatograms and these two parameters, which is then used by the ANN to recognise chromatograms and respond with the corresponding parameter values.

\subsection{Monitoring and prediction problems in the polishing phase}
Typically, within mAb platform processes, the protein A capture and low pH virus inactivation are succeeded by two steps of polishing chromatography \citep{shhu07}. While the capture chromatography step in the downstream purification is usually utilised to eliminate process-related impurities (e.g., residual HCPs, DNA, and leachables), polishing steps are usually deployed to remove product-related impurities such as variants, aggregation, fragmentation, chemical impurities (residual solvents, reagents), and glycosylation variants (See Table \ref{cqa_table} for more details). It is important to monitor and control these product-related impurities during biopharmaceutical manufacturing to ensure the safety and efficacy of the final product. Ion chromatography, which is a a technique for separating, analysing charged species and measuring the specific ion concentrations in a sample, is usually used within these polishing steps. There have been limited studies in the literature regarding the applications of ML and data analysis to address the monitoring and prediction problems in the polishing steps.

One of the issues associated with ion-exchange chromatography is selecting ion-exchange materials with high binding capacity to capture charged biomolecules in high-salt environments. \cite{yabr07} utilised a quantitative structure-property relationship (QSPR) modeling method to examine the relationship between the physicochemical properties and structural components of multi-modal ion-exchange ligands and their ability to bind proteins under high-salt conditions. The analysed ion-exchange ligands contained diverse substructures that may contribute to secondary interactions, leading to protein binding. In their work, the authors created a series of molecular descriptors based on the structures of the cation-exchange ligands. Then, a support vector machine model was built from 2D and 3D molecular descriptors to derive QSPR models of elution conductivity of the interested proteins. The results indicated that the trained models can predict the performance of multi-modal ion-exchange ligands with their ability to bind proteins under high-salt conditions. Furthermore, the analysis of the selected descriptors offered valuable insights into the critical physicochemical properties and structural characteristics necessary for protein binding in high-salt conditions. The outcomes showed that while aromatic rings promote protein binding, intermediate hydrophobicity (e.g., aliphatic side chain) and hydrogen bond donors (e.g. NH and OH) tend to suppress it. The results also indicated that negative partial charge regions of the ligands promote protein binding under high-salt conditions. The results suggest that QSPR models can be employed to design mixed-mode chromatographic systems for protein separations.

Product-related impurities usually exhibit similar physicochemical properties as the product and thus are difficult to separate. The proteins which have similar physicochemical properties regularly elute at the same time, resulting in co-elution. Co-eluting proteins can be challenging to separate and identify, particularly if they have similar molecular weights or other properties \citep{brbr14}. \cite{brsa15} demonstrated that the combination of mid-UV absorption spectra with PLS modelling can be successfully applied in real-time monitoring of co-eluting proteins in chromatography via two case studies. In the first case study, the PLS model was used for inline quantification of mAb, its aggregates (High Molecular Weights - HMWs) and fragments (Low Molecular Weights - LMWs) within a cation exchange (CEX) step. The CEX process usually aims to decrease the presence of both HMW and LMW species, which share comparable physicochemical characteristics with the mAb product. The obtained results indicated a good inline quantification of the concentrations of mAb monomer, LMWs, and HMWs in comparison with the offline reference analytics. The second case study addresses the purification of transferrin (trf) and IgG from Cohn supernatant I utilising flow-through mode anion-exchange (AEX) chromatography. The PLS models calibrated on average mid-UV absorption spectra from inline data was used to predict the selective protein concentrations of IgG, trf, IgM, and IgA in the validation runs. The empirical outcomes showed that accurate elution profiles for the higher concentrated target proteins IgG and trf were successfully predicted. Furthermore, the prediction of the lower concentrated IgM exhibited a strong agreement with the offline reference analytics. As a result, the PLS models enabled to successfully quantify co-eluting serum proteins in an AEX-based purification of Cohn supernatant I. In another study, \cite{brru18} demonstrated the use of variable pathlength UV/Vis spectroscopy in combination with PLS modelling to inline protein concentration quantification of mAb monomer and its HMW variants. As shown in the analysis, the differences in the spectra data of mAb monomer and HMW are relatively small because it is hypothesised that both the mAb monomer and HMW contain an equivalent mass proportion of aromatic amino acids and disulfide bridges. Therefore, any differences in their spectra are likely attributed to changes in tertiary structure or the effects of light scattering \citep{brru18}. However, the obtained results indicated that a well calibrated PLS model can identify slight differences in the absorption spectra. As a result, the concentrations of mAb monomer and HMW predicted by the PLS models closely matched the offline reference analytics. The use of a Savitzky-Golay filter for smoothing the spectra data can enhance the prediction performance of PLS models. \cite{goum20} used Raman spectroscopy and PLS models to rapidly determine the total protein concentration and monomer purity of Fc-fusion protein during CEX chromatography purification to identify the next prioritised analytical methods, thus operator and instrument time can be saved. The outcomes showed that if the trained PLS model is used to make prediction on spectral data with the operation conditions different from the one on which it was trained, the prediction accuracy will significantly decrease. Therefore, the authors suggested that separate models should be developed for specific conditions.

As shown in \cite{faiy99}, the performance of the chromatography process is greatly influenced by the flow rate, and that higher flow rates, along with lower column length and dynamic capacity, represent the optimal conditions for maximising productivity. \cite{niti21} proposed to use reinforcement learning to estimate the optimal process flow rate within the CEX chromatography for maximum separation of charge variants aiming to maximise yield and productivity. The proposed method was verified on a simulator of the CEX chromatography using mechanistic models for mass transfer dynamics in ion-exchange columns and extended Langmuir equations for salt concentration as presented in \cite{kule15}. The simulation was employed to create elution profiles further used for designing the learning policy for optimised flow rates.

\subsection{Control of a chromatography process}
In protein A capture chromatography, control of the load phase is crucial for efficient and selective purification of monoclonal antibodies \citep{thhe19}. It plays critical roles in optimising binding capacity, minimising non-specific binding, preventing resin overloading, ensuring buffer compatibility, and maintaining process consistency. By effectively controlling the load phase, it is possible to achieve optimal binding of the target protein to the column and efficient purification of high-quality monoclonal antibodies. However, the loading phase in the protein A capture process is usually not controlled in real-time, instead the load volume frequently relies on the offline quantification of the mAb and a conservative column capacity computed from resin-life time studies \citep{rubr16}. Therefore, \cite{rubr16} proposed a new method for real-time control of the load phase and terminated loading based on the online estimated values of mAb in the effluent of a Protein A capture step generated by PLS models using UV/Vis absorption spectra. To verify the performance of the proposed method, the trained PLS model was deployed for real-time controlling the load phase of a Protein A capture step through two different runs. The load phase was automatically terminated if a mAb concentration in the effluent reached 50\% of product breakthrough (the first run) or 5\% of product breakthrough (the second run). The empirical outcomes showed that in both runs, the corresponding load phases were successfully terminated close to the targeted breakpoints. In another study, \cite{thhe19} developed near infrared spectroscopy flow through cells, located before the inlet of the load column and optionally at the outlet of the load column, to capture spectra data in the harvested broth and flow-through every three seconds. These spectra are subsequently fed to online PLS models, which are calibrated with the reference spectra, to quantify the concentration of mAb in both the harvested broth and flow-through. The proposed method was verified using two experimental setups. In setup A, only the mAb concentration in the load stream was monitored. The control algorithm received concentration data every three seconds and made decisions whether or not to stop the flow of the loading pump based on the percentage breakthrough estimated from the mAb concentrations before column inlet and after column outlet. In setup B, the mAb concentrations were monitored in both the load and flow-through. The control algorithm was designed to alter the valve configuration and redirect the flow-through material to the second pass column instead of the waste tank once a certain level of mAb breakthrough was detected. The real-time loading control method achieved the optimal resin utilisation while still maintaining periodic elutions.

After the target protein is captured by the chromatography resin, several washing and elution steps are conducted to eliminate impurities and obtain the purified protein. Each chromatography step produces multiple fractions or eluates that include different levels of the target protein and other impurities. The pooling decisions are then performed to select and combine fractions from different chromatography runs or columns to achieve the desired level of purity and yield. These decisions are based on an analysis of the purity and quantity of each fraction, and they help to reduce batch-to-batch variability and minimise the overall process time and cost. \cite{brru18} demonstrated an inline control strategy of pooling decisions based on either the eluting protein concentrations or the purity of pools predicted by PLS models using variable pathlength UV/Vis spectra. In the first verification case study where lysozyme (the target product) was purified from cytochrome c, the pooling decision was triggered when the concentration of lysozyme exceeded 2 g/L and the cytochrome c concentration fell bellow 1.8 g/l based on the estimated values provided by the PLS models. In the second case study regarding the separation of HMWs from the mAb monomer, the pooling was triggered if the mAb monomer concentration is larger than 2 g/l and this pooling step was terminated if the purity fell bellow 95\%. In both case studies, the pool purity was then measured using offline reference analytics to compare the practical results with the one predicted by the PLS model. For the first case study, a predicted purity of 99.0\% was obtained by the PLS model, while offline analytics resulted in a measured purity of 99.7\%. In the second case study, the PLS model predicted a purity of 94.4\%, while off-line analytics measured a slightly lower purity of 94.2\%. The obtained results confirmed that the proposed method may be deployed for in-line control of pooling decisions within a chromatography process.

\subsection{Applications of ML in filtration}
Depth filtration (DF) is a frequently employed method to clarify cell culture broth when producing mAb products due to its low initial cost, straightforward equipment, ease of operation, and validation \citep{obbr12}. It can operate as either the initial step (direct depth filtration) or a subsequent step in the clarification process. A sequence of one or more stages of depth filters can be deployed, with each successive filter eliminating progressively smaller particles, generating a clarification sequence (starting with coarse and ending with fine depth filters) \citep{vazy07}. In recent years, as bioreactor processes have progressed towards longer fermentation times and higher cell concentrations to achieve greater product titers and yields, there has been a notable increase in cell debris and organic constituents, leading to lower cell viability \citep{agra16}. As a result, the loading capacity of depth filters is reduced. To address this, manufacturing scale processes may require large, single-use assemblies with a filter area of several hundred square meters. It is crucial for the depth filtration step to be stable, because any inconsistency in its performance can lead to decreased efficiency in downstream purification or loss of product \citep{nopa09}.

A significant challenge in developing an effective depth filtration-based clarification strategy for large volume bioreactors is predicting filter loading capacity at scale \citep{nopa09}. The current approach for scaling up depth filtration assumes that the fouling mechanism of the filter remains the same as operating conditions are maintained, enabling linear scaling up of filter area from the lab to pilot to commercial scale \citep{agra16}. However, a filter sizing depends on several process factors and raw material attributes, such as operating flux variation, lot-to-lot variation in the filter, feed variability, and operation scale \citep{nopa09}. To address these issues, \cite{agra16} described an application of artificial neural network modelling to predict depth filter loading capacity for mAb clarification during commercial manufacturing. The proposed model used feed turbidity, cell count, feed cell viability, flux, and time as input parameters to predict the pressure increase in a depth filtration unit operation. The model was trained with experimental data and showed excellent agreement between predicted and experimental results with a regression coefficient of 0.98. The research also deployed Monte-Carlo simulations to estimate the benefits of employing variable depth filter sizing compared to using a fixed filter area. The empirical outcomes showed a 10\% cost saving in using variable depth filter sizing for different clarification lots instead of using the fixed filter sizing. This study demonstrated the potential of ANN modelling to design efficient and cost-effective manufacturing scale filtration processes.

In the production of monoclonal antibody products, the viral filtration step is crucial to effectively remove both enveloped and non-enveloped viruses based on their size. Membrane-based systems have become increasingly popular as a highly efficient technology for filtration and purification in a wide range of applications, including the viral filtration \citep{vire17}. \cite{bimi22} highlighted the significance of a membrane pore structure and reversible mAb aggregate formation on filtrate flux during a virus removal filtration. One of the major challenges in the use of a virus filter membrane technology is the membrane fouling. This fouling can compromise the virus clearance and reduce membrane productivity, defined as the product recovered per membrane surface area \citep{isqi22}. Fouling often occurs due to the product and process-related variants such as host cell proteins, proteases, and endotoxins instead of any rejected virus particles because the concentration of virus particles in the process is usually several orders of magnitude lower than that of the product \citep{bagh19}. Moreover, fouling might also affect the selectivity of the used membrane and the contents of the fractions produced during filtration, thereby influencing the efficiency and cost-effectiveness of the filtration process \citep{vire17}. Therefore, understanding the mechanisms of fouling is crucial to controlling and minimising it \citep{cuya11}.

Detecting and identifying early-stage membrane fouling in real-time is crucial to understanding fouling phenomena. Therefore, \cite{vire17} explored the use of normal Raman spectroscopy as an online tool for monitoring membrane fouling. Such techniques capable of foulant characterization and online monitoring are essential for controlling fouling in membrane processes. Results from the study indicated that normal Raman spectroscopy is a promising tool for early-stage membrane fouling monitoring. The method is both qualitative and quantitative, allowing for using the peak heights of the foulant to track the progress of fouling over time. Additionally, the study demonstrated that multivariate methods, such as principal component analysis, can reveal the dynamic behavior of fouling over time. The combination of normal Raman spectroscopy and PCA can provide semi-quantitative results describing the accumulation of organic foulant on the membrane structure.

\subsection{Optimisation of purification sequences}
The biopharmaceutical industry faces increasing pressures to design flexible and cost-effective multi-product facilities that can accommodate diverse drug candidate characteristics and process variations \citep{ke09}. This is due to high clinical failure rates, cost containment pressures, and greater variability of biopharmaceutical products compared to chemical drugs \citep{situ12}. To prevent bottlenecks and delays, as well as to meet final product specifications and cost targets, multi-product biopharmaceutical facilities require flexible process configurations that can accommodate products with varying characteristics and impurity loads. To assist in designing such facilities, \cite{situ12} proposed a meta-heuristic optimization approach using genetic algorithms to design multi-product biopharmaceutical facilities. The approach handles multiple decisions, trade-offs, and constraints simultaneously, including the selection of purification sequences for each product, optimisation of chromatography column sizing strategies, and quantifying the optimal ratio of upstream to downstream trains. The application of the proposed method to an industrially relevant case study on mAbs production allowed identification of the most cost-effective purification sequences and column sizing strategies for each product, while meeting purity targets. The results could help biopharmaceutical process development teams identify promising strategies and allocate resources to optimise critical areas (e.g., optimise the HCP removal abilities of resin). The proposed approach can assess large sets of purification strategies for their financial benefits and critical performance values, which can guide further experimental verification and analysis.

In addition to meta-heuristic algorithms, there have been few studies which attempt to integrate data-driven modelling into optimisation problems for biopharmaceutical purification processes. \cite{name04} presented a hybrid model framework for optimisation of chromatographic processes, which enables the exploration of various design scenarios and coupling of optimal column designs with optimal operating conditions. The strategy involves using experimental data to estimate parameters for a physical model, which is then used to generate data for neural network-based empirical models. These empirical models are then deployed with an optimisation algorithm to maximise objective functions (e.g., production rate, yield, maximum solute
concentration). The use of hybrid empirical models reduces computational time and allows for multivariable optimisation and rapid exploration of different scenarios for optimal designs of complex chromatographic systems. \cite{piva17} introduced a novel method that integrates detailed mechanistic models with artificial neural networks to optimize a process involving three chromatographic columns which takes into account all possible process options with a specific set of unit operations. The purification process was optimised using artificial neural networks to identify initial optimal conditions, which were then used for a local search with mechanistic models. This enabled the determination of the optimal sequence of chromatographic units and their corresponding operating conditions. The inclusion of neural networks significantly accelerated the optimisation process. The high computational speed of the proposed method makes it suitable for extending to more unit operations, which could greatly aid in accelerating downstream process development. \cite{lipa19} presented a multiscale optimisation approach for purification processes of mAb fragments, aiming to optimise chromatography decisions such as column size, the number of cycles per batch, and operational flow velocities. At the operational level, data-driven models were developed to estimate chromatography throughput using simulated datasets based on microscale experimental data with loaded mass, flow velocity, and column bed height as the inputs. The piece-wise linear regression models were deployed for this modelling step. At the process design level, in order to determine the optimal chromatography column-sizing strategies, the developed data-driven models were used to formulate two mixed-integer nonlinear programming models for the purification processes aiming to identify the optimal chromatography column-sizing strategies to minimise the total cost of goods per gram. The approach demonstrated promising results in achieving optimal throughputs and affinity resin costs in two manufacturing-scale case studies.

Downstream processing of hydrophobic recombinant proteins (such as immunoglobulin Gs - IgGs) can lead to aggregation \citep{sagr10} when exposed to stress factors such as pH, temperature, and agitation \citep{prfr12}. Therefore, a real-time evaluation of the aggregation can help optimise the downstream purification of biopharmaceuticals, which ultimately enhances the product quality during manufacturing campaigns \citep{zhsp19}. The study conducted by \cite{ohle15} showed that multi-wavelength fluorescence spectroscopy in combination with the PLSR model can be used for the real-time quantitative monitoring of IgG aggregation within various process conditions (concentration, pH, and temperature) happening during the mAb downstream purification. The PLSR model was used to classify the final product samples into the classes with different aggregation degrees based on fluorescence spectra. The trained model showed high selectivity and sensitivity in the discrimination ability, which has potential to be used for the quality control and optimisation of downstream processing. 


\section{The applications of ML to problems in the product formulation and stability}\label{production_stab}
Formulation refers to the process of combining active chemical substances with other chemical additives, such as excipients, to create a safe, effective, and convenient product for patient administration \citep{ha12}. This process aims to ensure that the final product is suitable for its intended use and can be administered to patients safely and efficiently. Typically, the formulation step of biologics takes place after ultrafiltration/diafiltration step, during which the processing buffer is replaced with the formulation buffer, and the concentration is adjusted to the desired level. In addition to safety and effectiveness, it is also important to ensure stability of pharmaceutical products in which their quality, potency, and purity are maintained over time under specified storage conditions. According to \cite{puda22}, developing formulations and stability of mAbs poses significant challenges due to their structural complexity, high sensitivity, and diversity. Additionally, in biopharmaceuticals, which are usually protein-based products, stability can be particularly challenging because of the potential for degradation of the protein structure and loss of efficacy. To address these challenges, researchers are investigating and utilising different ML and deep learning models to predict how formulation components, physicochemical factors, and stress conditions can impact drug stability. This section will review applications of ML models in addressing typical problems for the formulation and stability of biopharmaceutical products.

\subsection{Design of biopharmaceutical formulations}
The formulation design process for biopharmaceuticals, which are usually carried out by analytical screening and prior insights, can greatly benefit from the use of computational tools based on machine learning. \cite{nadi21a} proposed a Bayesian optimisation algorithm to identify a formulation which optimises the thermal stability of three tandem single-chain Fv variants of the marketed antibody Humira with only 25 experiments, compared to the much larger number required by traditional design of experiments methods, full screening or grid search approaches. The proposed algorithm also allows for transferring prior knowledge to speed up the development of new biologics or the use of new buffer agents. This proposed method can also efficiently optimise multiple biophysical properties (e.g., both thermal and interface stabilities) and minimise the amount of surfactant in the formulation, ultimately accelerating the formulation design and drug development.

\subsection{Rapid assessment of the relationships among formulated protein structure, thermal stability, and aggregation}
During the development of a therapeutic protein product, there are various stages that require quick evaluation of protein structure, thermal stability, and aggregation. For instance, when developing a therapeutic monoclonal antibody (mAb), multiple antibody variants are often compared for their thermal stability (e.g., under heating) and aggregation propensity to identify the most suitable candidate with the most desirable physical properties for further development \citep{naji10}. Later, during pre-formulation and formulation development, the chosen product candidate can undergo similar testing to determine the impact of solution conditions like pH and various excipients on protein aggregation and unfolding \citep{waku13}. \cite{zhqi15} proposed the use of combined Raman spectroscopy and dynamic light scattering system to provide both protein secondary and tertiary structure information and hydrodynamic size data for therapeutic proteins including human serum albumin (HSA) and intravenous immunoglobulin (IVIG) at a high concentration during real-time heating studies and isothermal incubations at different temperatures. This data was then analysed using PLS models to quantify the impacts of temperature and time of incubation on the secondary and tertiary structure and aggregation of the protein. The experimental results indicated that during heating, the stability and aggregation of HSA were not affected by the protein concentration or heating interval. The thermal stability of HSA was tested at different pH levels (3, 5, and 8), and the greatest stability was observed at pH 8. However, distinct unfolding and aggregation behaviors were observed at each pH level. A forced oxidation study revealed that the hydrogen peroxide treatment reduced the thermal stability of HSA.

\subsection{Classification of subvisible particles in protein formulations}
The presence of particulate matter in therapeutic protein products is a growing concern for both industrial quality control and patient safety reasons \citep{cara09}. Subvisible particles, defined as objects with a size of 25mm or less, are found in all commercial therapeutic protein formulations and can consist of aggregated proteins or nonbiological materials like a silicone oil \citep{sapa17}. However, determining which particle characteristics pose a risk to patients is challenging due to the high heterogeneity of size, shape, and composition of protein aggregates and other particles \citep{caco17}. Flow-imaging microscopy (FIM) is a commonly used method for the subvisible particle characterisation in biotherapeutics \citep{zowe13}. Despite the fact that pharmaceutical companies accumulate vast FIM image repositories of protein therapeutic products, the existing methods for analysing these images depend on low dimensional lists of morphological features while disregarding a significant portion of the other information that can be derived from the image databases (e.g., structural information in FIM images) \citep{cada18}. Deep convolutional neural networks (Deep CNNs) have proven their capability to extract predictive information from raw macroscopic image data in various tasks, without the need for the selection or specification of morphological features \citep{lebe15}. However, the main challenge in training deep CNN models for classifying FIM images of particles in therapeutic protein formulations is the lack of appropriately labeled datasets which provide finer levels of detail, such as identifying particles associated with increased risk of adverse immune responses \citep{cada18}. As a result, \cite{cada18} proposed a new strategy to overcome this drawback. The proposed method combined a simple data pooling strategy with deep CNN model to develop a classifier that can predict a class of processing conditions or stress states like a mechanical shaking, and freeze-thawing that produced different protein particles and aggregation, using only 20 images. The proposed method can effectively differentiate between protein formulations in various scenarios relevant to protein therapeutics quality control and process monitoring.

A recent study conducted by \cite{loma22} demonstrated the effectiveness of combining transfer learning with FIM images as a robust and accurate approach for analysing data and identifying subvisible particles. The study proposed utilising a CNN model pre-trained on ImageNet for particle classification, enabling the identification of various subvisible particles commonly found in pharmaceutical protein formulations. These particles include protein particles, cap fiber particles, glass particles, silicone oil microdroplets, silicone tubing particles, rubber closure particles, and polystyrene beads. The findings revealed that the implementation of the pre-trained CNN model led to a significant improvement in accuracy of about 8\% compared to traditional machine learning methods that relied solely on morphological parameters. Additionally, the pre-trained CNN model required less training time and exhibited better adaptation to the specific dataset, in contrast to training a CNN from scratch. Furthermore, the study highlighted the potential for leveraging features learned from the ImageNet dataset to effectively classify subvisible particles as a method to deal with the lack of labeled datasets.

\subsection{Automatic identification of stress sources for protein aggregation}
Prefilled syringes containing biopharmaceuticals are subject to various sources of stress prior to administration \citep{toma17}. These perturbations have the potential to cause degradation of the product, including chemical alterations and the formation of aggregates, which could potentially affect the drug's effectiveness and pose risks to the patient's safety \citep{krts15}. Therefore, identifying the primary sources of stress that trigger protein aggregation in biopharmaceutical products stored in prefilled syringes is crucial for developing effective quality control strategies during development. \cite{gash20} proposed ML methods to build FIM images-based classifiers that are able to identify five types of stress sources including friability tests, freeze-thawing, agitation and heating at 60 and 90\textdegree C to which an mAb formulation has been exposed to. CNN models trained on FIM images of subvisible particles were used to generate stress signatures. These stress signatures were then used to train classifiers such as decision trees, KNN, SVM, and ensembles for distinguishing stress source types. The authors also created subvisible particle classifiers using CNNs to automatically identify protein aggregates, air bubbles, and silicone oil droplets in biopharmaceuticals in prefilled syringes. These classifiers can assist in mitigating aggregation in biopharmaceuticals by identifying sources of stress that are most likely to result in protein aggregation.

\subsection{Assessing and quantitation of aggregation and particle formation for antibody therapeutics}

The physiologic environment in the body after administration of therapeutic proteins can affect their stability and result in changes in the safety such as protein aggregation and fragmentation \citep{scma21}. To tackle this issue, it is desirable to conduct studies for assessing the stability of therapeutic proteins using simulated physiologic conditions as covered in a review of \cite{scma21}. In addition, testing stability of biologics is crucial in the quality control process to guarantee that the medication administered to patients is safe, effective, and consistent with previous clinical and toxicological findings \citep{zhsp19}. The examination of protein aggregation is especially important because it can result in a loss of pharmaceutical properties, as well as induce unwanted immunogenicity when interacting with the immune system \citep{siba16, pako20}. The interaction between physicochemical factors, such as protein and ion concentrations, pH, temperature, and particulate contamination, is critical in causing aggregation within therapeutic formulations \citep{zhsp19}. The inclusion of stabilizers, like surfactants, in the product formulation phase can decrease the extent of aggregation, but it is important to avoid any negative impact on the antibody structure \citep{wawu17}. Furthermore, the activity of proteases from the host cells can result in the antibody fragmentation and expose hydrophobic regions, contributing to aggregation. This can be mitigated by using protease inhibitor cocktails \citep{pako20}.

Real-time quantification of aggregation and protein particle formation in the mAb therapeutics is critical in the quality control and testing. \cite{zhsp19} demonstrated the use of Raman spectroscopy in combination with ML models as a solution for this problem. The authors developed a support vector regression model to rapidly and accurately predict a wide range of protein aggregation levels in the mAb products based on Raman spectra. It is worth mentioning that each measurement took less than three minutes, including the acquisition of spectra and data analysis, which is appropriate to be used in the biopharmaceutical manufacturing. Moreover, the study also identified the specific regions of spectral changes caused by the formation of protein particles, and linked them to the mechanisms of aggregation using PCA and 2D correlation analysis approaches. The outcomes of the study also emphasised the capability of Raman spectroscopy as an analytical tool for the in-line monitoring of a protein particle formation.

\subsection{Monitoring quality attributes for the release and stability of formulated mAbs}
For a formulated protein therapeutic, a typical quality control system includes multiple assays, each monitors product quality attributes (PQAs) for the release and stability belonging into various categories such as appearance, formulation components, purity, impurity, and bioactivity \citep{wewo21}. However, this approach is labor-intensive, time-consuming, and cost-ineffective. To improve efficiency and cost-effectiveness in monitoring PQAs of the formulated products while maintaining the desired quality, advanced analytical solutions are necessary. Therefore, \cite{wewo21} proposed a method combining Raman spectroscopy, design of experiments, and multivariate data analysis (using PLS and PCA) for monitoring of multiple PQAs through a single spectroscopic scan in formulated protein therapeutics. Variable importance in the projection analysis was employed in the proposed method for the identification of the association of the spectral and chemical basis for PQA measurements. Preliminary feasibility studies have shown that MARS has the capability to accurately quantify protein concentration, osmolality, pH, PS20 concentration, methionine concentration, and N-acetyl tryptophan concentration, potentially replacing existing methods for these PQAs. Additionally, the proposed method was also shown as a promising tool for characterising PQAs related to protein modifications, such as aggregation, fragmentation, and oxidation. The authors concluded that the proposed method can be used as a potential complementary technique to the liquid chromatography mass spectrometry (LC-MS) peptide map-based multi-attribute method in the analysis of the formulated protein therapeutics. This is because the proposed technique can measure quality attributes related to the formulation and size, which are not within the scope of the LC-MS peptide map-based multi-attribute method.

\subsection{Product identification of biotherapeutics}
Product identification is an essential analysis for biotherapeutics and is required by regulatory bodies to ensure the identity of final drug products. In addition to final product, in-process identity testing is also conducted to minimise business risks during fill operations and combat counterfeiting \citep{ropo10}. However, identifying biotherapeutics, especially monoclonal antibodies, can be challenging due to their chemical structure's similarity \citep{pasi16}. Traditional methods deployed for identification can be time-consuming and labor-intensive, requiring the development of quick, reliable, and cost-effective identification methods. Therefore, \cite{pasi16} demonstrated the use of spontaneous and surface-enhanced Raman spectroscopy in combination with partial least squares discriminant analysis for the highly specific and accurate identification of closely related human and murine antibody drugs in solution. It is worth noting that the empirical sample set comprised antibodies of the same isotypes with identical amino acid sequences in their constant region. Despite the relatively minor differences in the variable regions, the proposed method demonstrated the high accuracy of Raman spectroscopy and PLS discriminant analysis in identifying mAb drugs. These results highlighted the ability of Raman spectroscopy to detect subtle structural variations and the efficacy of modelling techniques in utilising spectral markers to identify drugs correctly. The positive outcomes can establish a path towards the real-time validation and utilisation of these advanced spectroscopic techniques in quality control testing facilities and manufacturing sites.

\subsection{Prediction of the biophysical stability using a primary antibody sequence}
Ensuring biophysical stability is a crucial purpose in the development of antibody-based therapeutic solutions \citep{puda22}. \cite{kiwo11} proposed to use support vector regression models to estimate the thermal and pH stability of the formulated molecules based on the sequence information of Pfizer-generated antibodies alone which represent various species (mouse, rat, and human), isotypes, and germlines. These predictions were then used to select the molecules which were likely to behave well under large-scale GMP facilities and at clinical sites. The empirical outcomes showed an accurate prediction of the pH stability, which is represented by pH\textsubscript{50}-the midpoint of the transition from the
high-pH to the low-pH conformation, within pH $\pm$ 0.2. Nonetheless, the thermal stability values could not be accurately estimated by the support vector regression models based on the primary antibody sequence alone.

\subsection{Prediction of the long-term stability of mAbs in real storage conditions}
\cite{gero20} proposed to use ANN models to predict the monomer retention (long term stability) of therapeutic protein formulations in real storage conditions including refrigerated condition, room temperature and elevated temperature after long-term storage, i.e., six months. The authors claimed that the analysis of samples stored for two weeks at elevated temperatures, i.e., 25\textdegree C and 40 \textdegree C, using a size exclusion chromatography coupled with a multi angle laser light scattering in accelerated stability studies is a crucial factor for predicting monomer retention after long-term storage at all refrigerated, room, and elevated temperatures. In addition, the research presented the use of ML models such as ANNs, linear regression, and decision trees for prediction of the real-time protein stability in real storage conditions. Fourteen protein therapeutics under twenty-four different formulation conditions were used in the study. The authors hypothesised that by using the outcomes of appropriately designed short-term accelerated stability assays, protein stability fingerprints that depend on the formulation can be produced, and these characteristics would then serve as inputs for ML models to estimate the actual protein stability in real-time.

\section{Open data sources for research on the applications of ML in biopharmaceuticals}\label{open_datasource}

Obtaining data in biopharmaceutical studies is costly and time-consuming due to the need for conducting biological experiments that take a significant amount of time. Furthermore, these data are often associated with strategic products and confidential information of pharmaceutical companies, making it challenging to publicly share them with the research community. This section attempts to describe several open data sources in the literature to assist in accelerating the applications of ML algorithms within the biopharmaceutical area.

Table \ref{table_protein_db} summarises antibody and protein databases, which are synthesised from \cite{akba22, khcu22, dodo21, noam20}. These are useful databases for the studies related to discovering prospective biologic candidates, evaluating their computational developability as well as designing mAbs and protein products as presented in section \ref{section_early_stage}.

\setlength\LTleft{0cm}
{
\scriptsize{
\begin{longtable}[t]{lL{3cm}L{8cm}L{4cm}}
    \caption{Relevant antibody and protein databases for the studies related to early stages of biologics development (synthesised from \cite{akba22, khcu22, dodo21, noam20}).} \label{table_protein_db} \\
    \toprule
    \textbf{ID} & \textbf{Database name} & \textbf{Description} & \textbf{Link} \\
    \toprule
    \multicolumn{4}{l}{\textbf{Sequence databases}}\\
    \toprule
    1 & abYsis &  An integrated database of antibody sequence and structure data. & \url{http://www.abysis.org/}\\
    \midrule
    2 & International Immunogenetics Information System (IMGT) & 
    This is a specialised knowledge resource focused on sequences, genomes, and structures of the immunoglobulins, T cell receptors, major histocompatibility complex (MHC), MHC super-families as well as related proteins within the immune systems of both vertebrates and invertebrates. & \url{http://www.imgt.org/}\\
    \midrule
    3 & iReceptor & This database contains antibody/B-cell and T cell receptor repertoires from many different independent repositories. & \url{https://gateway.ireceptor.org/login}\\
    \midrule
    4 & Observed Antibody Space (OAS) & This database includes immune repertoires of more than one billion annotated antibody sequences spanning various immune states and organisms. & \url{https://opig.stats.ox.ac.uk/webapps/oas/}\\
    \midrule
    5 & Patented Antibody Database & This database comprises antibody sequence information found in patent documents with 267,722 antibody chains (148,774 heavy chains and 118,948 light chains) from 19,037 patent families. & \url{https://www.naturalantibody.com/pad}\\
    \midrule
    6 & Integrated Nanobody Database for Immunoinformatics (INDI) & An integrated repository of nanobodies (single domain antibodies) with structure and sequence information from heterogeneous public sources. & \url{http://research.naturalantibody.com/nanobodies} \\
    \midrule
    7 & Antibody database & A repository of antibody sequences collected from all major public sources of therapeutic antibodies. & \url{https://naturalantibody.com/antibody-database/} \\
    \midrule
    8 & AntiBodies Chemically Defined (ABCD) & A manually curated repository of sequenced antibodies. &  \url{https://web.expasy.org/abcd/}\\
    \midrule
    9 & EMBLIg & This repository contains antibody sequences (light and heavy antibody variable domain) extracted from EMB antibody database. & \url{http://www.abybank.org/emblig/}\\
    \midrule
    10 & cAb-Rep & This database comprises 306 B-cell immunoglobulin sequence repertoires collected from 121 human individuals. & \url{https://cab-rep.c2b2.columbia.edu/}\\
    \midrule
    11 & sdAb-DB & This database includes natural and synthetic camelid single domain antibody sequences collected from various literature sources. & \url{http://www.sdab-db.ca/}\\
    \toprule
    \multicolumn{4}{l}{\textbf{Structure databases}}\\
    \toprule
    12 & Protein Data Bank (PDB) & This database contains 3D structure data for large biological molecules including proteins, DNA, and RNA suitable for research in fundamental biology, energy, health, and biotechnology. & \url{https://www.rcsb.org/} \\
    \midrule
    13 & Structural Antibody Database (SAbDab) & A database contains the structures of antibodies and nonobodies in the PDB annotated with a number of properties such as experimental details and heavy-light pairings. This database also includes affinity data for antibody-antigen complexes. & \url{https://opig.stats.ox.ac.uk/webapps/newsabdab/sabdab/}\\
    \midrule
    14 & Thera-SAbDab & This database curates immunotherapeutic variable domain sequences and their corresponding structural representatives of all WHO-recognised antibody therapeutics. & \url{https://opig.stats.ox.ac.uk/webapps/newsabdab/therasabdab/search/}\\
    \midrule
    15 & AbDb & This antibody data resources is a compilation of antibodies and nanobodies extracted from the PDB with information on redundancy and structures tackled with and without antigens for Fv fragments. & \url{http://www.abybank.org/abdb/}\\
    \midrule
    16 & AAAAA & An atlas of antibody anatomy with the structural alignment of antibody and T cell receptor sequences. & \url{https://plueckthun.bioc.uzh.ch/antibody/index.html}\\
    \midrule
    17 & SACS & A summary of crystal structures of the antibodies extracted from the PDB. & \url{http://www.abybank.org/sacs/}\\
    \midrule
    18 & PyIgClassify & This database provides the clusters and associated information of the antibody complementarity determining regions loop conformations of 3,065 antibodies extracted from the PDB. & \url{http://dunbrack2.fccc.edu/PyIgClassify/}\\
    \toprule
    \multicolumn{4}{l}{\textbf{Immunogenicity}}\\
    \toprule
    19 & Immune Epitope Database (IEDB) & This database contains experimental data on humans, non-human primates, and other animal species antibody and T cell epitopes. & \url{https://www.iedb.org/}\\
    \midrule
    20 & Bcipep & This repository contains B-cell epitopes with the peptides showing various measures of immunogenicity. & \url{https://webs.iiitd.edu.in/raghava/bcipep/info.html}\\
    \midrule
    21 & MHCBN 4.0 & A database contains the information about MHC binding, non-binding peptides, and T-cell epitopes. & \url{http://crdd.osdd.net/raghava/mhcbn/}\\
    \midrule
    22 & Leadscope Toxicity Database & This toxicity database contains over 200,000 chemical structures with over 600,000 toxicity study results of genetic toxicity, cancer, reproductive/developmental toxicity, acute and chronic toxicity, and many other endpoints & \url{https://www.leadscope.com/product_info.php?pro
ducts_id=78}\\
    \midrule
    23 & SYFPEITHI & This database consists of over 7000 peptide sequences known to bind class I and class II MHC molecules. & \url{http://www.syfpeithi.de/0-Home.htm}\\
    \toprule
    \multicolumn{4}{l}{\textbf{Antibody-antigen binding/Protein-protein interactions}}\\
    \toprule
    24 & PROXiMATE & A mutant protein-protein interaction kinetics and thermodynamics database consists of thermodynamic data for mutations in protein-protein complexes (including antibody-antigen complexes). & \url{https://www.iitm.ac.in/bioinfo/PROXiMATE/}\\
    \midrule
    25 & AgAbDb & A database stores the interactions between the  antigen-antibody molecules including residues of binding sites and interacting residue pairs. & \url{http://bioinfo.net.in/AgAbDb}\\
    \midrule
    26 & CoV-AbDab & A curated database of published or patented binding antibodies and nanobodies able to bind to coronaviruses, including SARS-CoV2, SARS-CoV1, and MERS-CoV & \url{http://opig.stats.ox.ac.uk/webapps/covabdab/}\\
    \midrule
    27 & SKEMPI v2.0 & A database stores information regarding alterations in thermodynamic parameters and kinetic rate constants that occur as a result of mutation for protein-protein interactions (including antibody-antigen complexes)  where the structure of the complex has been solved and can be found in the PDB. & \url{https://life.bsc.es/pid/skempi2/}\\
    \midrule
    28 & AntiJen & A database contains quantitative binding data for peptides in the context of vaccinology and immunology. & \url{http://www.ddg-pharmfac.net/antijen/AntiJen/antijenhomepage.htm} \\
    \midrule
    29 & PCLICK & A respository stores antibody-antigen structures based on the analysis of 403 antibody-antigen complexes. & \url{https://mspc.bii.a-star.edu.sg/minhn/cluster_pclick.html}\\
    \midrule
    30 & AB-Bind & A database stores experimental results for mutants for antibody-antigen structures including the change in Gibbs binding free energies. & \url{https://github.com/sarahsirin/AB-Bind-Database}\\
    \toprule
    \multicolumn{4}{l}{\textbf{General Information, Regulatory}}\\
    \toprule
    31 & Tabs: Therapeutic Antibody Database (commercial use) & A database stores information on about 2000 therapeutic antibodies targeting over 750 antigens, which are being developed by more than 600 companies. & \url{https://tabs.craic.com/static_pages/4} \\
    \midrule
    32 & ImmPort & An open access platform for clinical and molecular datasets. & \url{https://www.immport.org/shared/home} \\
    \midrule
    33 & AbMiner & This is a search tool for antibodies that enables users to scan through commercially available antibodies and match each one with its corresponding genomic identifier. & \url{https://discover.nci.nih.gov/abminer/}\\
    \bottomrule
\end{longtable}
}}

Compared to the diversity of biophysical, structural, and sequence data available for antibodies and proteins, open data on cell cultures and downstream processing is very limited. This is because the bioprocess development and manufacturing are expensive and time-consuming, and pharmaceutical companies often keep biological product information confidential and copyrighted. \cite{gase21} provided an open dataset\footnote{\url{https://ars.els-cdn.com/content/image/1-s2.0-S0098135421000041-mmc3.xlsx}} covering seven years, from 2010 to 2016, for the upstream process development of CHO cell lines used in the manufacturing of various antibody products. However, to protect the proprietary rights, the dataset was anonymised and normalised between 0 and 1. The dataset includes 106 cultures that represent a wide range of operational scales, from bench-top (5L volume) to manufacturing (500L volume). For each culture, 25 parameters were recorded for up to 17 days, including culture days, elapsed culture time, elapsed generation number, average cell compactness, average cell volume, average cell diameter, pH, temperature, pCO\textsubscript{2}, DO, osmolality, viable cell density, total cell density, cell viability, cumulative population doubling level, concentrations of glucose, glutamine, glutamate, ammonium, lactate, sodium, potassium, and bicarbonate, monomer percentage of the final product, and mAb concentration. The authors provided two versions of the dataset: one with missing values and another in which missing values were imputed using the method proposed in \cite{gase21}. In addition, another subset of the data including 45 cultures contains five additional variables, i.e., endpoint day, midpoint day, batch, culture volume, and cell line, apart from the above 25 parameters. In contrast to the first dataset, this dataset is not a time-series data, and it only stores the values for variables on the harvest day and the day corresponding to halfway through the process. In this subset, 10\% of batches are the failed runs. By analysing the failed and successful runs, it is possible to identify the events and impact of variables leading to the failure \citep{gase21}. This second dataset is also appropriate for assessing the knowledge transferability of ML models across different cell lines.

Due to the scarcity of open upstream cell cultures data, several studies have used mechanistic models to generate synthetic data from which ML models will be developed. \cite{hust21} built an emulator model using mechanistic models presented in \cite{nabe20}. This emulator was then used to generate concentration measurements of glucose, lactate, glutamine, ammonia, VCD, and titer using different configurations of parameters for mechanistic models of mammalian cell cultures, and these simulated values were added with small Gaussian noise to replicate the analytical measurement errors. An example dataset was shared online\footnote{\url{https://github.com/rauwuckl/BioprocessExampleData}}. This dataset was used to assess the knowledge transferability across cell lines of hybrid models using entity embedding vectors \citep{hust21}.

\cite{gost15} developed a simulator for an industrial-scale penicillin product semi-batch 100,000L bioreactor using first principle mechanistic models representing the mechanisms of substrate intake, cell growth, and penicillin production. The simulator also integrated the impact of multiple environmental parameters such as temperature, pH, pCO\textsubscript{2}, DO, viscosity, nitrogen and phenylacetic acid flow rates on the biomass and penicillin concentrations. The source code of the simulator is free to download at \url{www.industrialpenicillinsimulation.com}. In the later study, \cite{godu19} incorporated a realistic simulated Raman spectroscopy device into the above \textit{IndPenSim} simulator for the development, evaluation, and implementation of advanced and innovative control solutions in biotechnology facilities. The dataset generated by \textit{IndPenSim} for 100 batches of process and Raman spectroscopy measurements is shared publicly\footnote{\url{https://data.mendeley.com/datasets/pdnjz7zz5x/1}}. The 100 batches include various control strategies and batch lengths for a typical biopharmaceutical manufacturing facility. Batches 1-30 follow a recipe-driven approach, while batches 31-60 are operator-controlled. Batches 61-90 employed an advanced process control solution with Raman spectroscopy. Batches 91-100 exhibited process deviations due to faults. This dataset is well-suited for developing big data analytics, ML and artificial intelligence algorithms that can be applied to the biopharmaceutical industry sector.

The impact of chemical stability on both the efficacy and safety of protein therapeutics is a significant concern during development. Protein hotspots, which are amino acid residues susceptible to various post-translational chemical modifications such as deamidation, isomerization, glycosylation, oxidation, etc., play a crucial role in the chemical stability of protein therapeutics. The early elimination or reduction of these hotspots in the drug discovery process can be achieved through accurate prediction methods for identifying these potential hotspot residues. \cite{jisu17} presented a method of using ML models for prediction of protein asparagine deamidation. The authors built the training datasets from asparagine residues on 25 protein structures, which combines experimental measurements (such as penta peptide deamidation half-life) with structural features to describe the deamidation mechanisms. The dataset is freely available for download\footnote{\url{https://doi.org/10.1371/journal.pone.0181347.s001}}.

In the studies regarding antibody stability, the aggregation rates and viscosity can be used to develop predictive models for antibody stability \citep{laga22}. \cite{laga22} built a dataset based on molecular dynamics of the full-length antibody and sequence information of 20 preclinical and clinical-stage mAbs (18 IgG1 and 2 IgG4 were manufactured by AstraZeneca). The accelerated aggregation rates were measured at 45\textdegree C, while the viscosity was measured at multiple concentrations ranging from 80 mg/ml to 250 mg/ml depending on mAb sample. These features were then used to train ML models. The dataset is available for download \footnote{\url{https://www.tandfonline.com/doi/suppl/10.1080/19420862.2022.2026208/suppl_file/kmab_a_2026208_sm3851.zip}}.

To enhance the understanding in the field of protein formulation, long-term stability, and thermal stability characterisation, the researchers from the Technical University of Denmark created a database, called `Protein-excipient Interactions and Protein-Protein Interactions in formulation (PIPPI)'\footnote{\url{https://pippi-data.kemi.dtu.dk/overview}}. \cite{gero20} used the data in this database to train ANNs which are able to precisely predict the real-time aggregation of therapeutic proteins in pharmaceutically relevant formulations and their long term stability in real storage conditions.

\section{Opportunities for potential applications of ML towards BioPharma 4.0}\label{potential_direction}

The biopharmaceutical industry is undergoing a digital transformation towards Biopharma 4.0 due to the demand for more effective treatments and streamlined manufacturing processes \citep{puda22}. This is being driven by the development of new technologies, including advanced computational power, the deployment of smart sensors, PAT instrumentation, digitisation, and analytical tools. To achieve such a digital transformation, it is necessary for the biopharmaceutical industry to implement advanced ML based models for data analysis and predictions, as well as automation and Internet of Things solutions for the system connectivity. By leveraging these technologies, biopharma companies can enhance operational procedures, increase process efficiency, and boost productivity, leading to faster decision-making during the bio-manufacturing process. Machine learning has made significant contributions to almost all areas within the biochemical engineering as reviewed in \cite{mosa21}. However, there are still fundamental challenges that need to be addressed to facilitate its future applications. Additionally, there are several new machine learning techniques that show great potential for use in the bioprocess development and manufacturing of biologics, but they have yet to be fully explored in the literature. This section will cover potential applications of machine learning in the innovation of the biopharmaceutical industry.

\subsection{Digital twins}
A digital twin (DT) is a virtual representation of a physical system that mimics its behavior and dynamics. A complete DT consists of both physical and virtual components, with automated data communication between the two facilitated by an integrated data management system \citep{chya20}. According to \cite{sost21}, in the context of biopharmaceuticals, the goal of a DT is to create a digital technology that can access all available information in process data archives across different scales and sites aiming for a real-time information exchange with the complete process control system and all human stakeholders involved. The DT would be able to transform data independently of their format into a relevant context, capturing the required level of complexity and simulating various future scenarios to derive relevant decisions. Therefore, this technology has a huge potential to transform the biopharmaceutical industry by improving process efficiency, prediction, decision making, reducing costs, risk analysis, and increasing product quality. DTs can benefit scientists by providing real-time insights and suggestions for the process validation. Process development teams can use DTs for efficient planning and analysis of iterative campaigns. DTs also play a central role in managing data, knowledge, and operations, providing a transparent basis for risk assessment. They connect end-users, regulatory authorities, vendors of hardware and software solutions, unifying data standards and managing expectations for digital transformation \citep{sost21}. As a result, DTs are crucial for achieving a real-time monitoring and control of process parameters and critical product quality attributes in the biopharmaceutical industry.

\begin{figure}[!ht]
    \centering
    \includegraphics[width=0.95\textwidth]{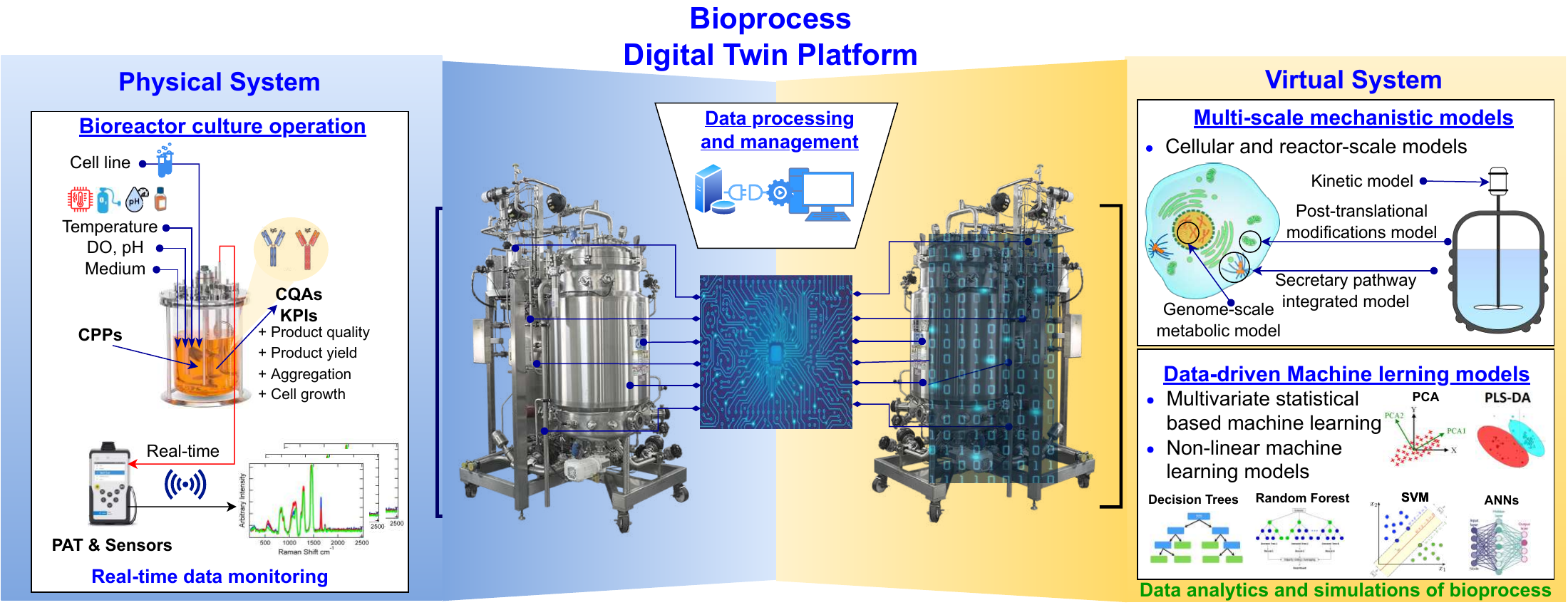}
    \caption{A concept of bioprocess digital twin platform including physical system, data processing and management, and virtual system blocks (adapted from \cite{papa21}).}
    \label{ditigal_twin}
\end{figure}

\cite{papa21} introduced a bioprocess digital twin platform that integrates physical and virtual systems through a data management system collecting a real-time culture data as shown in Fig. \ref{ditigal_twin}. The bioprocess DT includes multiple interconnected components, such as physical operations of cell culture, real-time bioreactor monitoring, data processing and management, as well as mechanistic and ML modeling of both cells and bioprocesses. Recent advancements in sensor and PAT tools such as Raman and near-infrared spectroscopies have enabled real-time monitoring of multiple process parameters in bioreactors. The historical and real-time data can be processed and stored in a database management system that can be used to connect the physical and virtual parts of the bioprocess DT platform. The virtual system replicates the physical processes, including cells and bioreactors, and their dynamic behaviors are simulated using knowledge-based mathematical and mechanistic models. Additionally, process inputs and outputs can be analysed using data-driven ML approaches. The combination of two modeling approaches, namely mechanistic and data-driven, can be mutually beneficial to capture the metabolic and regulatory states of entire cells and culture behaviors. This approach can effectively identify bottlenecks in the process and key engineering targets, as well as suggest various operational strategies to manage product quality. \cite{sost21} argued that hybrid models are a backbone technology to achieve an effective implementation of the DT platform.

However, the integration of virtual and physical plants in biopharmaceutical manufacturing has not yet been fully developed so far \citep{chya20}. In order to achieve biopharmaceutical digital transformation, several requirements still need to be fulfilled, including the development of real-time data acquisition, a dedicated data transfer system, effective control and execution techniques, robust simulation methods, anomaly detection, prediction tools, and easy access to a secure cloud server platform. We refer the readers to recent review papers with perspectives on biopharmaceutical DTs for more details, such as \cite{papa21, sost21, chya20, smju20}.

\subsection{Emerging of optical spectroscopy}
To develop DT solutions, it is essential to deploy PAT tools to measure multiple process parameters and quality attributes in the
bioreactor and collect a large sets of cell culture data during the monitoring process of the physical systems \citep{chya20, papa21}. New spectroscopy techniques have recently emerged that enable precise and real-time monitoring of both upstream and downstream processes as presented in Sections \ref{upstream} and \ref{downstream}. Industrial applications of spectroscopy are primarily concentrated on monitoring cell growth and quantifying nutrient and metabolite concentrations in culture fluids. In particular, UV/Vis and multiwavelength UV spectroscopy have been employed to measure protein concentrations in real-time during production \citep{clau17}. Meanwhile, NIR has been utilised for testing raw materials and final products \citep{clau17}. Raman spectroscopy has been deployed for the measurements of viable cell density, metabolite, nutrient, and antibody concentrations \citep{bedo15, mela15, poma22}. Moreover, spectroscopy techniques can be employed to monitor critical quality attributes of the manufacturing process, such as host cell proteins and post-translational modifications of proteins \citep{roma16, brru18, goum20}. Studies have also demonstrated that in-line Raman spectroscopy and Mid-IR are capable of monitoring various parameters, including protein concentration, aggregation, host cell proteins, and charge variants \citep{escu21, ware20}. However, to build high accurate ML models from spectroscopy data, it is required to perform several data preprocessing steps, such as background correction, spectral smoothing, and noise removal. Many various applications of spectroscopic sensing in upstream and dowstream processing have been reviewed in the literature \citep{wewo21, escu21, yime20, gust19, clau17, bury17, roma16}.

As presented in Section \ref{monitoring_cell}, the monitoring of cell culture processes within bioreactors has been dominated by advanced in-line Raman monitoring systems. Nevertheless, the utilisation of Raman-based monitoring methods is currently constrained by the inherently weak signal and the significant fluorescence interference from various biomolecules in the culture broth \citep{bury17}, which may have negative impacts on the quantification of parameters related to cells and bioprocess. Hence, the deconvolution of Raman signals and the implementation of advanced analyses using ML algorithms for complex biological compounds are crucial for eliminating distorted process information, thereby facilitating enhanced interpretation and deeper comprehension of the bioprocess \citep{papa21}. In the future, Raman spectroscopy could be coupled with advanced data analysis techniques and robust statistical models, using ML and deep learning algorithms for signal processing. Furthermore, Raman spectroscopy could be integrated with other biosensors, such as dielectric spectroscopy, to create a multi-attribute sensing system that enables rapid and simultaneous monitoring of all aspects of the bioprocess and cell culture data \citep{escu21}, and ensuring the consistent quality of biotherapeutic products \citep{papa21}.

\subsection{Handling small data}
The requirement for a large amount of training data is a fundamental limitation of machine learning \citep{mosa21}. However, in the biopharmaceutical industry, datasets often have a higher number of measurements than observations \citep{seva15}. Typically, measurements are taken only during a few planned times as the batch moves through the production process, and few replicates are conducted due to time and cost limitations. This can pose significant challenges when applying machine learning to small datasets with infrequent feedback in bioprocess development and manufacturing data. To build high performing predictive models, it is required to design specific algorithms for small data sets. Regularization techniques have been recognized as potential solutions for such issues, as they can handle both input selection and model estimation at the same time to avoid over-fitting \citep{pasc11}. \cite{seva15} combined a regularization method, i.e., the elastic net, with Monte Carlo sampling techniques to build an effective learning model from limited training datasets. The elastic net with Monte Carlo sampling algorithm merges the advantages of the elastic net algorithm, which enables simultaneous model selection and parameter estimation, with the strength of Monte Carlo sampling to prevent potential data overfitting, resulting in a reliable, understandable, and straightforward process model.

The second method to deal with the limited amount of training data is the use of various data augmentation strategies. One simple method is to introduce random noise to the initial data. Given the intrinsic stochastic nature of bioprocesses, it can be assumed that any data point might deviate within an acceptable range from the original data and still reflect the underlying physical nature of the process. This range of noise can originate from experimental measurement variations or empirical knowledge. As a result, a considerable number of synthetic data can be generated from a typically limited set of experimental data. This data augmentation technique has proven to be effective for constructing machine learning models in bioprocess modeling as well as soft-sensing as demonstrated in \cite{tuga18a, tuga18}. In additional to random noise, \cite{tuga19} proposed a probabilistic machine learning framework to build a probabilistic model from empirical batch process data. Then, this model can be used in combination with probabilistic programming to create an arbitrarily large number of dynamic \textit{in silico} datasets for training ML models. However, a drawback of these data augmentation methods is that they consider the small historical training sets as a representative of the whole sample space to generate synthetic datasets.

Another approach to tackle the small training data issues is incorporating process prior knowledge into data augmentation by building and fitting mechanistic models, such as classical kinetic models, based on initial experimental datasets. Then, these mechanistic models can be used to generate synthetic data under various experimental conditions as demonstrated in \cite{hust21, godu19, kost19}. In contrast to the random noise technique, this method can provide extra biophysical understanding in the generation of synthetic data and enhance the precision of the machine learning model. Nonetheless, this method requires solid understanding underlying cellular and bioprocess mechanisms. In addition, minimisation of the difference between a straightforward kinetic model and experimental bioprocess data is not a straightforward issue. Therefore, this approach has not been commonly used in practice.

\subsection{Uncertainty quantification}
Biochemical processes are often less reproducible compared to chemical processes because of the complex and stochastic nature of metabolic reaction networks and enzymatic reaction kinetics \citep{mosa21, stki16}. Therefore, models used to simulate biochemical processes need to have not only high accuracy but also low uncertainty. There are two main approaches to estimate uncertainty in a data-driven model \citep{mosa21}. The first approach is to use ML methods with ability to automatically estimate uncertainty. Bayesian learning-based algorithms, such as Gaussian processes and Bayesian neural networks, are often used for this type of machine learning model. These methods are best suited for simulating systems with small amounts of data. However, the use of these Bayesian learning-based algorithms in biopharmaceutical process development and manufacturing is less common and still in its early stages compared to PLSR, PCA, tree-based models, and artificial neural networks. The second approach is to use bootstrapping, which can be applied to most machine learning algorithms. However, it can only approximate the true model uncertainty instead of estimating it accurately. The idea behind this approach is to train several machine learning models with the same structure through bootstrapping and then calculate the variation between these models. In future studies, it is essential to consider uncertainty analysis since it plays a critical role in assisting decision-making tasks. These tasks include identifying the structure of data-driven models, optimising robust bioprocesses, predicting metabolic pathways \citep{mosa21}, estimation of missing values, and designing experiments that meet a maximum uncertainty constraint \citep{wage17}.

\subsection{Moving from fed-batch manufacturing to continuous manufacturing}
In addition to developing new PAT tools for online data collection and high performing predictive models, another trend aiming for a digital transformation of the biopharmaceutical manufacturing industry is the shift of numerous bioprocesses from a batch to continuous operations \citep{hose18}. According to \cite{bofl22}, continuous bioprocessing is becoming increasingly popular due to its benefits, such as consistent nutrient conditions, the elimination of by-products/waste, the ability to produce high-density cultures, and no lag phase once the system is in operation. \cite{pako20} argued that the implementation of end-to-end continuous biomanufacturing is considered as one of the major breakthroughs in the biologics field, as it has the potential to reduce processing times, enhance productivity, ensure consistent product quality, and lower costs \citep{pabu19}. Transitioning to continuous operation also presents the chance to cut down on capital costs by considerably reducing the size and footprint of equipment, such as employing smaller bioreactors and chromatography columns, and eliminating hold tanks \citep{koco15}.

According to \cite{koco15}, the availability of data provided by advanced PAT tools and related feed-forward and feedback control systems will be crucial in ensuring consistent and reliable continuous operations in the long run. To realise the practical implementation of end-to-end continuous biopharmaceutical manufacturing, the technological advances in both upstream and downstream are required \citep{pabu19}. In upstream bioprocessing, perfusion bioreactors have already been employed in the production of recombinant proteins, as evidenced by studies such as \citep{lupa15, lupa16, huwo18}, which demonstrate encouraging outcomes in terms of reducing product heterogeneity. \cite{bich19} compared the quality of a conjugated fusion protein produced through perfusion mode with that from two fed-batch operating strategies. Their findings indicated that the application of perfusion cell culture led to a decrease in product fragmentation and aggregation, while enhancing the homogeneity of its glycan profile. Compared to continuous upstream processing using perfusion bioreactors, continuous downstream processing in the biopharmaceuticals industry has been immature and limited in applications \citep{hose18}, although they have been studied in academia for decades \citep{ju13, loan18}. It is mainly because of the difficulty in fully understanding of complex process dynamics and challenges in process performance and capacity utilisation \citep{pako20}. Therefore, the practical implementation of end-to-end continuous processing in replacement of the batch-to-batch processing in biologics manufacturing has been considered relatively cautiously \citep{chma17}.

Continuous manufacturing processes bring about new challenges with regards to the process control \citep{paqu17}, particularly in managing the spread of impurities and other disruptions resulting from the tight integration of continuous unit operations. The primary source of difficulties in operating and controlling continuous processes is the absence of appropriate process analytics tools capable of providing the necessary measurements to ensure product quality \citep{pabu19}. Therefore, to ensure a successful realisation of continuous biopharmaceutical manufacturing, it is required to develop solutions for integrated continuous monitoring and control of whole upstream and downstream processes \citep{duhe18}. The successful implementation of plant-wide control strategies resulting in optimal overall process operations for small-molecule pharmaceuticals as reported in \citep{labe13, labe14},  offer excellent examples which are potentially transferable to biopharmaceuticals. However, the implementation of continuous manufacturing for biologic drugs is more complex compared to small-molecule pharmaceuticals. \cite{jibr16} indicated that on-line real-time sensors are less available in biomanufacturing than for small molecules, making control strategies for biopharmaceutical manufacturing more heavily reliant on open-loop design space rather than close-loop feedback control. Additionally, the mechanistic understanding of some bioprocesses is not fully understood, making it more challenging to control post-translational modifications.

In the literature, there have been several initial realisation of small scale continuous bioprocesses. \cite{wago12} presented one of the initial fully continuous, pilot-scale bioprocesses for producing a monoclonal antibody and a recombinant human enzyme. The design involved a 12L perfusion bioreactor connected to an Alternating Tangential Flow (ATF) cell retention equipment, as well as a downstream processing train that incorporated a filter, a surge bag, and a 4-column Periodic Countercurrent Chromatography (PCC) unit. \cite{goko15} demonstrated an end-to-end continuous bioprocess by connecting a perfusion bioreactor to an ATF cell retention device, followed by a downstream  purification process using two 4-column PCC systems. Moreover, \cite{kast16} introduced a laboratory-scale continuous production process for monoclonal antibodies, which involved a perfusion cell culture, a surge tank, and a continuous capture process. \cite{stul17} implemented an end-to-end integrated continuous lab-scale process for manufacturing monoclonal antibodies. The system consisted of a continuous cultivation with filter-based cell retention, a continuous two-column capture process, a virus inactivation step, a semi-continuous polishing step using twin-column countercurrent solvent gradient purification, and a batch-wise flow-through polishing step, which were all successfully integrated and operated together.

\subsection{Integrating bioprocess knowledge and knowledge transferability}
Section \ref{monitoring_cell} demonstrated how purely data-driven modelling approaches have significantly contributed to achieving the objectives of Biopharma 4.0 by providing robust data analytics tools and facilitating real-time monitoring through advanced PAT and soft sensors. However, their effectiveness in integrating systems within Biopharma 4.0 has been limited due to the absence of causal or mechanistic connections between process inputs and outputs. Despite the various machine learning techniques available, data-driven modelling inherently lacks physical insight and thus struggles to accurately extrapolate bioprocess behaviours \citep{mosa21}. Hence, it is crucial to integrate machine learning with established process mechanisms to expand the efficiency and comprehension of ML models in practical biopharmaceutical manufacturing. In contrast, purely knowledge-driven (mechanistic) models such as kinetic models can surpass the drawbacks of purely data-driven models by providing mechanistic links for growth rate \citep{krho17}, cell growth and death, dynamic metabolism \citep{nalu22, ohle14, xibi10}, or N‐linked glycosylation modelling \citep{koje19}. In general, the simplified kinetic models combine metabolic factors into a small set of differential equations, which typically include descriptions of cell viability, glucose and other nutrient concentrations, and ammonia and lactate concentrations. These models can then be integrated with process models to determine mass and energy balances, as well as fluid dynamics, which can be used for modeling cellular and metabolic processes in upstream bioreactors as demonstrated in \cite{kost19}. However, the adoption of these models within the biopharmaceutical manufacturing has been limited due to their complexity, the experimental requirements for parameterisation, and the high computational costs involved \citep{tsva21, basu20}. Moreover, the current kinetic models only represent the interactions of a small number of species within the cell culture, while media formulations can contain over 50 components, which can lead to hundreds of unknown metabolic interactions \citep{basu20}.

Essentially, knowledge-driven (mechanistic) modeling and data-driven modeling represent two opposing strategies for modeling \citep{gotu20}. Developing a mechanistic model requires a comprehensive understanding of the underlying processes, which can be time-consuming and labor-intensive. On the other hand, a data-driven model requires minimal physical knowledge, and can be rapidly deployed with flexible structures. However, constructing a data-driven model typically requires a larger dataset, and its validity may be limited to the specific conditions under which it was trained. According to \cite{sost21}, hybrid models address the limitations of the two original modeling approaches by leveraging the advantages of both. By combining the understanding and knowledge gained from the knowledge-centric approaches with the straightforwardness of the data-centric approaches, hybrid models can provide a more comprehensive understanding of the underlying behaviors of the systems. This approach is also cost-effective and modular, making it suitable for digital twin implementations \citep{tsva21}. Hybrid models can utilise the known underlying mechanisms to model specific cellular processes, such as metabolism, while relying on ML-based approaches to model other less explored processes \citep{tsva21}. Moreover, hybrid ML models can offer mechanistic insights into the relationships between CQAs and CPPs, enabling efficient real-time prediction and control through the use of appropriate bioprocess levers. These insights can also enable the rapid evaluation of clone performances and guide rational cell-line engineering by targeting the relevant and sensitive pathways \cite{wamy22}.

\cite{sohi17} discussed three main groups of combining mechanistic models with data-driven models:
\begin{itemize}
    \item \textit{Converting mechanistic models into data-driven models}. In this method, a mechanistic model is used to generate synthetic data, which is then used to train data-driven models, and the final predictive results on practical data are generated by data-driven models. There have been several typical studies \citep{hust21, godu19, kost19} in the literature using this method to overcome the difficulty of ML models development resulting from small training datasets. This method also enhances the insights into complex mechanistic models such as cell metabolism models as data-driven models can help discover the most influential parameters for a specific cell mechanism based on explanation approaches associated with the predictive models built from the synthetic data.
    \item \textit{Calibration of mechanistic models}. In this method, the data-driven models are used to estimate parameters of mechanistic models or are inserted as part of a mechanistic model, e.g. in a form of constraints, for process prediction \citep{nalu21, nalu22}. Most of the hybrid modelling studies presented in section \ref{hybrid_upstream} belong to this group. In this approach, the predictive results are generated by the mechanistic models. However, the parameters of the mechanistic models can be changed during operations due to the changes in the process conditions. In such situations, the mechanistic model loses its reliability and have to be re-calibrated during the process. In addition, the mechanistic models may have model parameters that change over time or have a dependency on process parameters that are not integrated into the model, such as biomass yields on substrates \citep{sohi17}. A data-driven model parameterised with historical data can be deployed to estimate such process variables with the desired frequency. However, key process variables, such as nutrient and metabolite concentrations, are typically measured offline with high experimental effort and time delay, which may be insufficient for highly dynamic processes. One solution is to use data-driven models that estimate these variables based on high-frequency process data, such as spectroscopy. An example of this approach is given in a study by \citep{crwh14}, where a data-driven model based on Raman data provides a feedback signal for a nonlinear model-predictive control in an online context.
    \item \textit{Hybrid models}. In this group, different data-driven models and meachanistic models can be connected in a block-oriented manner to create a hybrid model. The resulting model will be hybrid, meaning it combines both types of models. This approach has not yet been extensively applied in the biopharmaceutical field. However, the insights from the development of hybrid approaches in other engineering fields show a great potential for their applications in biopharmaceticals. \cite{vool14} reviewed three main composition structures for the hybrid models. The first one is the parallel composition, where a mechanistic model may not be accurate enough to model relevant phenomena, and thus the data-driven component is employed to compensate for the mismatch between the model and the actual process. The second structure is a serial composition, which is suitable for the case where no precise understanding of the underlying mechanisms is available but there is enough process data to deduce the unknown patterns. In this way, the empirical data can be used as inputs to the data-driven model to quantify the parts of the process that is difficult to compute by mechanistic models. These estimated outcomes or parameters are then adopted within the mechanistic models. The third composition structure is also a serial combination, where the predictions from the mechanistic models are considered as inputs to the data-driven models. In this way, the mechanistic part represents the balance equations and the inner relationships of the system, while the data-driven part describes the underlying phenomena that cannot be captured by the mechanistic approach. Similarly, \cite{bija20} proposed five new methods of combining first principles and machine learning models and assess them on a case study regarding multiphase flow-rate estimation in a petroleum production system. The first combining method uses the physically meaningful features generated by a feature engineering approach and the outputs from the first principles models as inputs to train the ML models. The second method employs the first principles models to compute the mismatch between the estimated values and actual values and uses these errors as targets to build ML models with the physically meaningful input features. The third method is similar to the second method but the ML model uses raw measurements as input features. The fourth approach deploys the first method to build sub-modules of the system and then combines the outputs from each sub-module into a linear meta-model. The fifth method combines the solution of the models built using the first method on physically meaningful input features with the model using the raw data as input features. In short, these combination approaches investigated from other engineering areas can be further assessed and applied to the biopharmaceutical field to enhance the quality of hybrid models.
\end{itemize}

\subsection{The use of new machine learning approaches and concepts}
Several novel machine learning methods such as reinforcement learning, transfer learning, online learning, meta-learning, and explainable ML have been widely used in many practical applications, but their applications in biochemical engineering have not been extensively explored with some changes which could only be observed in the last few years \citep{mosa21}. This section discusses potential applications of these algorithms to the bioprocess development and manufacturing.

\subsubsection{Transfer learning}
Transfer learning is a collective name used for a number of different methods used to maintain and transfer knowledge by adjusting the parameters or architecture of a trained ML model, allowing knowledge to be transferred from a previous inference task to a new one, even when the data source varies \citep{wekh16}. In recent years, researchers have shown great interest in transfer learning as a means of achieving efficient model training in situations where data is scarce \citep{gotu20}. By leveraging the knowledge gained through solving a well-defined problem, in principle, transfer learning enables that knowledge to be applied to a different, though related problem. 

In the biopharma research, it is difficult and costly to obtain large or evolving datasets from bioproduct manufacturing processes. There is, therefore, a potential for using transfer learning methods to build robust ML models for industrial-scale manufacturing processes based on the knowledge obtained from the construction of ML models for lab-scale processes during the bioprocess development phase. Furthermore, it is also important to investigate and propose novel knowledge transferability solutions between ML models developed for different cell lines \citep{hust21}. This can reduce the time and resources needed to develop new robust and reliable ML models for new cell lines leading to more efficient and cost-effective biopharma processes as it reduces the amount of experimental data required to build accurate models for different cell lines.

\subsubsection{Reinforcement learning}
Reinforcement learning (RL) \citep{suba18} is an approach which can be used for developing an optimal control policy in the presence of uncertain or stochastic process dynamics. RL offers several advantages over standard bioprocess control methods using mechanistic models and mathematical programming \citep{dubo22}. One key advantage is its flexibility to work with varying levels of mechanical knowledge and system structure. Model-free RL exhibits special characteristics that differentiate RL from other process control approaches such as as model-based RL. Rather than relying on explicit knowledge of the process dynamics, model-free RL uses Monte Carlo realisations of different control policies under the process dynamics to determine the optimal one \citep{mano20}. In this way, model-free RL is capable of addressing several types of complex systems and problems, including hybrid systems that combine continuous and discrete states, actions, and events, as well as problems with various objective functions such as tracking control, economic optimisation, and experimental design. Additionally, model-free RL can handle model uncertainties that are not limited to Gaussian distributions. The ability of model-free RL to provide control solutions without explicit knowledge of the process dynamics is particularly attractive in biopharmaceuticals, where the targeted systems are typically nonlinear and subject to uncertainty. This poses challenges for identifying mechanistic models that can be used for subsequent model-based optimisation and design tasks \citep{mano20, dubo22}. Inspired by these benefits of RL, several studies applied RL in optimising bioprocesses  \citep{niti21, pesa20}, developing data-driven control solutions \citep{mano20, pano18}, and building control strategies in combination with model knowledge \citep{ohpa22, mope22, kipa21} for bioprocesses. 

Despite the undeniable potential of model-free RL, applying it to biopharmaceutical applications presents several challenges. One of the main challenges is ensuring compliance with operational constraints to guarantee safety. A general RL solution policy aims to maximise expected rewards without providing an explicit mechanism for constraint satisfaction, which can pose a risk to bioprocess operations since microbial and mammalian cells are highly sensitive to their culture environments \citep{mosa21}. Another challenge associated with model-free RL for biopharmaceutical applications is the issue of data efficiency. RL algorithms rely on observations or samples of system responses under a given control policy. Due to the inherent stochastic nature of the systems, it may take numerous trial-and-error experiments to learn an optimal policy. However, obtaining a large volume of real-world data is often impractical and expensive in biopharmaceuticals. The combination of RL and process knowledge can address the challenge regarding the initialisation of an accurate RL policy by using a mechanistic model as a simulator for offline training. For example, \cite{kipa21} proposed a two-stage optimal control approach for closed-loop dynamic optimisation of a fed-batch bioreactor. The high-level optimiser employs differential dynamic programming, which is a model-based RL method using model gradients to maximise productivity over the long-term. In contrast, the low-level controller uses model predictive control for a short-term planning to track the high-level plan, eliminate disturbances, and account for the model-plant mismatch. Similarly, \cite{ohpa22} developed an integrated framework of a model predictive control and RL, where the value-based function obtained by a model-free method replaces the high computational cost function of the model predictive control. The authors validated their approach for optimising an industrial-scale penicillin bioreactor. The outcomes showed that the hybrid RL can effectively learn from fewer data than model-free RL approaches.

\subsubsection{Robust incremental learning with adaptation abilities}
Bioprocess data, such as nutrient and metabolite concentrations, typically undergoes changes during batch operations, and batch-to-batch data also exhibits dynamic changes. Hence, to construct robust machine learning models for this type of data characteristics, it is expected to use incremental (online) learning algorithms with adaptation capabilities \citep{kaga09a,kaga10,kagr11}. Incremental learning enables learning from new data samples as they become available during an ongoing process, assuming that this data comes from the same source. The objective of incremental learning is to extract the maximum amount of information from new data without forgetting previously acquired knowledge \citep{khru21}. In addition, the integration of different adaptive strategies \citep{bafa21} into ML algorithms can assist the automatic construction and updating of ML models to capture the dynamic changes in the input data over the course of process operation. Therefore, this learning approach is well-suited for dealing with the dynamic nature and diversity of bioprocess data.

According to \cite{gotu20}, online learning also contributes to improving model training under the limited data scenario, where information from new data can be integrated into the existing model when they are available \citep{zh14}. For example, in the context of batch process monitoring where data is limited, two adaptive PCA algorithms were proposed by \cite{liyu00} to recursively update the correlation matrix for adaptive process monitoring. \cite{luya04} introduced a two-dimensional dynamic PCA starting with a limited number of successful batches but which can be updated with the accumulation of successive batches to capture both within-batch and batch-to-batch dynamics simultaneously.

While there are some limited examples of using online learning with adaptive capabilities in biopharma, there have been a whole series of algorithms and approaches developed and applied for a variety of process monitoring and control applications in the broader process industry requiring adaptive soft sensors (see these publications for further details, \cite{kaga08,kaga09,kaga09c,kaga09b,kaga09a,kaga10,kaga11,kagr11,zlga12,buea14,sabu16,baga17}) which are readily available for a successful transfer to the bio-pharmaceutical production settings especially in the context of the recent push towards implementing various PAT technologies discussed previously in the review.

\subsubsection{Meta-learning}
Since every machine learning algorithm has its own inductive bias, and the No Free Lunch theorem \citep{adal19} states that no machine learning algorithm can outperform all others on all problems, a way to enhance the predictive performance of ML algorithms on a new dataset is to predict and suggest the most appropriate ML algorithm for a specific dataset. Meta-learning which is a \textit{learning to learn} method can be used to recommend the most suitable ML algorithm and parameter configurations for a new dataset based on previous learning experiences \citep{lega10,lebu15,albu20,kemu20}. Therefore, the use of meta-learning can automate, accelerate, and enhance the development of machine learning-based solutions \citep{kemu20, doke21, khke22}. Moreover, meta-learning can also help in the case of no or limited training data availability for a new problem \citep{alga18,albu19,gotu20}. For instance, \cite{fiab17} devised a model-agnostic meta-learning algorithm applicable to any neural network architecture that optimise initial parameters directly, enabling quick adaptation to new tasks using limited data. Their approach demonstrated superior performance in classification and reinforcement learning tasks compared to traditional transfer learning methods.

Constructing high-performing predictive models for upstream and downstream processes in biopharmaceutical manufacturing is challenging due to limited historical data and infrequent feedback from variables of interest that are difficult and expensive to measure. To overcome this drawback, a smart algorithmic solution is to take advantage of available prior information and transferable knowledge related to bioprocesses to establish meta-knowledge for complex decision-making systems \citep{so20}. In this way, meta-learning can be used to recommend the most suitable learning models based on the process mechanistic models and prior knowledge regarding previous development activities, then the proposed ML model can continue fine-tuning and adapting its predictive outcomes based on the limited available data \citep{lesh18}. The meta-features used in a meta-learning framework can be extracted directly from previous similar bioprocess data using statistical methods \citep{riga22}, but they also depend on the problem domain. In the biopharmaceuticals, apart from meta-features extracted from the bioprocess data, there have been many mechanistics models proposed in the literature which can assist in building meta-knowledge, such as existing mechanistic models for cell metabolism and protein glycosylation \citep{stru21, ma20, moku19, kost19, kova18, kyan17, garu17, spha16, crwh14, aebo12}, mechanistic models of charge variants \citep{kule15}, or mechanistic models for downstream chromatography \citep{homa16}.

\subsubsection{Explainable ML}
According to \cite{mosa21}, regardless of the type of machine learning algorithm used in biochemical engineering studies, the focus is always on the knowledge that can be derived from the data, rather than the data itself. Whether using a machine learning algorithm for bioprocess data classification or regression, if the results fail to provide additional physical insight or facilitate domain knowledge generation, the algorithm will not be widely used in biochemical engineering research. This fact also holds for the biopharmaceutical field. The main objective of analysing bioprocess data is to uncover valuable insights that may be concealed within vast amounts of data related to multiple process runs. Such insights can then be applied to improve and optimise the reliability and efficiency of production processes. In upstream processing, it is also important to explore the complex relationships between process variables and CQAs of the final product and identify the key variables impacting the quality attributes.

Therefore, it is highly desirable to use explainable ML algorithms \citep{tjgu20} for analysing and understanding the bioprocess data. Based on the explanations of underlying mechanisms leading to the predictive outcomes of data-driven models, insights related to the most influential parameters on a certain cell mechanism \citep{sohi17} can be captured and these insights may be used to improve the designs and control of bioprocesses or to better understand and develop new mechanistic models.

\subsubsection{Automated data preprocessing and feature engineering}
Feature engineering is a crucial step in the development of high performing predictive models \citep{khke22, kemu20}. To develop effective ML models for biopharmaceuticals, effective data preprocessing and feature engineering techniques play an even more important role as the bioprocess data is heterogeneous in both time scale and data types \citep{gase21, chhu08}. In particular, the electronic records of bioprocess data frequently come from multiple different data sources with a variety of inputs (e.g., nutrient and metabolite concentrations), process outputs (e.g., cell density and product concentration and quality attributes), control actions and physical parameters captured by various devices and at different time scales \citep{chhu08}. In addition, monitoring and control data collected from sensors and PAT tools typically includes a significant amount of noise. For example, spectroscopy data needs to be heavily preprocessed to eliminate unwanted signals and non-informative features prior to training ML models \citep{gava15}. Therefore, it is required to apply different data preprocessing and feature engineering techniques such as feature selection methods on heterogeneous
data and streaming data \citep{zlga12,lich17} to extract informative features which can assist in building effective ML model \citep{buea14}. Feature selection aims to achieve various goals, such as creating simpler and easier-to-understand models, enhancing data mining efficiency, and preparing clean and comprehensible data. To attain these objectives, it is possible to analyse a wide range of bioprocess runs to identify useful patterns, features, and valuable relations from different facets of the bio-manufacturing processes from the upstream, downstream purification, to product formulation and stability. Dynamic dimensionality reduction techniques can also be deployed to achieve this objective. 

In addition to feature selection and dimensionality reduction, automated feature engineering also aims to generate more complex, composite features with biophysical meaning which can enhance the performance of predictive models. According to \cite{bija20}, constructing physically meaningful features through basic feature engineering by combining the original raw measurements alone or with external data generated based on process knowledge can considerably improve the performance of learning models, compared to models trained on only raw features. Furthermore, by incorporating physics-based features into machine learning algorithms, it is possible to develop more interpretable models than the ML models using raw data directly. There have been several new methods of automated feature generation \citep{zhxu22, chli19} in the ML literature, which can be integrated with bioprocess knowledge to generate biophysically meaningful features aiming to improve the performance and interpretability of ML models in the biopharmaceuticals.

\subsubsection{Automated and autonomous machine learning}
As this review clearly illustrates there has been a huge increase in the applications of ML methods in biopharma in recent years and this growth is set to continue.

Given increasing availability of data, there are well established machine learning models development workflows which can and indeed has already been used to produce well performing ML models. However, despite the availability of various sophisticated tools \citep{scke22}, the process of developing, deploying and maintaining these ML models, especially for robust monitoring and control of industrial processes, can still be costly and both labour and knowledge intensive. As also discussed in detail in the section on challenges of biopharma processes with respect to the use and deployment of ML solutions (see Section \ref{challenges_biopharma}) the availability of the data, their quality and open-source databases in the pharmaceutical industry can vary quite dramatically depending on the specific application area.

The realisation of ambitious goals and aspiration behind the digital transformation, Biopharma 4.0, or development of Digital Twins will heavily rely on automation and ultimately, to a large extent, the autonomy of the ML models development, deployment and life-long accurate operation in changing environments.

The need and opportunities resulting from a full automation of ML models development, deployment and continuous adaptation was already recognised over a decade ago in the broader process industry where such fully automated solutions for the development of adaptive soft sensors were proposed \citep{kaga09a,kaga09,kaga09c,kagr11,sabu16,sabu18} and further challenges discussed in \cite{kaga09c}. Since then, the Automated Machine Learning (AutoML), as it has become to be refered to particularly in the recent years \citep{kemu20}, has undergone a major development (please see comprehensive recent reviews on the topic here \cite{kemu20,doke21,scke22,khke22}). Both AutoML and its next incarnation in the form of Autonomuos Machine Learning (AutonoML) \citep{kemu20}, espousing a much higher levels of autonomy, are likely to be instrumental in the realisations of digital transformations of the biopharma industry. 

\section{Conclusions} \label{conclu}
The biopharma industry is undergoing transformation due to the need for improved treatments and more efficient development and manufacturing processes. This is motivated by the development of advanced technologies such as computational power, smart sensor technologies, data management and communication systems, and analytical tools. A strategic initiative for digital transformation is necessary to take full potential advantage of these technologies. This requires the implementation of ML-based models for data analysis and predictions, along with automation and IoT for system connectivity \citep{puda22}. ML-based models have the potential to enhance operational procedures, process efficiency, and improve productivity in less time, enabling better decision-making during bio-manufacturing. Incorporating ML and AI-based infrastructure strategically offers the potential to transform the bioprocesses for biopharmaceutical development.

This paper presented a thorough assessment of how ML has been used in different stages of the development and manufacturing processes for mAbs and therapeutic proteins. Additionally, it seeks to identify the difficulties related to bioprocesses and process data and to propose potential research avenues in this domain towards building a digital transformation platform for the biopharmaceutical industry. A characteristic of bioprocess data is that it always involves knowledge that underlies the data, rather than the data itself, upon which ML models are built. Therefore, a critical note regarding the applications of ML in the biopharmaceutical field is that regardless of which machine learning algorithm is employed for bioprocess data classification or regression, their outcomes will not be useful in biopharmaceutical research if they cannot offer additional biophysical insights or contribute to domain knowledge generation and understanding. As a result, simply relying on machine learning to uncover the hidden knowledge of bioprocess data may be overly optimistic, given the data-driven nature of this approach. Hence, the most critical step towards enabling the continuous application of machine learning in future studies is to improve the accuracy of predictions and reduce uncertainty by integrating bioprocess knowledge into the model building process using different methods. In addition to exploring other advanced machine learning techniques, it is also important to give high priority to constructing novel mechanistic models for biopharmaceutical processes in both upstream and downstream processing. Overall, it is believed that the emerging relationships between data-driven, mechanistic, and hybrid modelling will continue to be fruitful and complementary to each other. The intelligent integration of these modelling techniques will also allow for the development of advanced digital twins, which can significantly expedite the development of biopharma from identifying lab-scale designs of experiments and optimise, monitor, and control industrial-scale operations.

\section*{Acknowledgment}
This research was supported under the Australian Research Council's Industrial Transformation Research Program (ITRP) funding scheme (project number IH210100051). The ARC Digital Bioprocess Development Hub is a collaboration between The University of Melbourne, University of Technology Sydney, RMIT University, CSL Innovation Pty Ltd, Cytiva (Global Life Science Solutions Australia Pty Ltd) and Patheon Biologics Australia Pty Ltd.

\bibliography{myref}

\end{document}